\documentclass[runningheads]{llncs}

\usepackage{eccv}

\usepackage{eccvabbrv}

\usepackage{graphicx}
\usepackage{booktabs}

\usepackage[accsupp]{axessibility}  % Improves PDF readability for those with disabilities.

\usepackage{adjustbox}
\usepackage{multirow}
\usepackage{amsmath}
\usepackage[table]{xcolor} % Enables row coloring
\usepackage{pgfplots}
\usepgfplotslibrary{groupplots}
\pgfplotsset{compat=1.18}
\usepackage{caption}  
\usepackage{floatrow}
\newfloatcommand{capbtabbox}{table}[][\FBwidth]
\usepackage{xspace}
\usepackage{tikz}
\usetikzlibrary{calc,positioning,arrows.meta}
\usepackage{hyperref}
\usepackage{orcidlink}

\usepackage{xcolor}
\definecolor{lightblue}{HTML}{eaf6ff}
\definecolor{lightgray}{HTML}{f4f7f8}
\definecolor{darkgreen}{HTML}{05961E}

\newcommand{\name}{SetDiff\xspace}
\newcommand{\p}{\mathbf{p}}

\newcommand{\y}{\mathbf{y}}
\newcommand{\z}{\mathbf{z}}
\newcommand{\I}{\mathbf{I}}
\begin{document}

\title{Enhancing Novel View Synthesis via \\ Geometry Grounded Set Diffusion}

\titlerunning{\name: Geometry Grounded Set Diffusion}

\author{
Farhad G. Zanjani \quad Hong Cai \quad Amirhossein Habibian \\ {\tt\small \{fzanjani, hongcai, ahabibia\}@qti.qualcomm.com}
}

\authorrunning{F. Zanjani et al.}

\institute{Qualcomm AI Research\footnotemark}

\maketitle
\begin{abstract}
We present \textit{\name}, a geometry-grounded multi-view diffusion framework that enhances novel-view renderings produced by 3D Gaussian Splatting. Our method integrates explicit 3D priors, pixel-aligned coordinate maps and pose-aware Plücker ray embeddings, into a set-based diffusion model capable of jointly processing variable numbers of reference and target views. This formulation enables robust occlusion handling, reduces hallucinations under low-signal conditions, and improves photometric fidelity in visual content restoration. A unified set mixer performs global token-level attention across all input views, supporting scalable multi-camera enhancement while maintaining computational efficiency through latent-space supervision and selective decoding. Extensive experiments on EUVS, Para-Lane, nuScenes, and DL3DV demonstrate significant gains in perceptual fidelity, structural similarity, and robustness under severe extrapolation. \name establishes a state-of-the-art diffusion-based solution for realistic and reliable novel-view synthesis in autonomous driving scenarios.

\url{https://qualcomm-ai-research.github.io/setdiff}
\end{abstract}

%%\href{https://qualcomm-ai-research.github.io/setdiff}{https://qualcomm-ai-research.github.io/setdiff}
%% This is required for papers in LNCS proceedings.
% \keywords{Novel View Synthesis \and Image Diffusion \and Gaussian Splatting}

\section{Introduction}
\begin{figure}[!b]
\centering
 \resizebox{0.8\textwidth}{!}{
  \includegraphics[width=\textwidth]{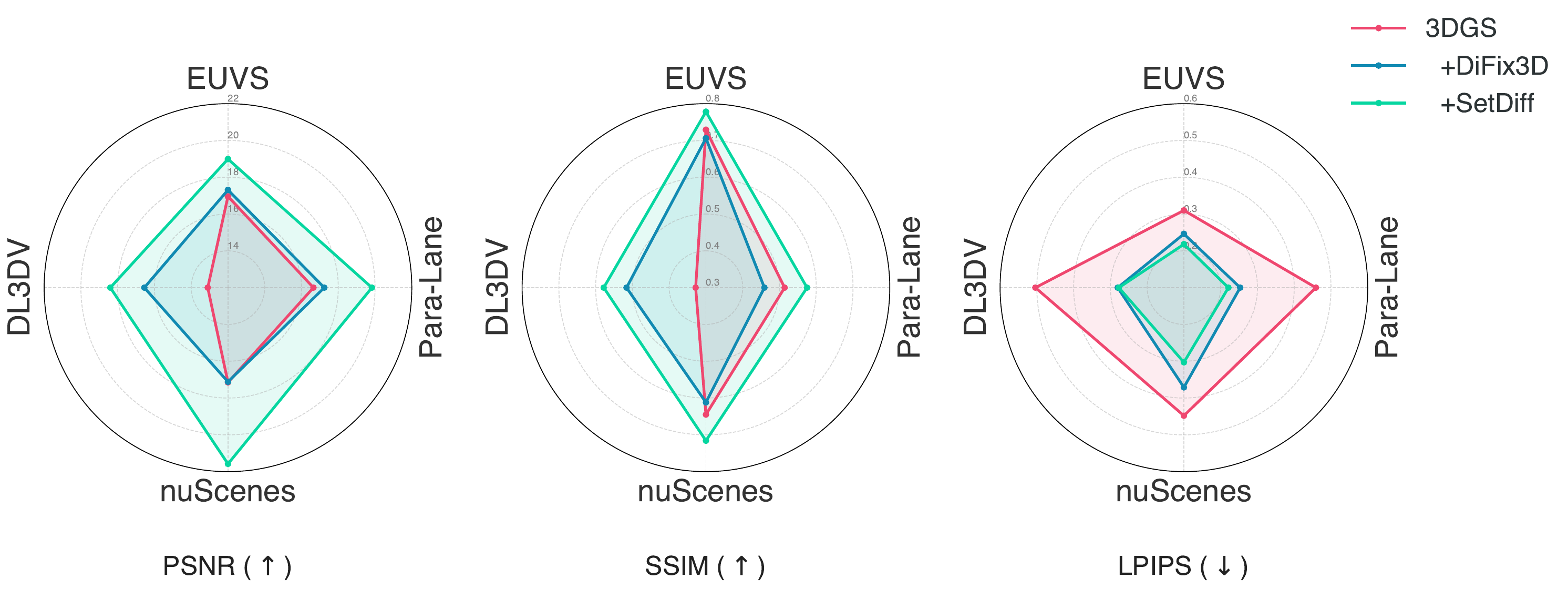}
  }
  \caption{Our proposed \name consistently improves 3DGS~\cite{liu20243dgs} performance for NVS, on four datasets, while outperforming DiFix~\cite{wu2025difix3d} as the SOTA diffusion-based enhancer.}  
\label{fig:teaser}
\footnotetext{** Qualcomm AI Research is an initiative of Qualcomm Technologies, Inc.}
\end{figure}

\begin{figure}[t!]
\centering
 \resizebox{\textwidth}{!}{
  \includegraphics[width=\textwidth]{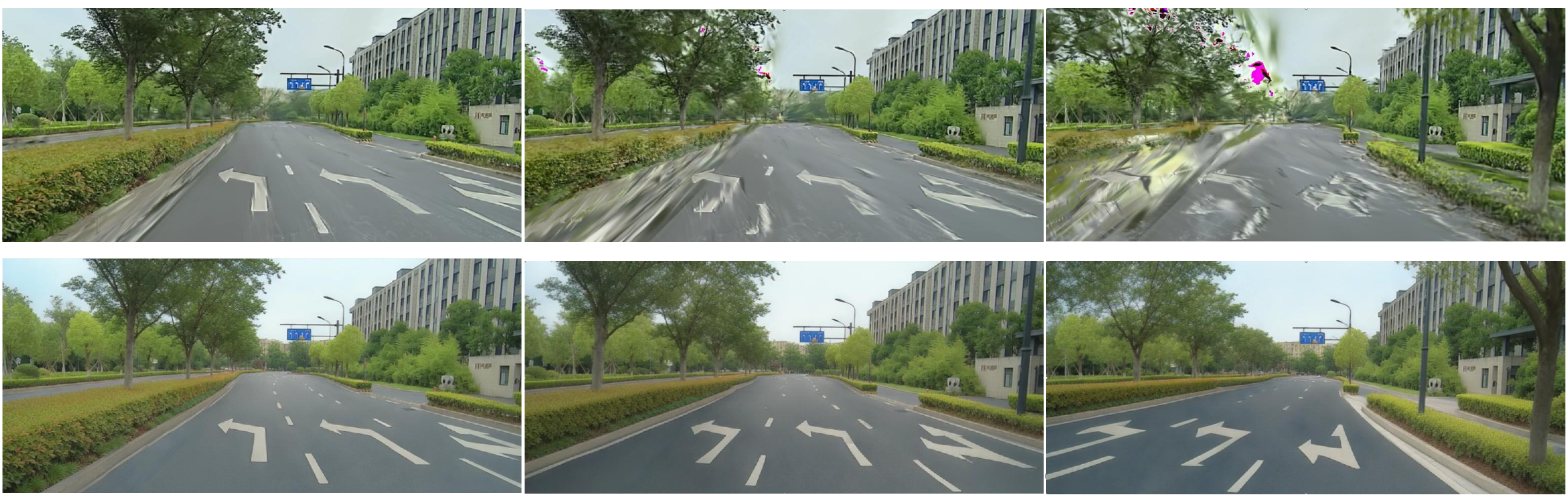}
  }
  \caption{Example of 3DGS rendering artifacts (top row) when deviating from the training trajectory (left) by one (middle) and two (right) lanes. The diffusion enhancement greatly removes the rendering artifacts (bottom row) even under extreme novel views.}
\label{fig:teaser_nvs}

\end{figure}

The recent paradigm shift toward end-to-end intelligent systems based on vision–language–action models~\cite{kim2024openvla,li2025recogdrive,renz2025simlingo} has intensified the need for photo-realistic simulators~\ie world models, which support closed-loop training and evaluation. Despite significant progress in fully diffusion-based simulators~\cite{ali2025world,russell2025gaia,gao2025magicdrive}, these approaches still fall short of the high temporal consistency and cross-view fidelity required for realistic simulation. As an alternative, a parallel line of work leverages 3D Gaussian Splatting (GS)~\cite{liu20243dgs} for sensor simulation~\cite{yang2023unisim,alpasim_2025,zhou2025hugsim, ljungbergh2024neuroncap}. While GS-based methods offer superior temporal coherence, geometric fidelity, and computational efficiency, they tend to degrade when the agent deviates from the trajectories observed in training, a common occurrence in closed-loop settings as illustrated in~\cref{fig:teaser_nvs}, resulting in the challenging novel view synthesis (NVS) problem. In this work, we address this limitation by improving novel view synthesis quality under challenging \textit{autonomous driving} scenarios.

% Existing solutions
Prior efforts to improve novel view synthesis can be broadly grouped into three categories.
\textit{i)} Geometry‑regularized reconstruction methods~\cite{wu2023dngaussian, cheng2024gaussianpro, chen2024pgsr} enforce depth, normal, or structural constraints to promote geometric consistency. While effective to some extent, these methods remain sensitive to noisy supervision, sparse viewpoints, and wide‑baseline extrapolations, often leading to brittle or unstable reconstructions.
\textit{ii)} Optimization methods that inject generative priors, such as RegNeRF~\cite{niemeyer2022regnerf} and VEGS~\cite{veg2023}, can hallucinate plausible content for unseen regions, but often rely on single-view supervision or static priors, which limits multi-view consistency and temporal coherence. \textit{iii)} Most relevant to our setting are diffusion-based enhancement methods~\cite{liu20243dgs, wu2025difix3d}. In particular, DiFix3D~\cite{wu2025difix3d} formulates NVS as an image-to-image diffusion enhancement problem trained on carefully curated pairs of noisy renderings and clean ground-truth images. While these methods can produce visually appealing outputs, they condition primarily on RGB renderings and rely heavily on the diffusion prior to hallucinate corrupted regions. As a result, they lack explicit geometric guidance and struggle to effectively incorporate reference image evidence. When input signal‑to‑noise ratios are low, the enhanced results may appear perceptually compelling but often deviate from the true underlying scene in terms of photometric fidelity.

We address these limitations with \emph{\name}, a geometry-grounded diffusion enhancer designed to process an arbitrary-length, arbitrarily ordered \textit{set} of images rendered from diverse camera poses and time steps. Our model employs an orderless \textit{set-mixer} attention mechanism that enables the diffusion process to attend to and aggregate complementary information across viewpoints, thereby maximizing enhancement quality. To ensure efficiency while capturing cross-frame and cross-view dependencies, we construct the diffusion denoiser using a 2D U-Net architecture distilled into a single-step inference model. This design yields a highly efficient enhancer without sacrificing the ability to leverage rich multi-view geometric context.

Crucially, rather than relying solely on RGB renderings, we introduce a \emph{geometry-grounded} conditioning that exposes the denoiser to explicit 3D scene geometry extracted from the 3DGS representation. Specifically, we condition the diffusion model on rendered coordinate maps that provide pixel-aligned 3D correspondences, along with pose-aware embeddings~\eg Plücker ray encodings. These geometric cues enable the denoiser to reason about where geometry is consistent, uncertain, or conflicting across views, allowing it to suppress floaters, respect occlusions, and recover high-frequency details without compromising structural plausibility. We experimentally demonstrate that such geometry grounding substantially improves content restoration and overall novel view synthesis quality.
Our contributions are summarized as follows:
\begin{itemize}
    \item We introduce a geometry-grounded conditioning strategy that renders 3D geometric cues from Gaussian Splatting representations and integrates them into the diffusion denoiser, improving reconstruction fidelity and reducing hallucinations when RGB renderings are degraded.
    \item We propose a variable-length diffusion architecture that jointly processes a set of reference and target views, enabling effective content propagation across images for stronger multi-view visual aggregation. Owing to its scalable design and training strategy, our model supports real-time enhancement of high-resolution images.
    \item We establish new state-of-the-art results in novel view synthesis on three challenging autonomous driving datasets~\ie EUVS, Para-Lane, nuScenes, as well as on general-scene benchmarks~\ie DL3DV as summarized in~\cref{fig:teaser}.
\end{itemize}
\section{Related Work}

\subsubsection{Neural Rendering and Gaussian Splatting for Autonomous driving}
Neural scene representations—such as Neural Radiance Fields (NeRF)~\cite{mildenhall2020nerf} and in particular 3D Gaussian Splatting (3DGS)~\cite{kerbl2023gaussian}—have recently been extended to autonomous driving scenarios~\cite{chen2025omnire, yan2024street, zanjani2025gaussian, zhou2024drivinggaussian, wu2023mars, zhou2024hugsim}. While these approaches substantially improve photorealism and support reconstruction of large-scale driving scenes, they remain fundamentally limited by the shortcomings of their underlying representations in novel view synthesis. Their performance degrades markedly under sparse or uneven sensor coverage and during wide-baseline viewpoint extrapolation, often resulting in artifacts such as floaters, geometry inconsistencies, and texture bleeding. Such limitations hinder their viability for direct use in generalizable closed-loop simulation pipelines, where stable view synthesis, robustness to sparse observations, and camera-pose extrapolation are essential.

\subsubsection{Generative Priors in 3D Optimization}
To alleviate reconstruction artifacts, several approaches incorporate generative models as priors during optimization. RegNeRF~\cite{niemeyer2022regnerf} and DietNeRF~\cite{jain2021putting} introduce semantic and photometric regularization to guide appearance learning. ReconFusion~\cite{reconfusion2023} and VEGS~\cite{veg2023} leverage pretrained generative models to hallucinate unseen content and improve novel view synthesis. In particular, VEGS distills diffusion-based priors into 3DGS to support urban-scale view extrapolation. While these techniques improve fidelity, they often struggle in severely degraded settings and typically do not ensure strong multi-view or temporal coherence.

\subsubsection{Diffusion-Based Enhancement}
Another line of work formulates novel-view enhancement as an image-level restoration problem using diffusion models. DiFix3D~\cite{wu2025difix3d} applies latent diffusion to denoise rendered views, but its reliance on RGB-only conditioning can cause the model to over-hallucinate, sacrificing fidelity to the underlying scene. DiFix++~\cite{omran2025hybrid} improves restoration quality through refined objectives. 3DGS-Enhancer~\cite{liu20243dgs} frames the task as enhancing intermediate rendered sequence of frames between two high-quality reference views, introducing a video diffusion prior to enforce temporal coherence. However, its single-camera formulation and sequential generation pipeline limit scalability to multi-view or multi-camera simulation settings.

Inspired by prior work, we introduce a set-based diffusion enhancer that jointly restores multi-view renderings using a distilled diffusion prior. In contrast to existing approaches, our framework naturally handles variable numbers of cameras and time steps, and incorporates explicit geometric conditioning through occlusion‑aware cross-view correspondences and known camera poses. This design enables robust enhancement even under severe degradation by transferring missing information from diverse reference views into novel target views, ultimately improving both perceptual quality and structural fidelity.

\section{Geometry Grounded Set Diffusion}

\begin{figure*}[!t]
\centering
 \resizebox{\linewidth}{!}{
  \includegraphics[width=\textwidth, clip]{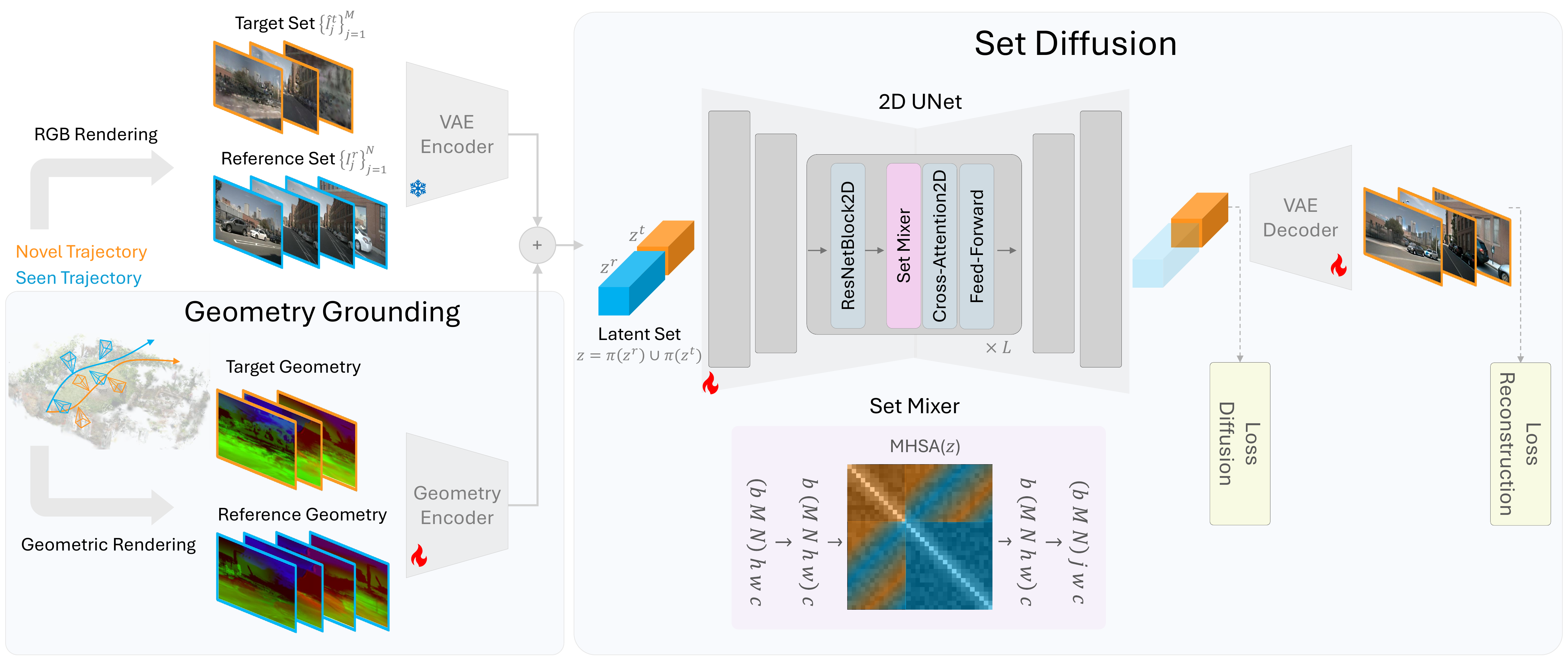}
  }
  
  \caption{
    Overview illustration of \name. The set of rendered novel-view images are enhanced via an image diffusion model, conditioned on reference views, camera poses and 3D geometric priors.
  }
  \vspace{-5pt}
   \label{fig:overview}
\end{figure*}

Given a 3DGS model trained on a set of images from seen camera trajectories, called \textit{reference set}, our goal is to enhance the visual quality when rendering the scene from novel camera poses $\{\p^t_j\}_{j=1}^M$, denoted as the \textit{target set} $\mathcal{I}^t=\{\hat\I^t_j\}_{j=1}^M$.
Due to incomplete observations, the target set often suffers from severe rendering artifacts, especially when the novel camera poses deviates further from the seen trajectories.
Following the recent trend~\cite{liu20243dgs,wu2025difix3d,omran2025hybrid}, we rely on diffusion based image enhancement to remove  novel view rendering artifacts.
We propose two major extensions to prior diffusion enhancers: \emph{i)} We ground the diffusion model on rich 3D scene geometries to overcome the lack of explicit 3D modeling in image diffusion models, as described in~\cref{sec:grounding}. \emph{ii)} Instead of processing the target frames and reference frames one-by-one, we extend the diffusion model to process a variable number of target frames and reference frames jointly as a set, as described in~\cref{sec:setdiff}. This allows a more effective flow of information across target and reference frames leading to higher quality enhancements. An overview of our approach is shown in \cref{fig:overview}.

\subsection{Geometric Grounding}
\label{sec:grounding}
Being trained on web-scale images, the existing image diffusion models lack explicit understanding of 3D structures. The lack of 3D understanding limits their effectiveness in enhancing severe rendering artifacts, leading to unrealistic and inconsistent generations. As a remedy, we propose to extract the rich scene geometries~\ie 3D coordinate maps and camera poses, readily available in 3DGS and encode them into the diffusion process.

\noindent\textbf{Extracting coordinate maps}
In addition to standard RGB rendering—where color values associated with 3D Gaussian primitives are projected onto the image plane—we compute a complementary \emph{geometric rendering} that projects the 3D world coordinates of Gaussian centers into the camera view.

\begin{figure}[!t]
 \resizebox{.85\textwidth}{!}{
\begin{tabular}{c}
%\includegraphics[width=\textwidth]{images/c_maps/fit_3dgs_hybrid_dl3dv_scene=42_rgb.png}
%\\
%\includegraphics[width=\textwidth]{images/c_maps/fit_3dgs_hybrid_dl3dv_scene=42_geo.png}
%\\
\includegraphics[width=\textwidth]{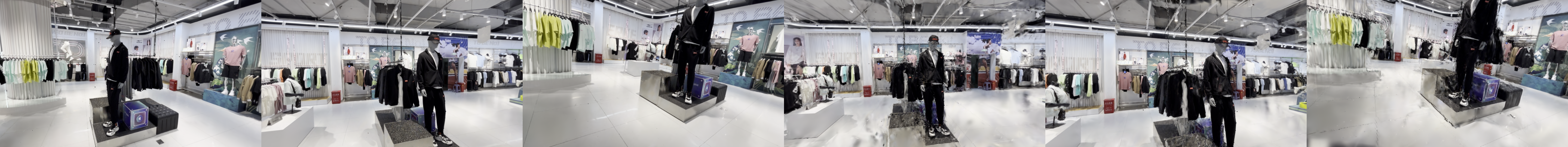}
\\
\includegraphics[width=\textwidth]{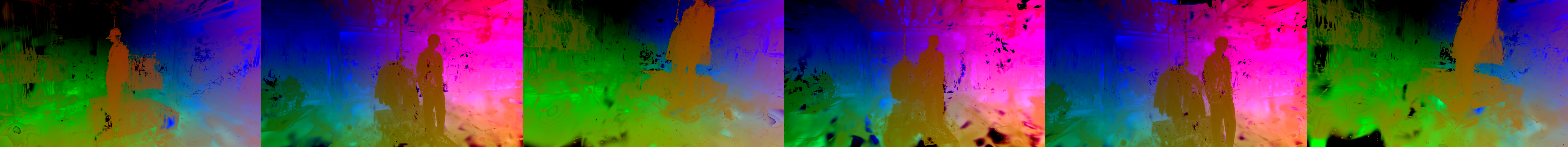}
\\
\includegraphics[width=\textwidth]{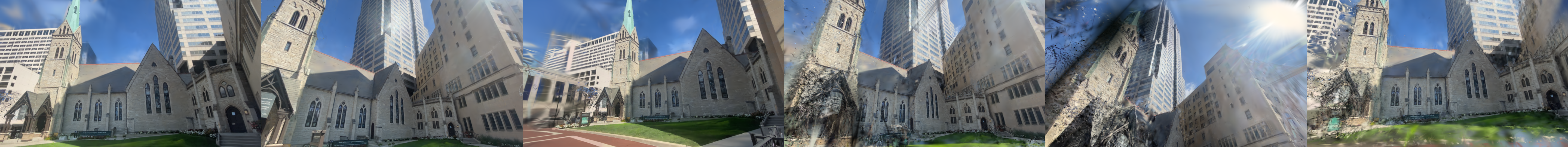}
\\
\includegraphics[width=\textwidth]{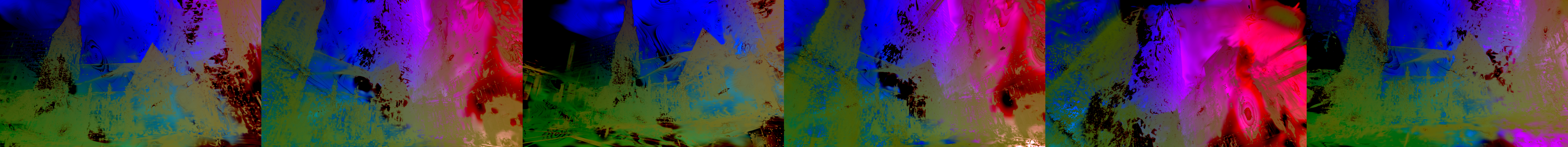}

\end{tabular} 
  }
  \caption{
 Rendered RGB and C-map images from 3DGS for DL3DV scenes. Col. 1–3 depict reference views (\ie training views), while col. 4–6 show target views. The C-maps encode occlusion-aware 3D correspondences across views (represented by colors).
  }
   \label{fig:c_maps_dl3dv}
   \vspace{-6pt}
\end{figure}

Whereas RGB rendering provides appearance information, geometric rendering produces per-pixel 3D coordinate maps (C-maps), which encode the estimated $\textit{xyz}$ position of the visible surface point from the camera’s perspective. Although C-maps for target views may contain noise due to extrapolation or imperfect geometry, they serve as valuable structural cues during diffusion-based restoration. Conditioning the diffusion model on C-maps for both reference and target views provides a 3D-aware guidance in multi-view enhancement task. Analogous to positional encodings, C-maps supply pixel-aligned world coordinates that help guide attention across an unordered set of frames, including heterogeneous cameras and temporal frames. \cref{fig:c_maps_dl3dv} shows rendered RGB images and C-maps from 3DGS for two scenes from DL3DV data.

Formally, a \textit{3D coordinate map} $\mathbf{c}(u,v) \in \mathbb{R}^3$ assigns to each pixel $(u,v)$ the accumulated world coordinate of the surface point along the corresponding camera ray. The coordinate is obtained through an opacity-weighted volumetric accumulation:
\begin{equation}
\mathbf{c}(u,v) = \sum_{i} \alpha_i(u,v) 
\prod_{j=1}^{i-1} \big(1 - \alpha_j(u,v)\big) \; \mathbf{x}_i,
\label{eq:coord_rendering}
\end{equation}
where $(u,v) \in \{1,\dots,H\} \times \{1,\dots,W\}$ denotes pixel coordinates for an image of resolution $H \times W$. Here, $\mathbf{x}_i \in \mathbb{R}^3$ represents the 3D world coordinate of sample $i$ along the ray (\eg, a Gaussian center), and $\alpha_i(u,v)$ is its effective opacity contribution. This formulation parallels standard volumetric rendering, where samples contribute to the final estimate based on visibility and transparency.

In 3DGS reconstruction, $\alpha_i(u,v)$ corresponds to the opacity of the $i$-th Gaussian after projection onto pixel $(u,v)$. An analogous formulation applies to NeRF-style radiance fields when computing coordinate maps. 

\noindent\textbf{Extracting camera poses}
Conditioning the enhancement process on camera poses enables the diffusion model to restore target views by effectively transferring and fusing visual information from high-quality reference views. 
Following prior work~\cite{xu2023dmv3d}, we represent camera geometry using \emph{Plücker ray embeddings}, which encode each pixel's viewing ray as a dense 2D feature map. A ray is parameterized by a pair $(\mathbf{o} \times \mathbf{d},\; \mathbf{d}) \in \mathbb{R}^6$, where $\mathbf{o}$ denotes the ray origin in world coordinates, and $\mathbf{d}$ is the normalized ray direction. The cross product $\mathbf{o} \times \mathbf{d}$ captures the ray's moment, forming together with $\mathbf{d}$ a minimal and rotation-equivariant representation of 3D lines.

To ensure stable conditioning across heterogeneous camera configurations, we apply a normalization procedure to all camera poses before computing Plücker embeddings. This includes normalizing extrinsic coordinate to a canonical reference. Detailed descriptions of the normalization scheme are provided in the supplementary material.

\noindent\textbf{Geometry encoding}
Given the coordinate maps $\mathbf{c}$, Plücker ray embeddings $\mathbf{p}$, we extract the geometric conditioning tensor $\mathbf{g} \in \mathbb{R}^{H/f \times W/f \times c}$ using a convolutional \textit{geometry encoder} as $\mathbf{g} = \Psi(\mathbf{c}, \mathbf{p})$. The encoder is made of five convolutional blocks with group-Norm and SiLU activations. Having the same number of channels and down-sampling factor ($f=8$) as the VAE encoder ensures the dimensionality of the encoders to match, which allows to combine the  photometric and geometric conditioning via element-wise addition. Our ablation studies show that this simple fusion outperforms using cross-attention to consume geometric conditioning.

\subsection{Set Diffusion}
\label{sec:setdiff}
Following previous work~\cite{wu2025difix3d,omran2025hybrid}, we rely on reference-conditioned image diffusion to enhance the noisy targets. Reference images are defined as a set of clean images retrieved from the seen trajectories, which are closest matches to the novel camera poses.

\subsubsection{Reference set construction}
To select the most relevant reference views for the target views, we compute a \textit{view overlap score} between each reference and target camera:
\begin{equation}
    \text{score} = \alpha \cdot \frac{v_i \cdot v_j}{\|v_i\| \cdot \|v_j\|} + \beta \cdot \left( 1 - \frac{d(c_i, c_j)}{\max_{i,j} d(c_i, c_j)} \right),
    \label{eq:score_fun}
\end{equation}
where $v_i$ and $v_j$ denote the view directions of the $i^{\text{th}}$ and $j^{\text{th}}$ cameras, respectively, and $d(c_i, c_j)$ is the Euclidean distance between their camera centers, normalized by the maximum pairwise distance in the scene. The parameters $\alpha$ and $\beta$ control the relative importance of view alignment and spatial proximity. In all experiments, we set $\alpha = 0.8$ and $\beta = 0.2$. It results in a set of $N$ clean reference images per target set, denoted as $\mathcal{I}^r = \{\I^r_j\}_{j=1}^{N}$.

\subsubsection{Latent construction} All images in the reference and target sets, $\mathcal{I}^r$ and $\mathcal{I}^t$, are encoded into the latent space using a frozen image encoder from Stable Diffusion~\cite{sauer2024adversarial}. Geometric conditionings,~\ie C-maps and Plücker ray embeddings, are embedded using a geometry encoder $\Psi$ whose outputs are added to VAE latents. It results in two sets of latents $\z^t \in \mathbb{R}^{M\times H/8 \times W/8 \times c}$ and $\z^r \in\mathbb{R}^{N\times H/8 \times W/8 \times c}$ for target and reference sets, respectively.

\subsubsection{Set mixer}
Unlike previous work~\cite{wu2025difix3d,omran2025hybrid}, which relies on a single image reference and target for diffusion enhancement, we extend the diffusion model to consume a \textit{set} of reference and target images. Using a set of reference images enables the diffusion model to aggregate the information across multiple complementary views rather than solely relying on the closest match, which may miss crucial details needed for an effective enhancement. Moreover, we do enhance the target frames jointly as a set instead of one-by-one which improves the enhancement quality as verified experimentally. 

Instead of relying on \textit{vector} representation, which fixes the number and the position of reference and target frames, we rely on a flexible set representation that allows to use an arbitrary number and an arbitrary order of target and reference images. This flexibility allows a single trained model to handle a variable number of target and their matched reference frames and generalize well to the arbitrary ordered reference and target views. 
For this purpose, we rely on an image-based denoiser with no spatio-temporal convolutions or position encoding in the temporal domain. In particular, we use the 2D UNet from Stable Diffusion which meets these requirements. Additionally, during the training, we apply a random temporal permutation function $\pi$ on the target and reference frames, $\pi(\z^t)$ and $\pi(\z^r)$, to increase the model robustness against temporal reordering.

To effectively aggregate visual information across the reference and target sets, we replace each self-attention layer in the UNet with a \textit{Set Mixer}. It unions all the tokens from target and reference set, \ie reshaping into $b(MNhw)c$ where $b$ denotes batch size, and performs multi-head self-attention (MHSA) over the unified token set as illustrated in \cref{fig:overview} . It allows each token from the target or reference images to attend to all other tokens to globally search for the missing details in any reference or target views.
As validated experimentally, our set diffusion imposes a minimal computational overhead to the frame-by-frame processing allowing the batch processing of frames in real time. 

\subsubsection{Diffusion training} 
We train our diffusion enhancer following the standard single step image-to-image diffusion training~\cite{parmar2024one, wu2025difix3d, omran2025hybrid} based on SD-Turbo~\cite{sauer2024adversarial}. Specifically, we finetune the denoising UNet $f_\theta$, to take the noisy target and clean reference latents, $\z^t$ and $\z^r$, as input and predict the clean target views $\y^t$ as the ground-truth. Following~\cite{wu2025difix3d}, we fix the noise level $\tau=200$, and adopt the v-prediction objective~\cite{salimans2022progressive} with the following latent diffusion loss:
\begin{equation}
    \mathcal{L}_{\text{Diffusion}}=\mathbb{E}_{\z^r, \z^t, \y^t} 
    \left[\left\| 
    \mathbf{y}^t - f_{\theta}\!\left([z^t; z^r], \tau\right) 
    \right\|_2^2\right]
\end{equation}
We augment the latent diffusion objective with a pixel-space reconstruction loss, $\mathcal{L}_{\text{Reconstruction}}$, enabling joint fine-tuning of the VAE decoder together with the geometry encoder and the denoising U-Net. The reconstruction loss is defined as an equally weighted combination of $\ell_2$, SSIM, and LPIPS terms. This formulation jointly enforces pixel-level accuracy, structural consistency, and perceptual fidelity, leading to improved enhancement quality. Given the high memory requirements involved in optimizing the pixel-space reconstruction loss, during training, we only decode a subset of target latents and rely on gradient accumulation and LoRA fine-tuning of the VAE decoder.
\section{Experiments}
\begin{table*}[b!]
\caption{\textbf{SOTA comparisons on EUVS} across three extrapolation settings. Neural reconstruction (top) and diffusion enhancers (bottom) comparisons are reported separately. All diffusion enhancers are applied on the same 3DGS backbone.}
  \label{tab:euvs_table1}  
  \centering
  % \begin{adjustbox}{width=\linewidth}
\setlength{\tabcolsep}{8pt} % default is 6pt
\renewcommand{\arraystretch}{1.1} % row spacing (default = 1.0)

\resizebox{\textwidth}{!}{
\begin{tabular}{lccc|ccc|ccc}
\toprule
\multirow{3}{*}{\textbf{Method}} & \multicolumn{9}{c}{\textbf{EUVS Novel-View Data Settings}} \\
\cmidrule(lr){2-10}
& \multicolumn{3}{c}{\textbf{Setting 1}} & \multicolumn{3}{c}{\textbf{Setting 2}} & \multicolumn{3}{c}{\textbf{Setting 3}} \\
\cmidrule(lr){2-4} \cmidrule(lr){5-7} \cmidrule(lr){8-10} 
& \textbf{PSNR} & \textbf{SSIM} & \textbf{LPIPS} & \textbf{PSNR} & \textbf{SSIM} & \textbf{LPIPS} & \textbf{PSNR} & \textbf{SSIM} & \textbf{LPIPS} \\
\midrule
3DGS~\cite{kerbl2023gaussian} & 16.37 & 0.720 & 0.260 & 19.53 & 0.751 &  0.267 & 14.99 & 0.717 &  0.405 \\
2DGS~\cite{2dgs2023} & 16.30 & 0.710 & 0.289 & 18.83 &  0.720 &  0.292 &  11.36 & 0.545 & 0.546 \\
GSPro~\cite{cheng2024gaussianpro} & 16.39 & 0.719 & 0.245 & 19.39 & 0.747 & 0.225 & 14.82 & 0.699 & 0.388 \\
PGSR~\cite{chen2024pgsr} & 16.32 & 0.710 & 0.273 & 18.38 & 0.712 & 0.253 &  14.25 &  0.698 &  0.436 \\
3DGM~\cite{3DGM2024} & 16.35 & 0.725 & 0.254 & 18.78 & 0.746 &  0.281 & 14.60 & 0.723 & 0.405\\
Feature 3DGS~\cite{zhou2024feature} & 16.01 & 0.724 & 0.257 & 19.59 &  0.786 & 0.228 & 14.33 & 0.639 & 0.382 \\
Zip-NeRF~\cite{zipnerf2023} & 14.06 & 0.692 & 0.342 & 17.36 & 0.672 &  0.358 & 14.42 & 0.656 & 0.454 \\
Instant-NGP~\cite{instantngp2022} & 12.65 & 0.625 & 0.594 & 17.15 & 0.721 & 0.517 & 14.39 &  0.710 & 0.659 \\
\midrule
3DGS~\cite{kerbl2023gaussian} & 16.37 & 0.720 & 0.260 & 19.53 & 0.751 &  0.267 & 14.99 & 0.717 &  0.405 \\
\quad + VEGS~\cite{veg2023} & 15.88 & 0.705 & 0.306 & \textbf{23.33} & 0.795 &  0.281 &  14.25 & 0.647 & 0.442 \\
\quad + DIFIX3D~\cite{wu2025difix3d} & 17.09 & 0.739 & 0.224 & 20.02 & 0.713 & 0.187 & 14.84 & 0.665 & 0.328 \\
\quad + DIFIX3D++~\cite{omran2025hybrid} & \textbf{17.94} & 0.782 & 0.218 & 21.68 & 0.772 & 0.209 & 15.91 & 0.743 & 0.335 \\
\rowcolor{lightblue} \quad + \textbf{\name} & 17.84 & \textbf{0.796} & \textbf{0.192} & 22.55 & \textbf{0.796} & \textbf{0.172} & \textbf{16.60} & \textbf{0.745} & \textbf{0.291} \\
\bottomrule
\end{tabular}
}
\end{table*}

\subsection{Experimental Setup}
\subsubsection{Dataset}
To evaluate the generalization ability of novel view synthesis (NVS) methods under challenging extrapolation, sparsity, and dynamic scene conditions, we conduct experiments across four diverse datasets: 
\emph{i)} EUVS~\cite{han2025extrapolated} as a large-scale benchmark designed to evaluate NVS under significant viewpoint changes—crucial for vision-centric autonomous driving.
\emph{ii)} Para-Lane~\cite{ni2025lane} as multi-lane urban driving sequences captured from parallel trajectories, enabling evaluation of NVS under realistic cross-lane displacements.
\emph{iii)} nuScenes~\cite{caesar2020nuscenes} to evaluate the model performance under significant scene dynamics, including moving vehicles, pedestrians, and riders.
\emph{iv)} DL3DV-10K~\cite{ling2024dl3dv} to evaluate the model on diverse bounded and unbounded scenes, beyond the autonomous driving domain. The details and experimental protocol for each dataset is provided in~\cref{sec:sup_dataset}.

\noindent\textbf{Evaluation metrics}
We evaluate the quality of synthesized novel views using PSNR, SSIM, and LPIPS. These metrics quantify complementary aspects of reconstruction quality: PSNR captures pixel-wise fidelity, SSIM measures structural similarity, and LPIPS assesses perceptual consistency. All metrics are computed against ground-truth novel-view images provided by each dataset.

\noindent\textbf{Implementation details}
During the training, we use target and reference sets with variable size ranging from $8$ to $12$ frames in total. At inference time, we limit the set size based on the available GPU memory. Using larger reference set generally improves enhancement quality by supplying richer context, while increasing the target set size enables joint processing and stronger cross-view message passing. 

C-maps are generated via global-frame 3DGS rasterization (\Cref{eq:coord_rendering}), clipped by scene scale, with zero assigned to rays that hit no primitives. No filtering, inpainting, or additional uncertainty modeling is applied beyond the 3DGS opacity and Gaussian parameters. Training includes classifier-free guidance (CFG). We apply a dropout rate of 15\% to both the C-map and camera pose. At inference time, the CFG scale is set to 2.0. All experiments are conducted at a resolution of 576 $\times$ 1024 pixels. Additional implementation choices are provided in the supplementary material.

\begin{figure*}[t!]
\centering
\resizebox{\textwidth}{!}{%
  \begin{tabular}{@{}c@{\hspace{-3pt}}c@{}} % 3pt gap; use @{} to remove default padding
    \begin{tabular}{@{}c@{\hspace{2pt}}c@{}c@{}c@{}c@{}}
%% Example 1
\rotatebox{90}{\makebox[0.8cm][r]{\tiny{3DGS}}}&
\includegraphics[width=.16\textwidth]{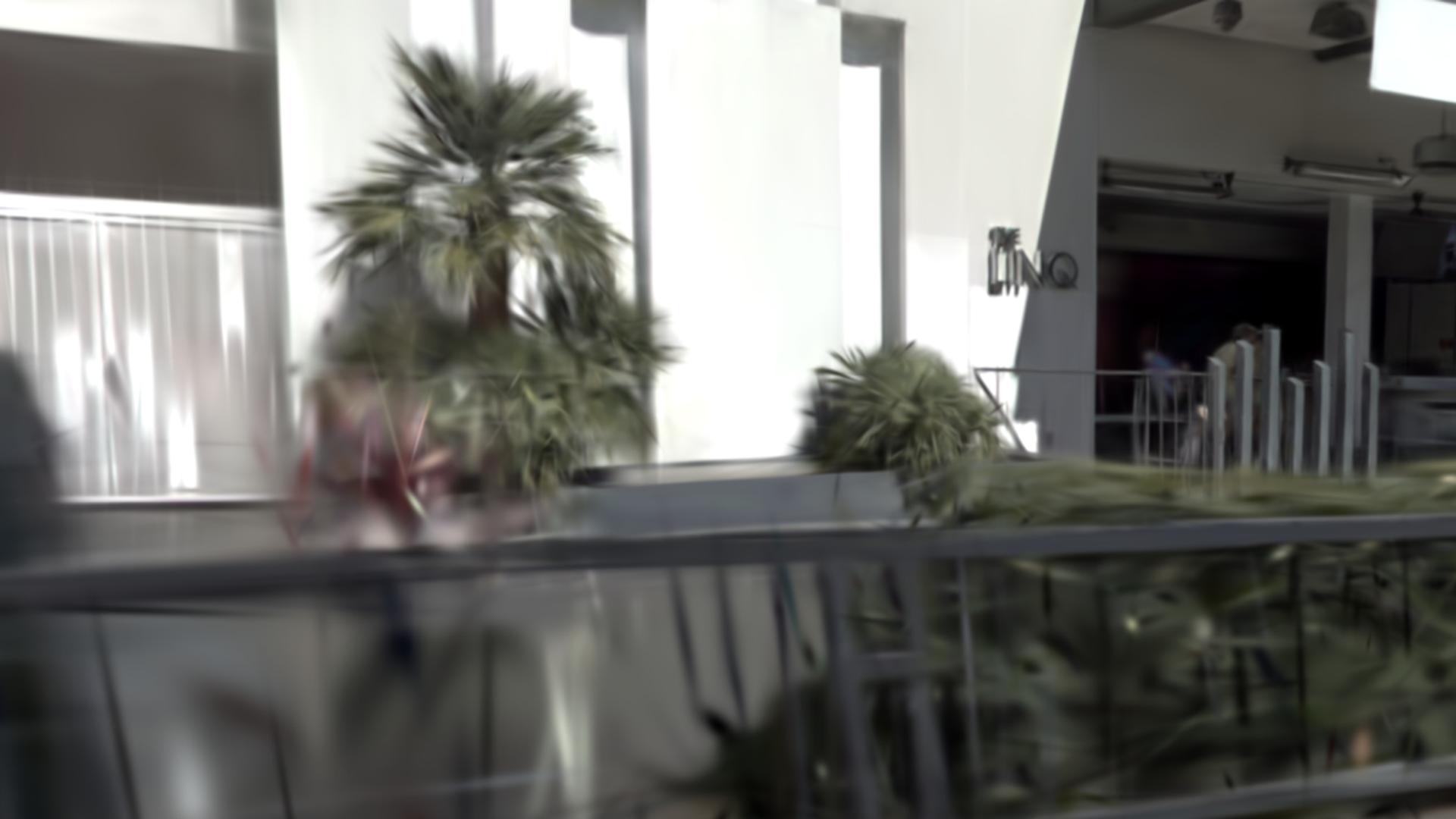} &
\includegraphics[width=.16\textwidth]{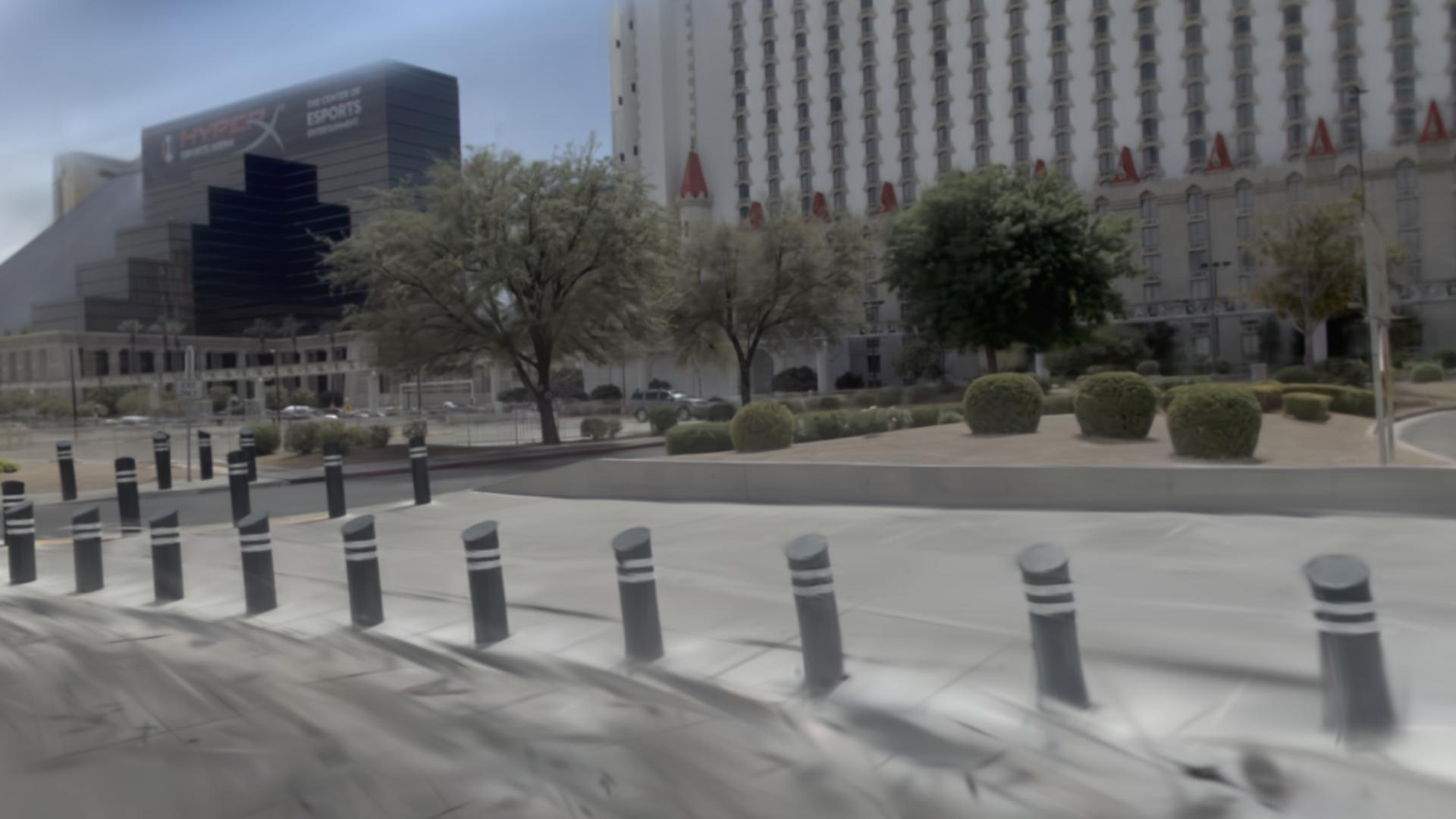} &
\includegraphics[width=.16\textwidth, trim={0cm 0cm 0cm 3.4cm}, clip]{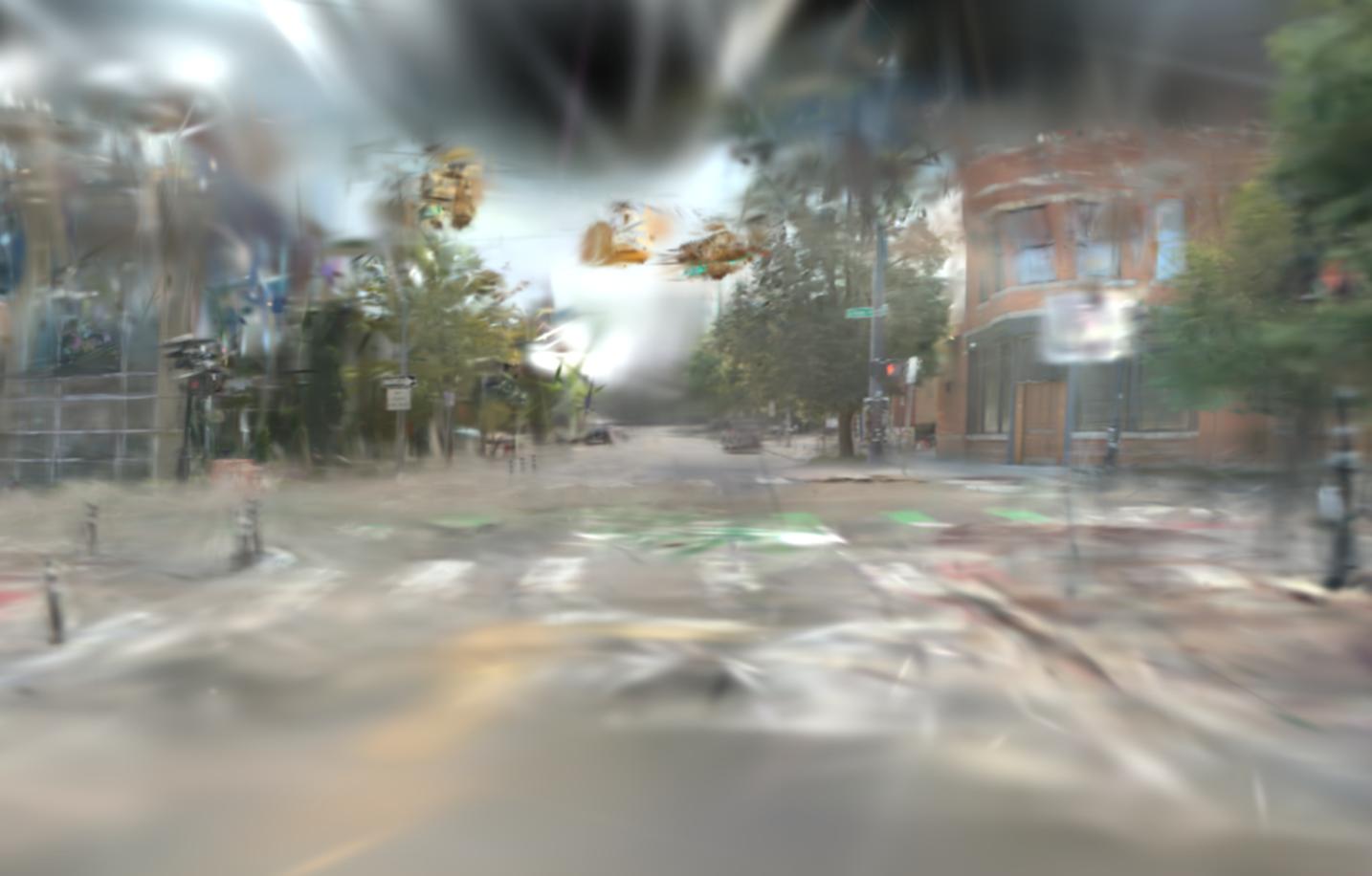} &
\includegraphics[width=.16\textwidth, trim={0cm 0cm 0cm 3.4cm}, clip]{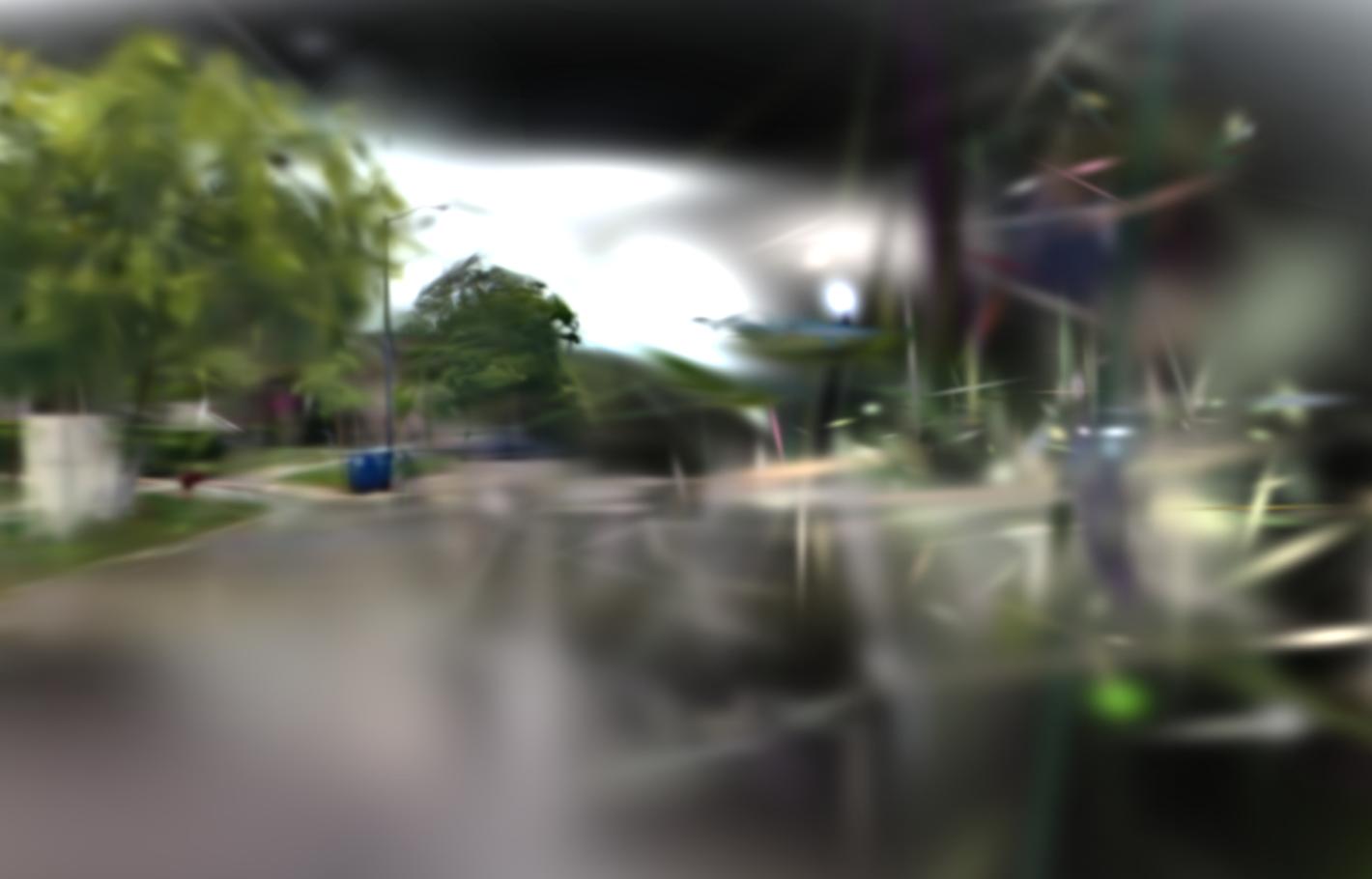}
\vspace{-3pt}\\
\rotatebox{90}{\makebox[1.0cm][r]{\tiny{DiFix3D}}}&
\includegraphics[width=.16\textwidth]{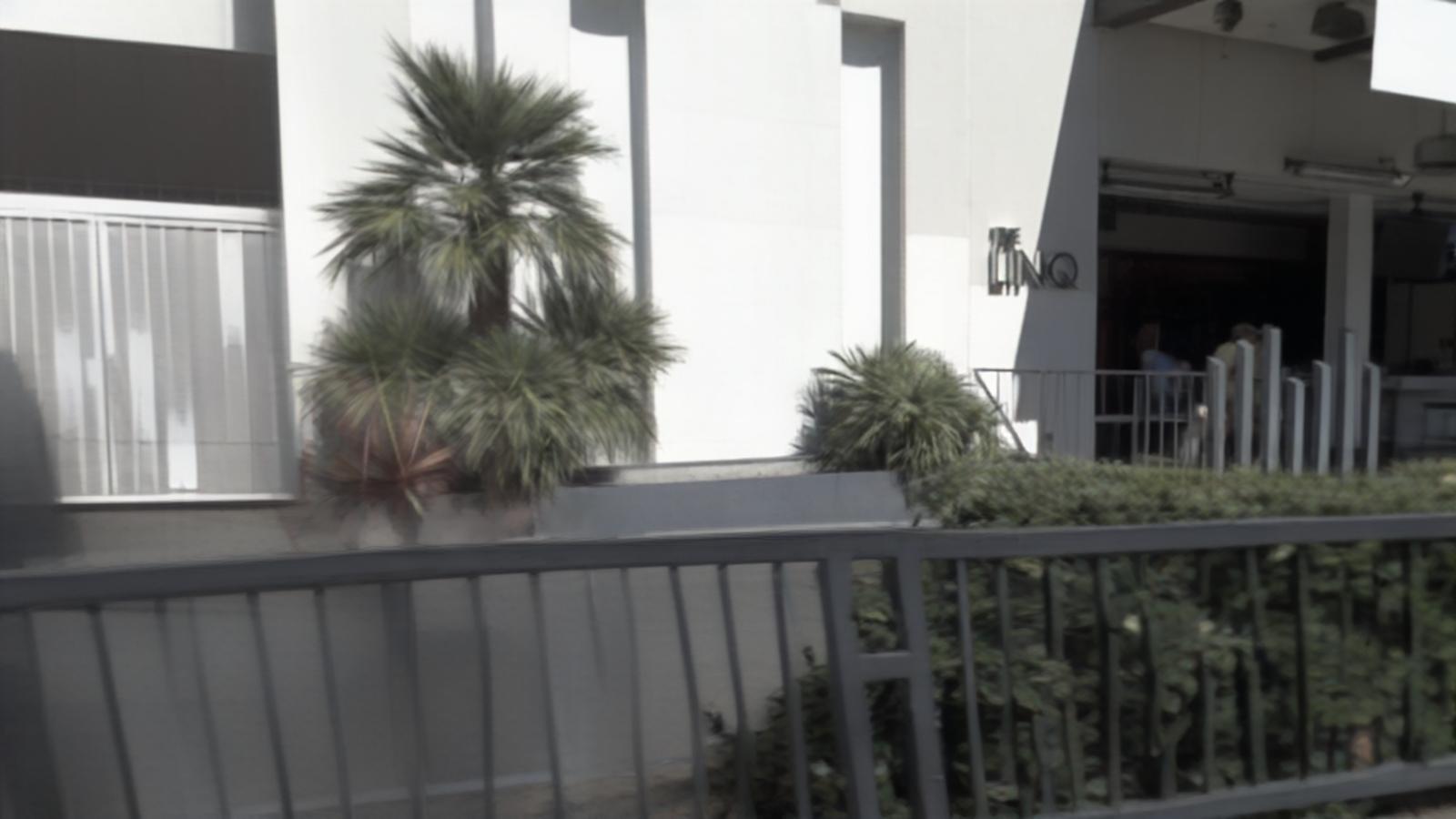} &
\includegraphics[width=.16\textwidth]{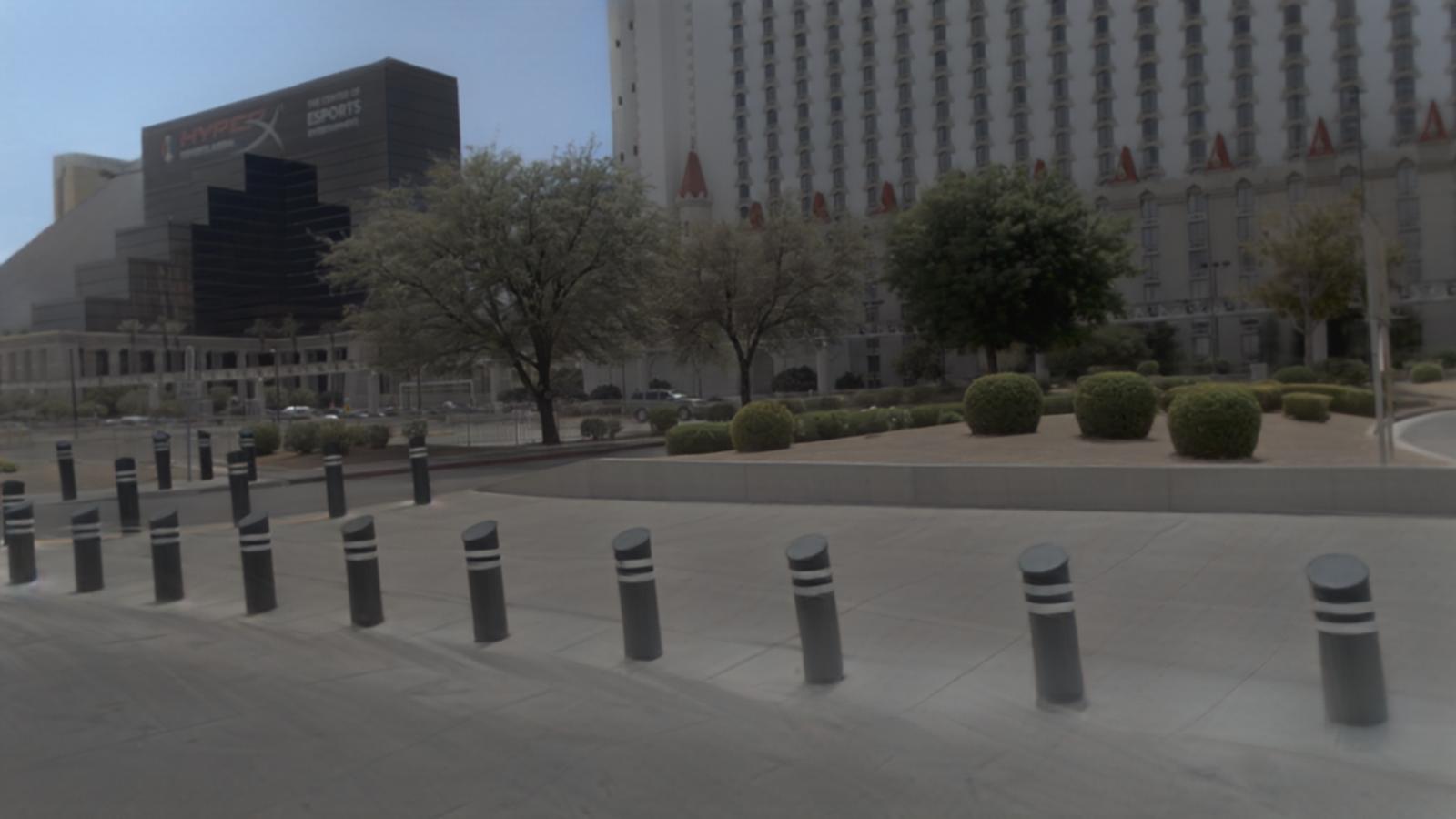} &
\includegraphics[width=.16\textwidth, trim={0cm 0cm 0cm 3.4cm}, clip]{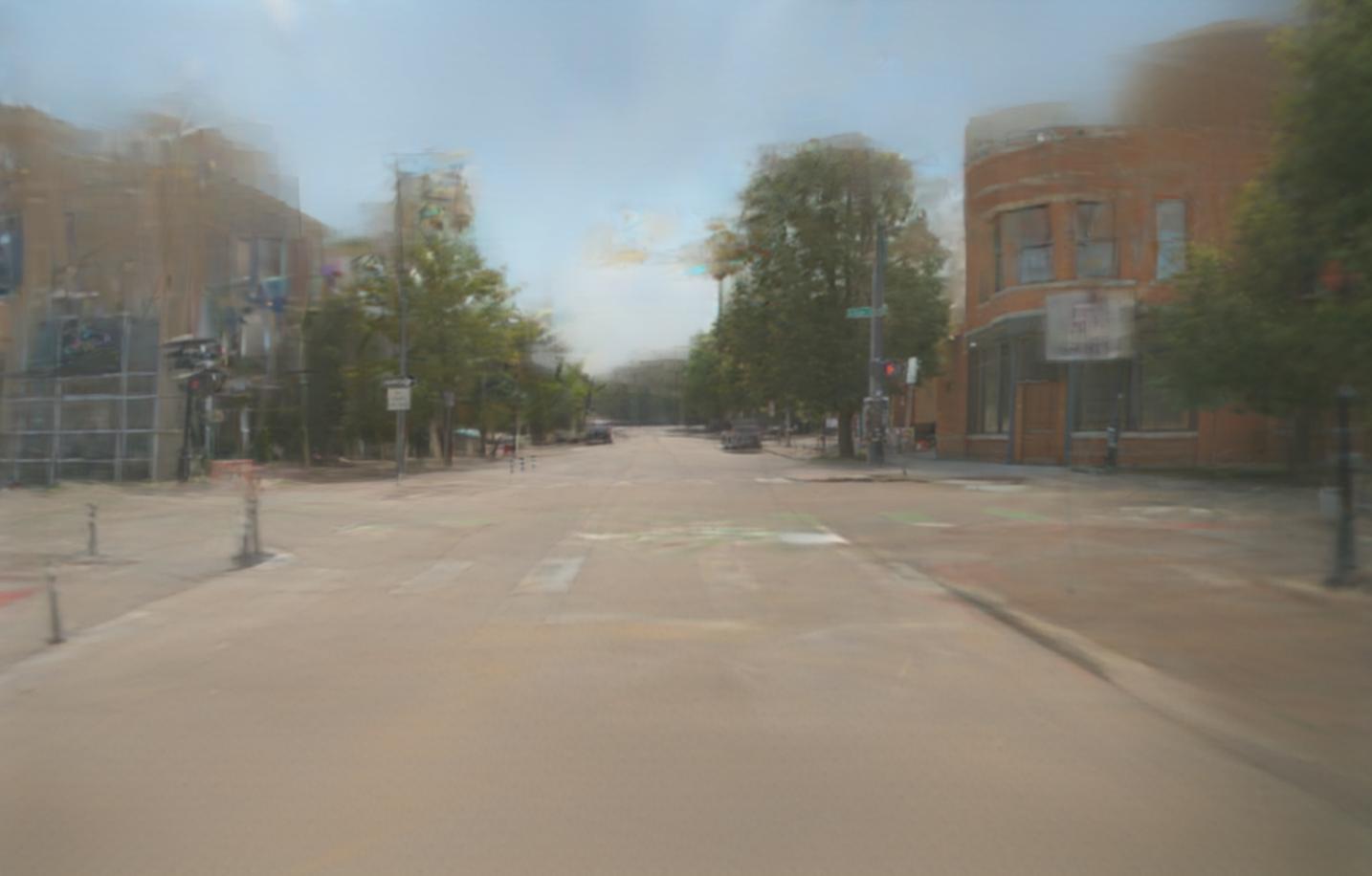} &
\includegraphics[width=.16\textwidth, trim={0cm 0cm 0cm 3.4cm}, clip]{ 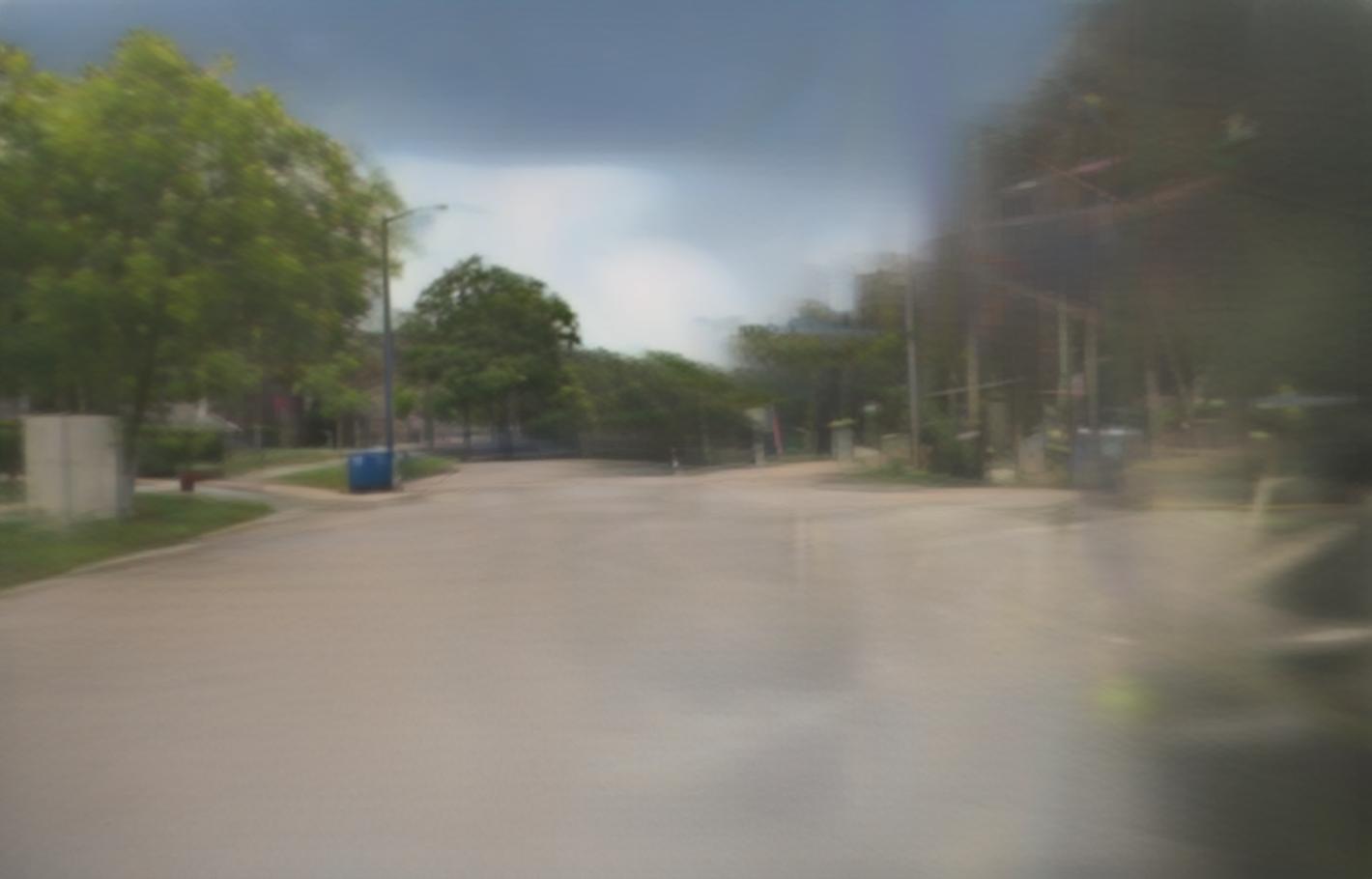}
\vspace{-3pt}\\
\rotatebox{90}{\makebox[0.8cm][r]{\tiny{Ours}}}&
\includegraphics[width=.16\textwidth]{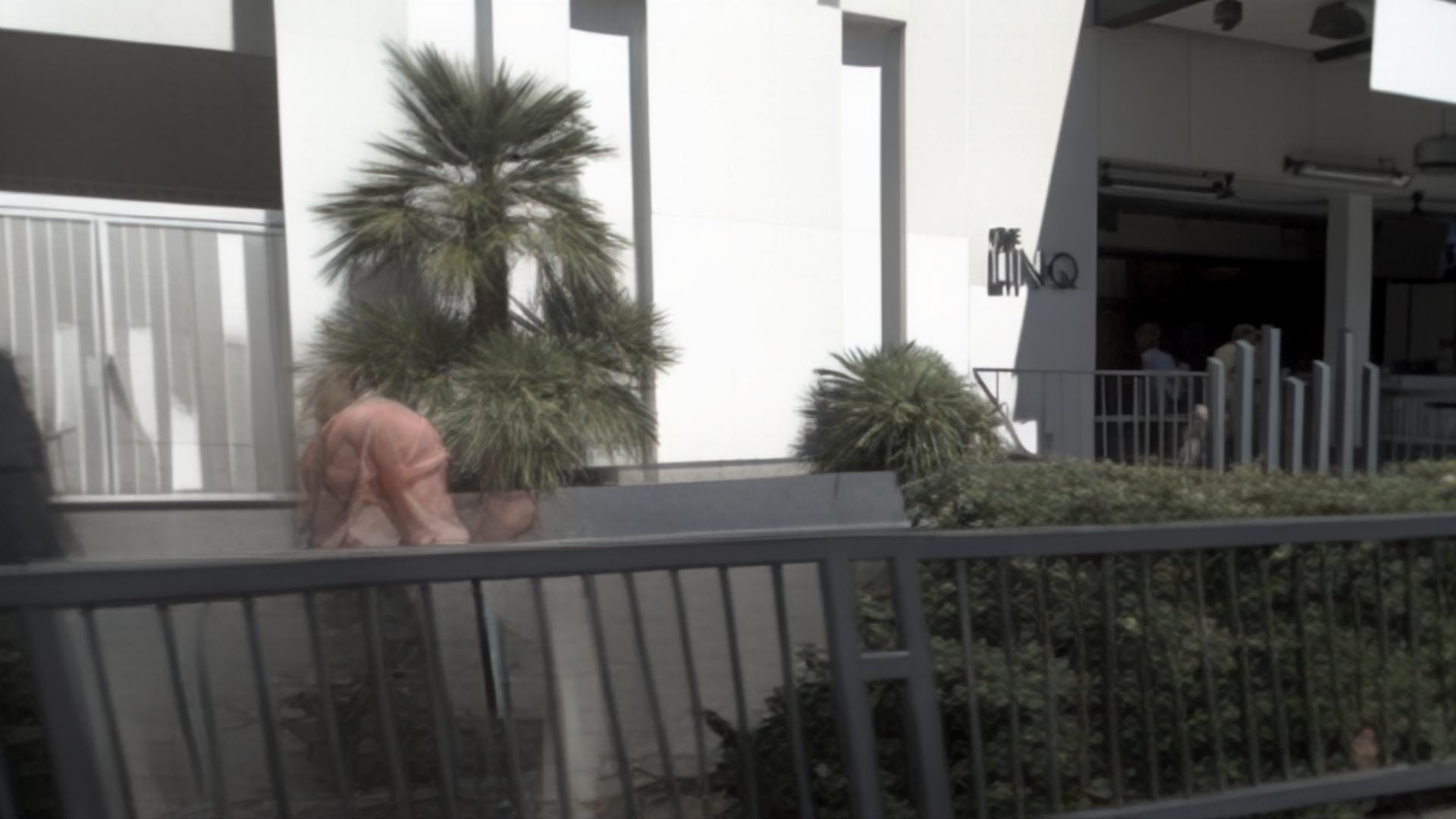} &
\includegraphics[width=.16\textwidth]{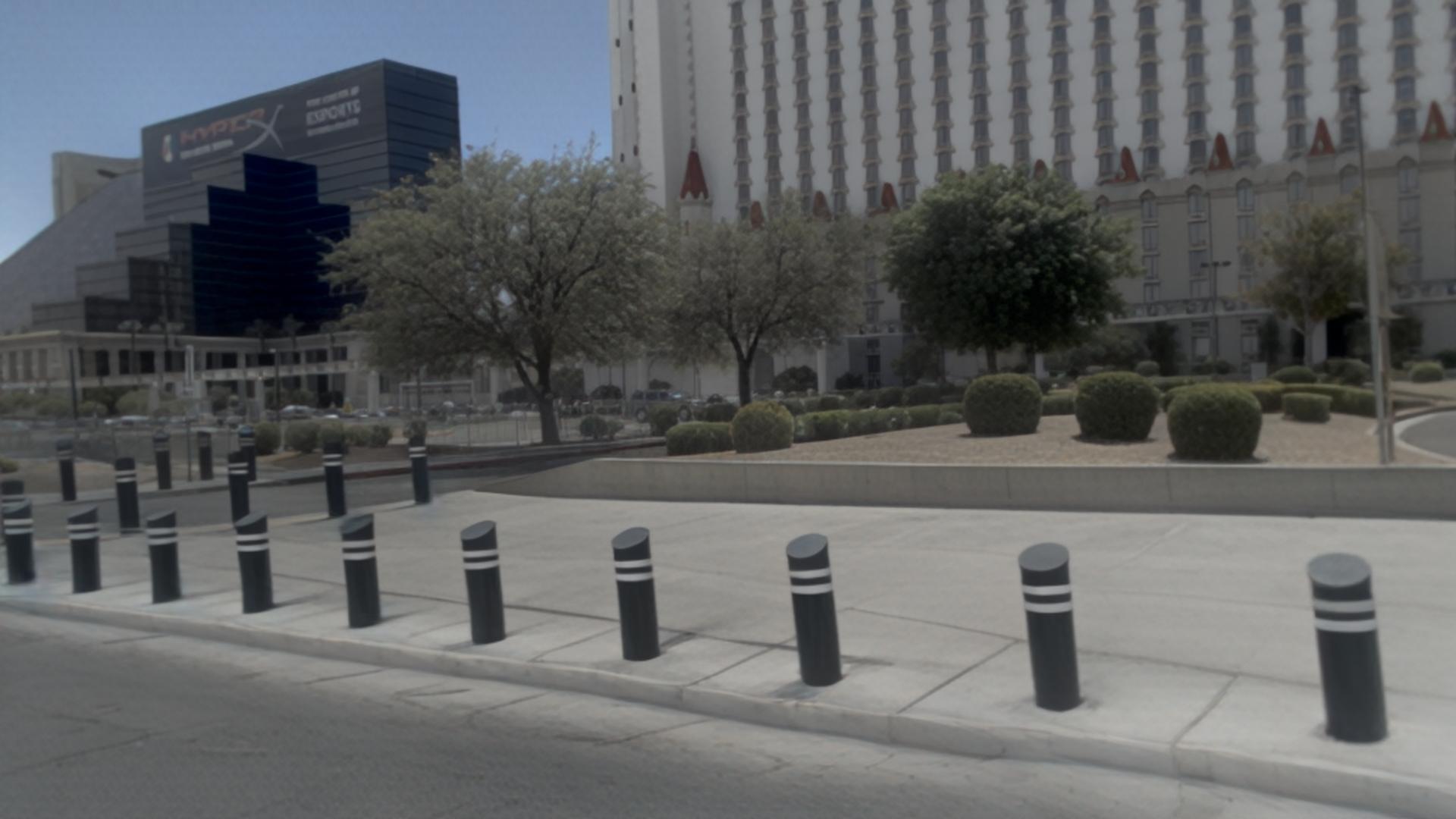} &
\includegraphics[width=.16\textwidth, trim={0cm 0cm 0cm 3.4cm}, clip]{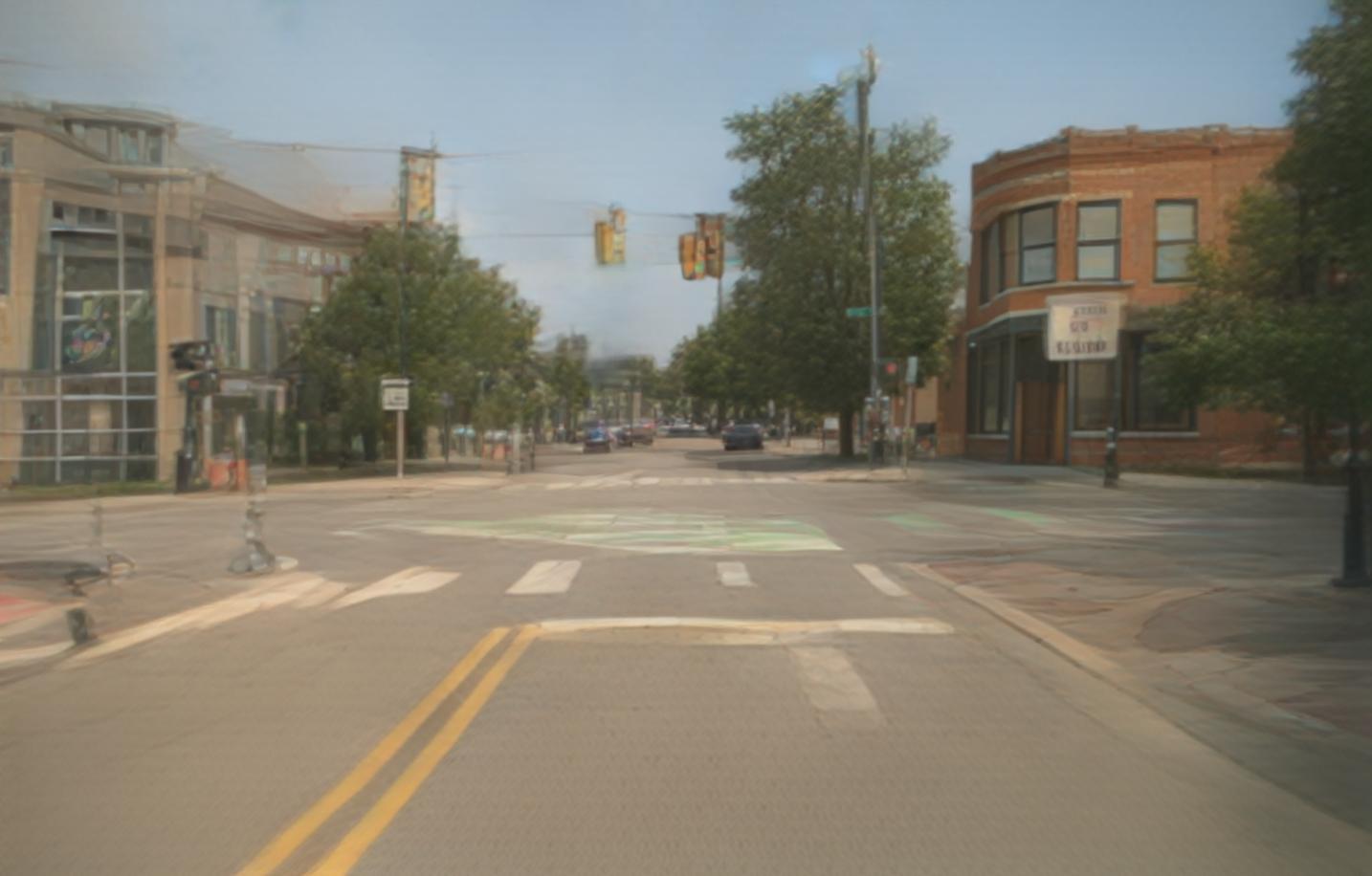} &
\includegraphics[width=.16\textwidth, trim={0cm 0cm 0cm 3.4cm}, clip]{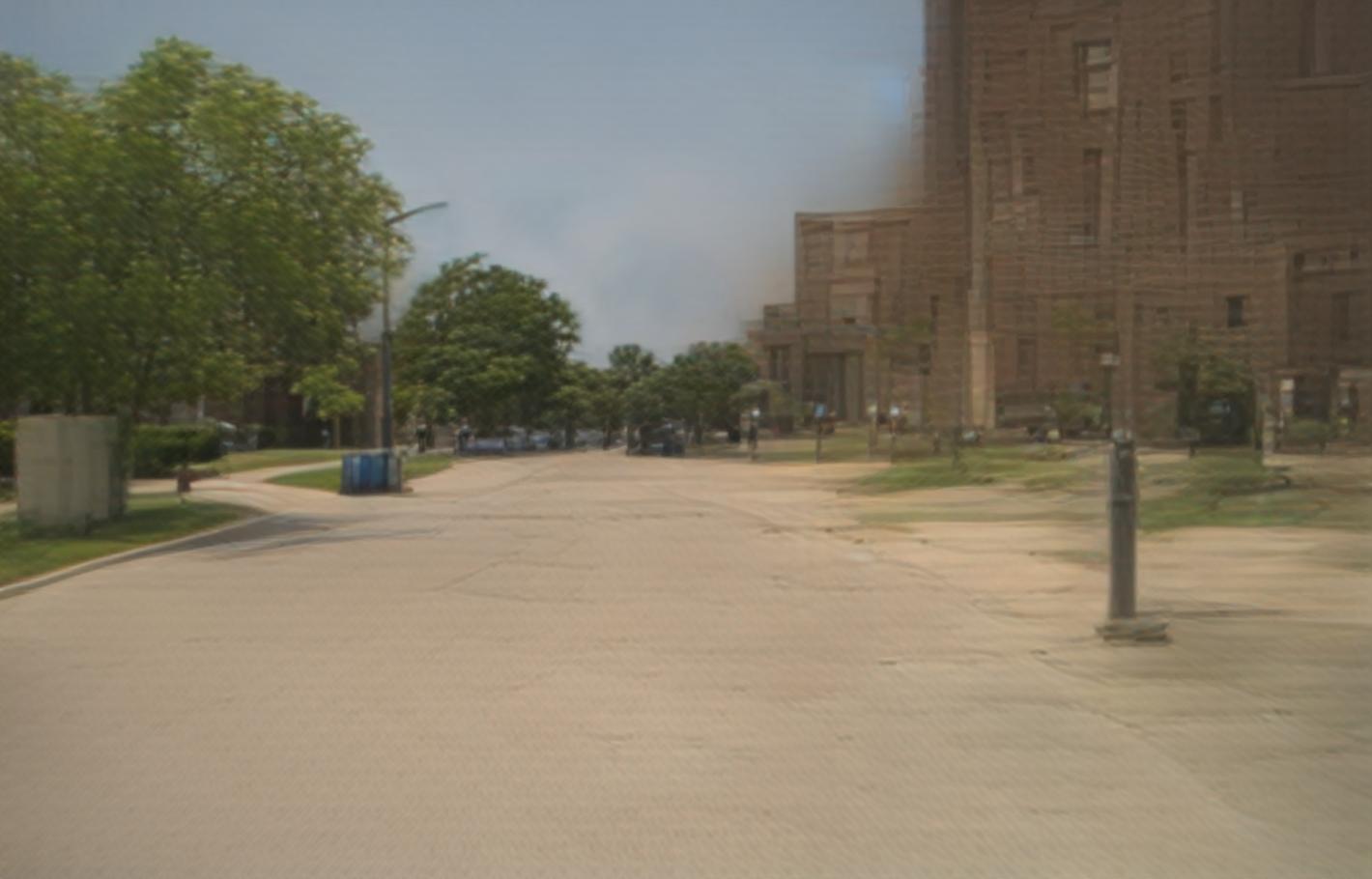}
\vspace{-3pt}\\
\rotatebox{90}{\makebox[0.8cm][r]{\tiny{GT}}}&
\includegraphics[width=.16\textwidth]{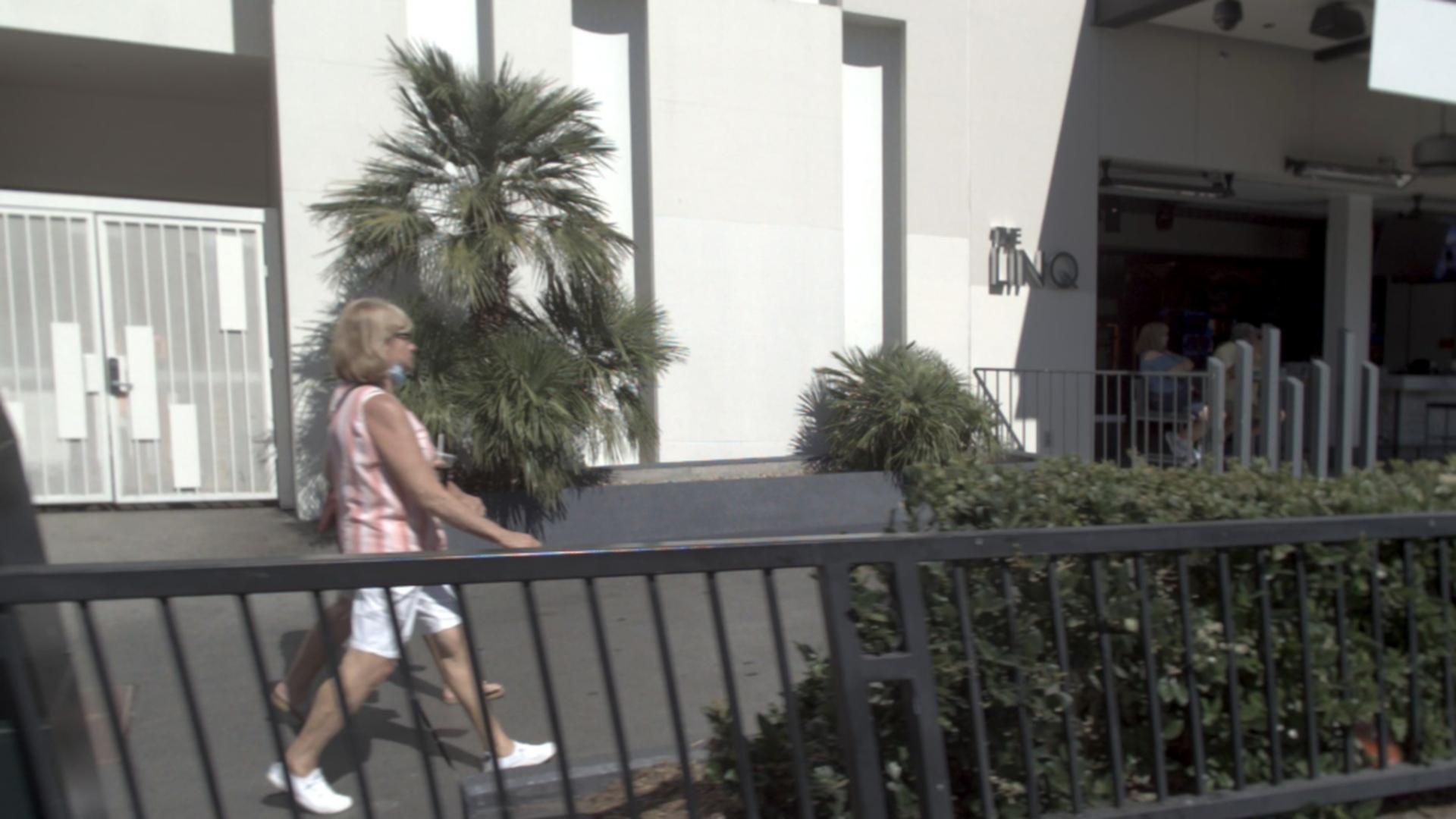}&
\includegraphics[width=.16\textwidth]{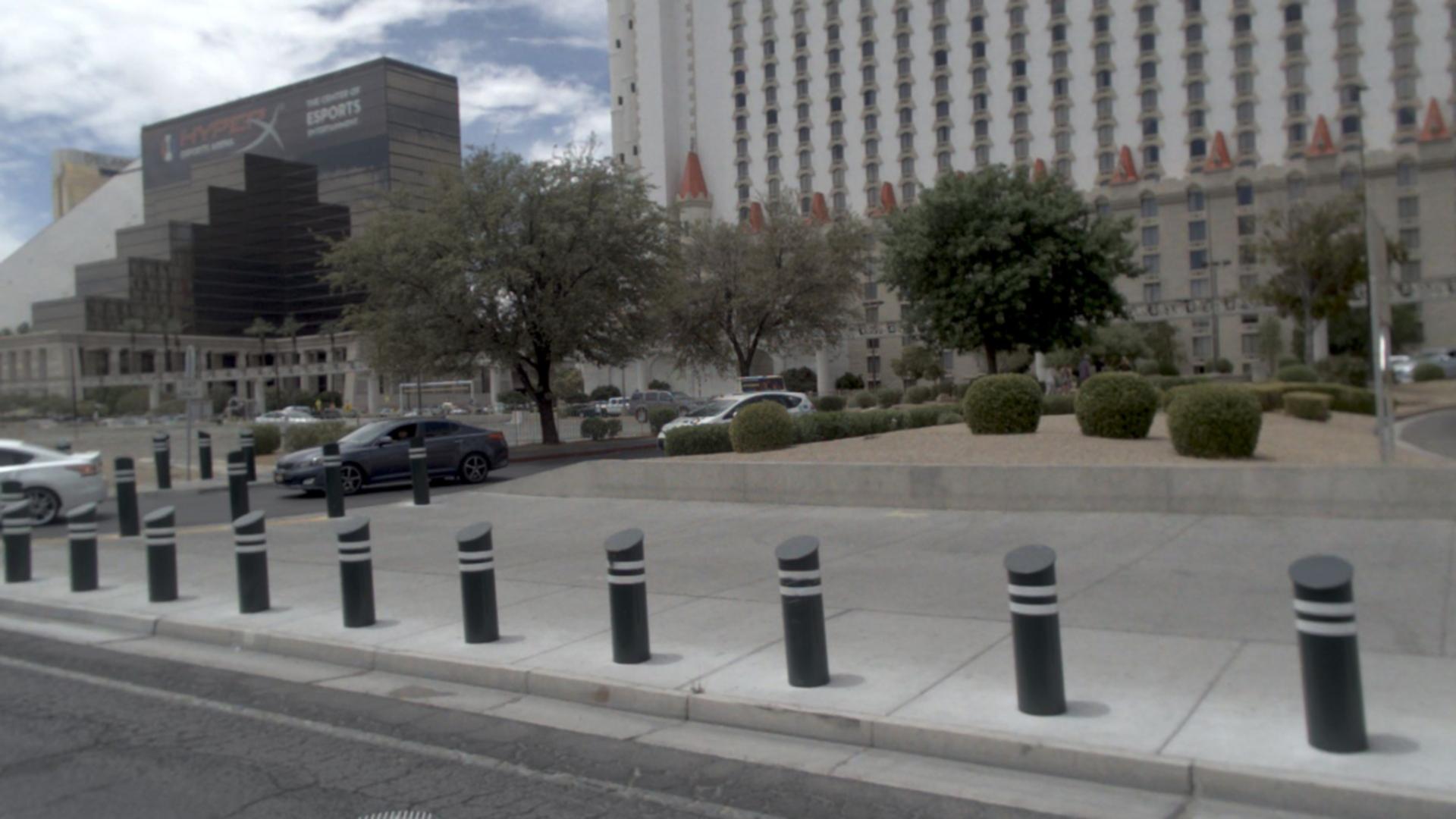}&
\includegraphics[width=.16\textwidth, trim={0cm 0cm 0cm 3.4cm}, clip]{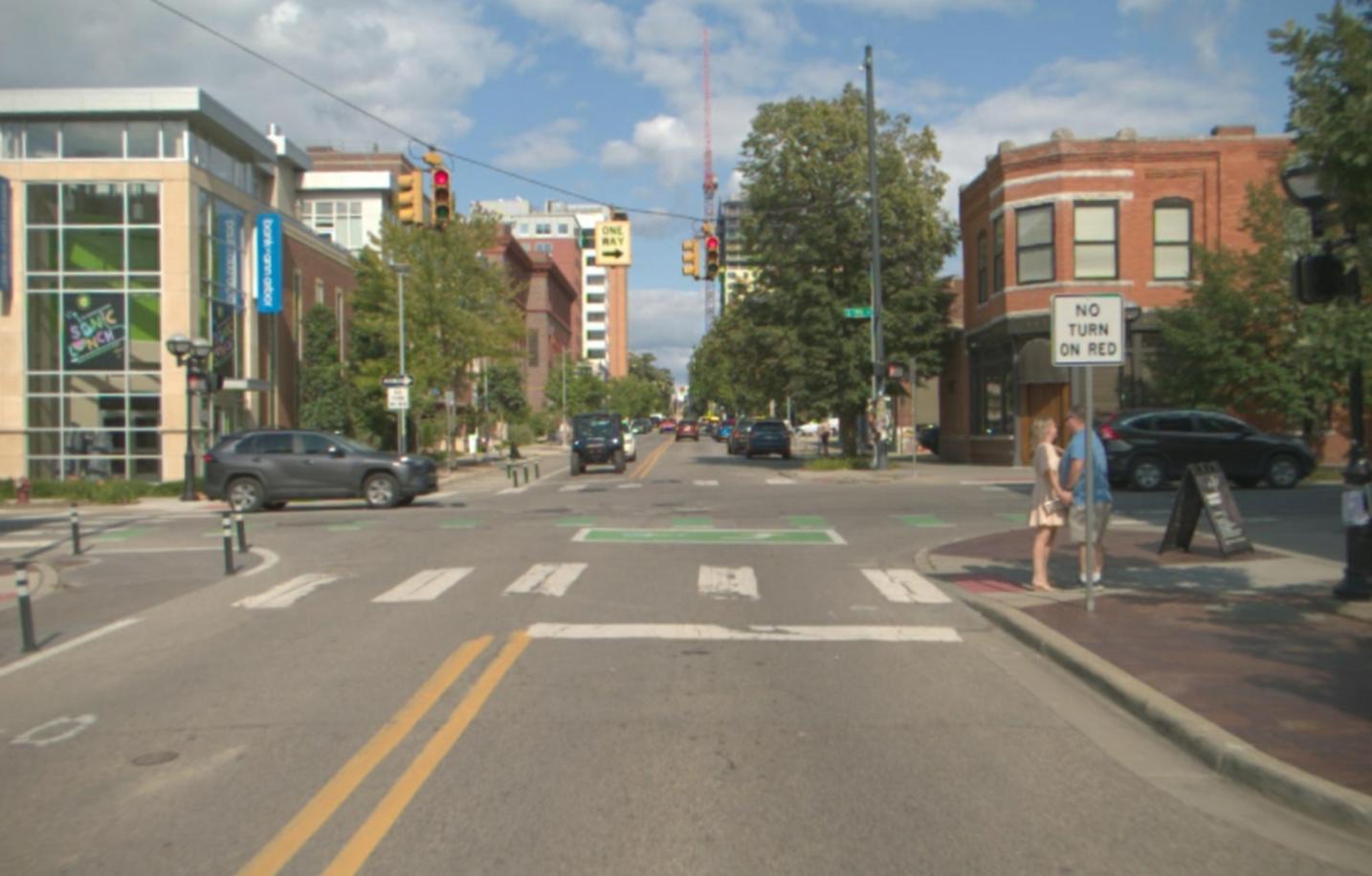}&\hspace{-6pt}
\includegraphics[width=.16\textwidth, trim={0cm 0cm 0cm 3.4cm}, clip]{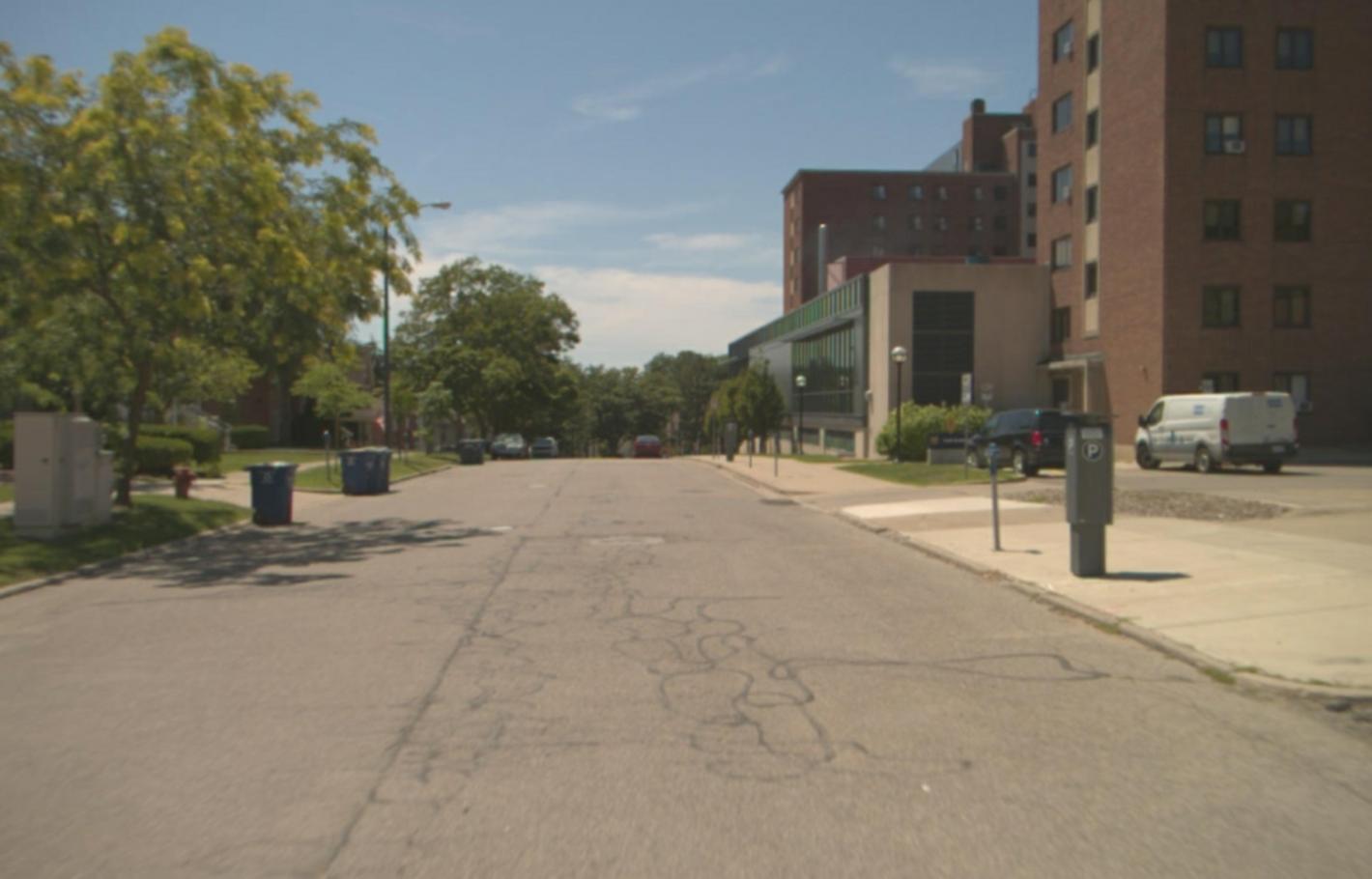}

\\
\end{tabular} &
    \begin{tabular}{@{}c@{}c@{}}
%% Example 1
\includegraphics[height=0.092\textwidth]{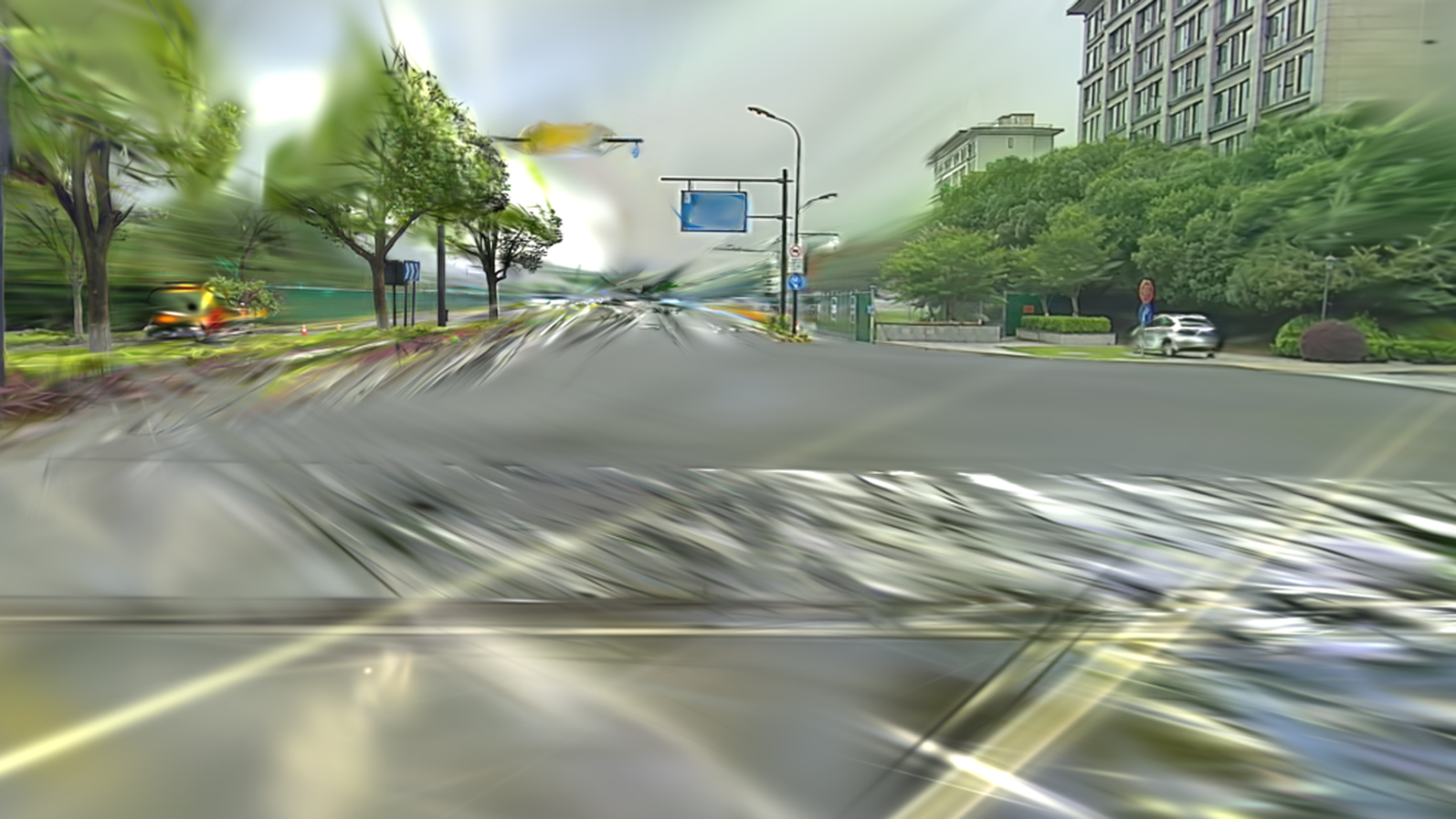}&
\includegraphics[height=0.092\textwidth]{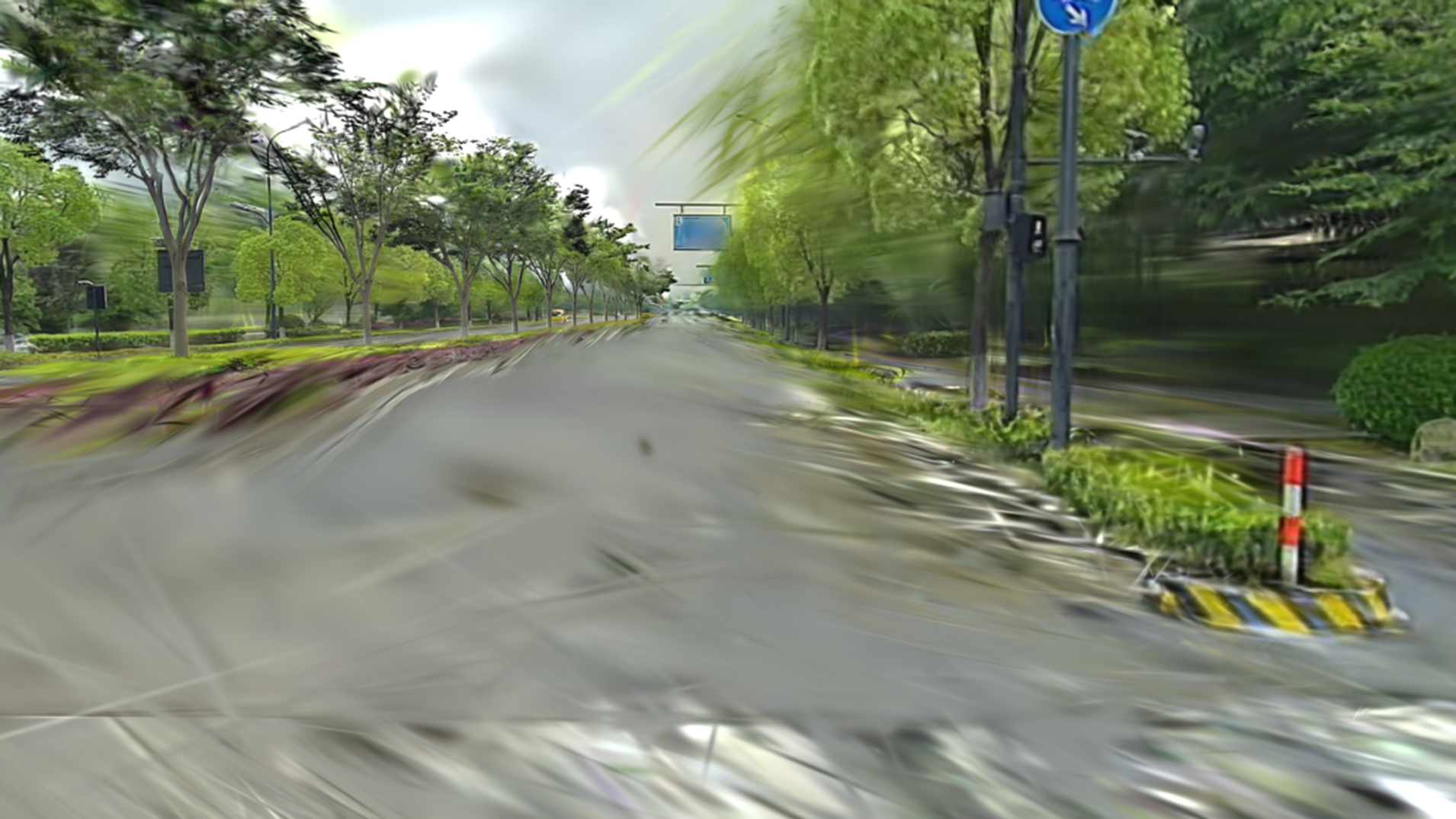} 
\vspace{-3pt}\\
\includegraphics[height=0.092\textwidth]{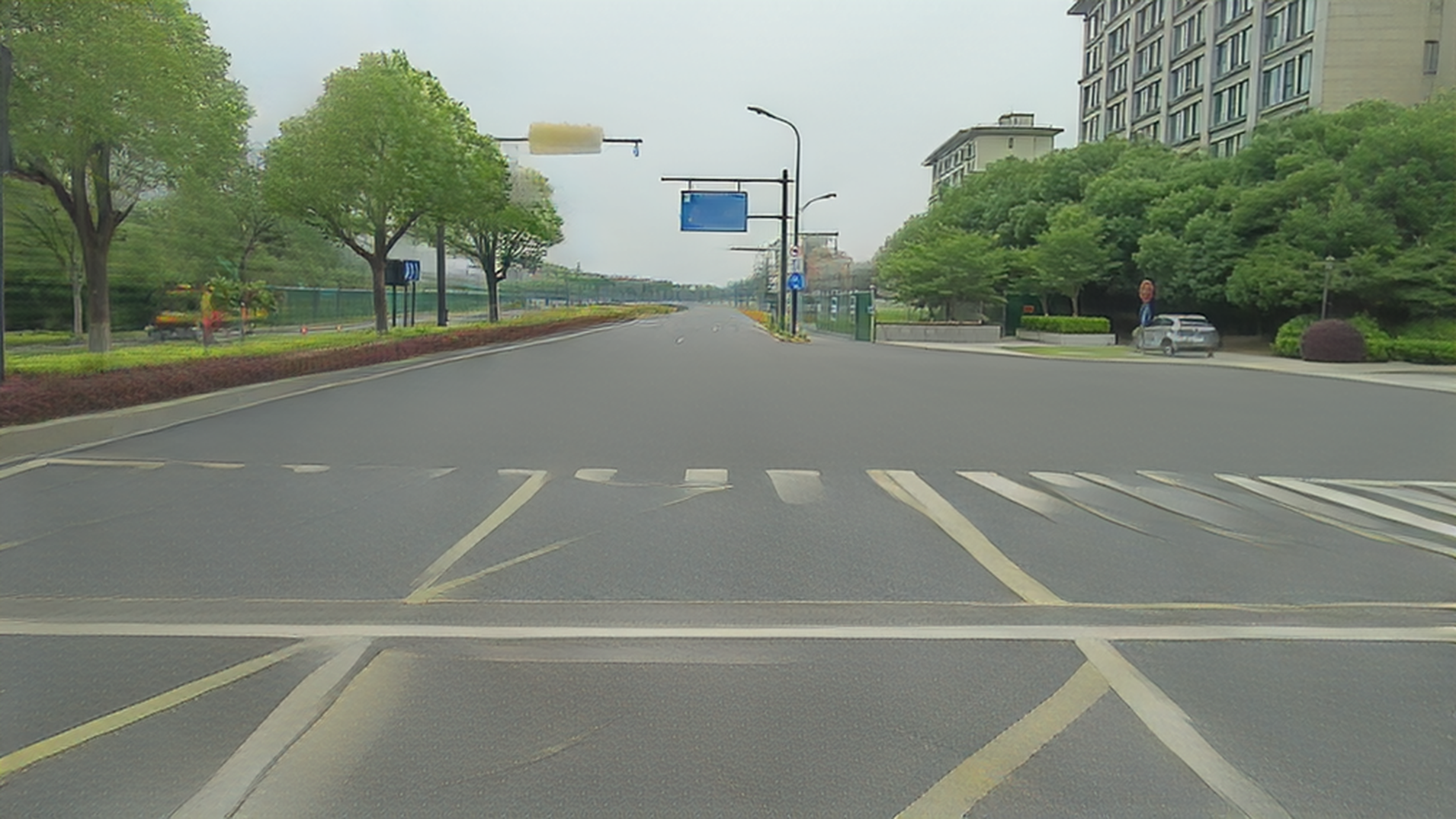}&
\includegraphics[height=0.092\textwidth]{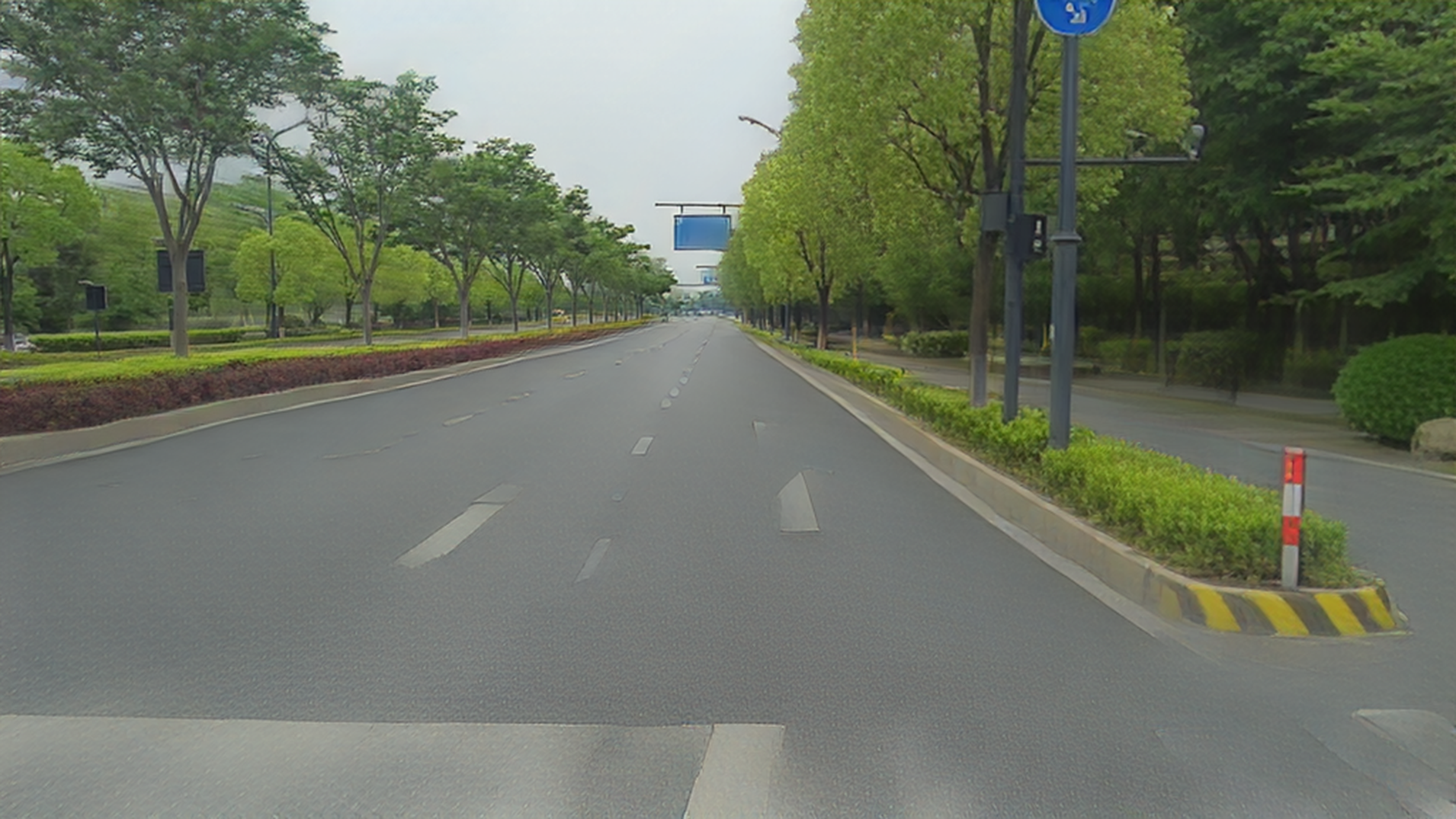}
\vspace{-3pt}\\
\includegraphics[height=0.092\textwidth]{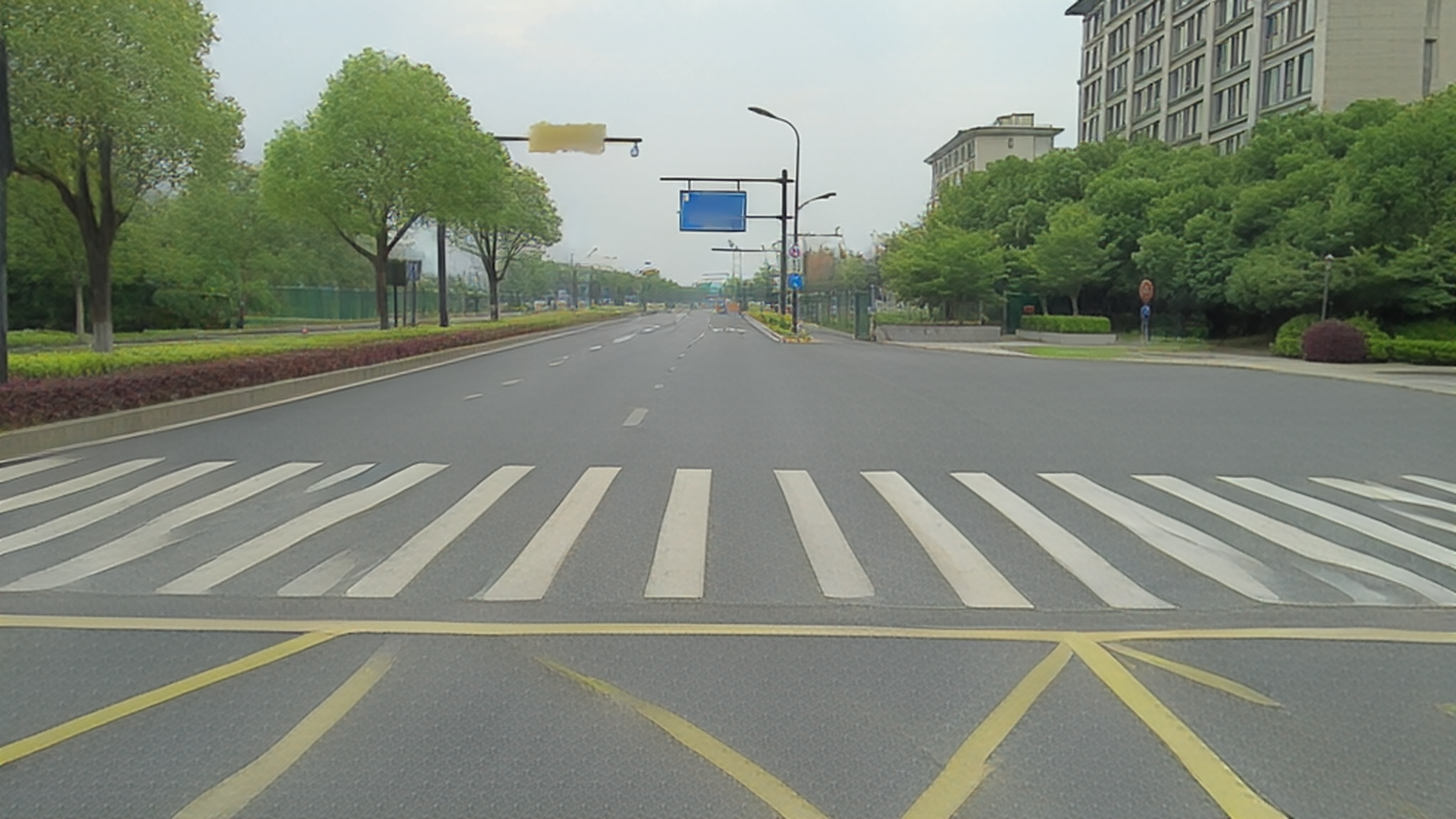}&
\includegraphics[height=0.092\textwidth]{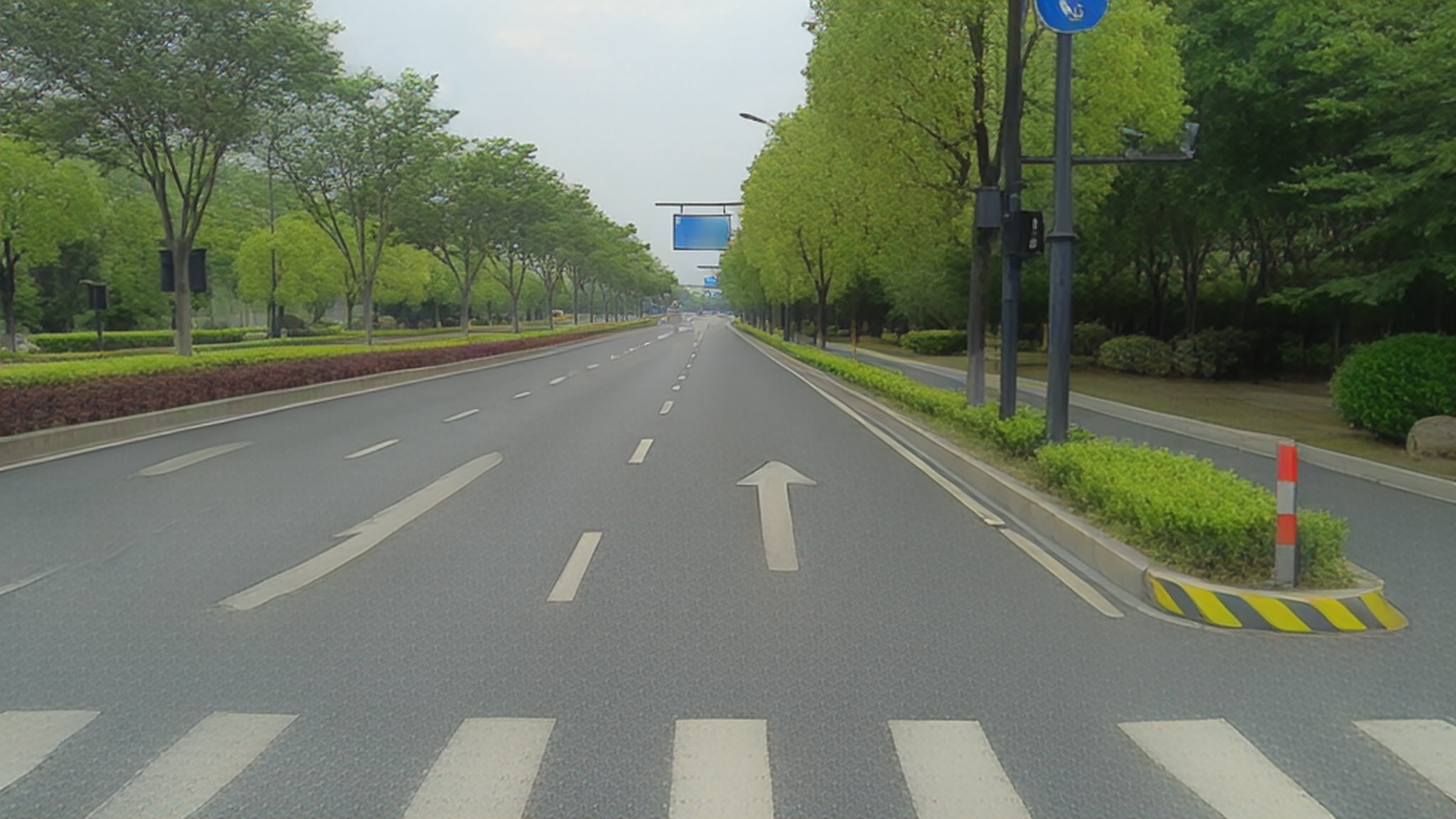}
\vspace{-3pt}\\
\includegraphics[height=0.092\textwidth]{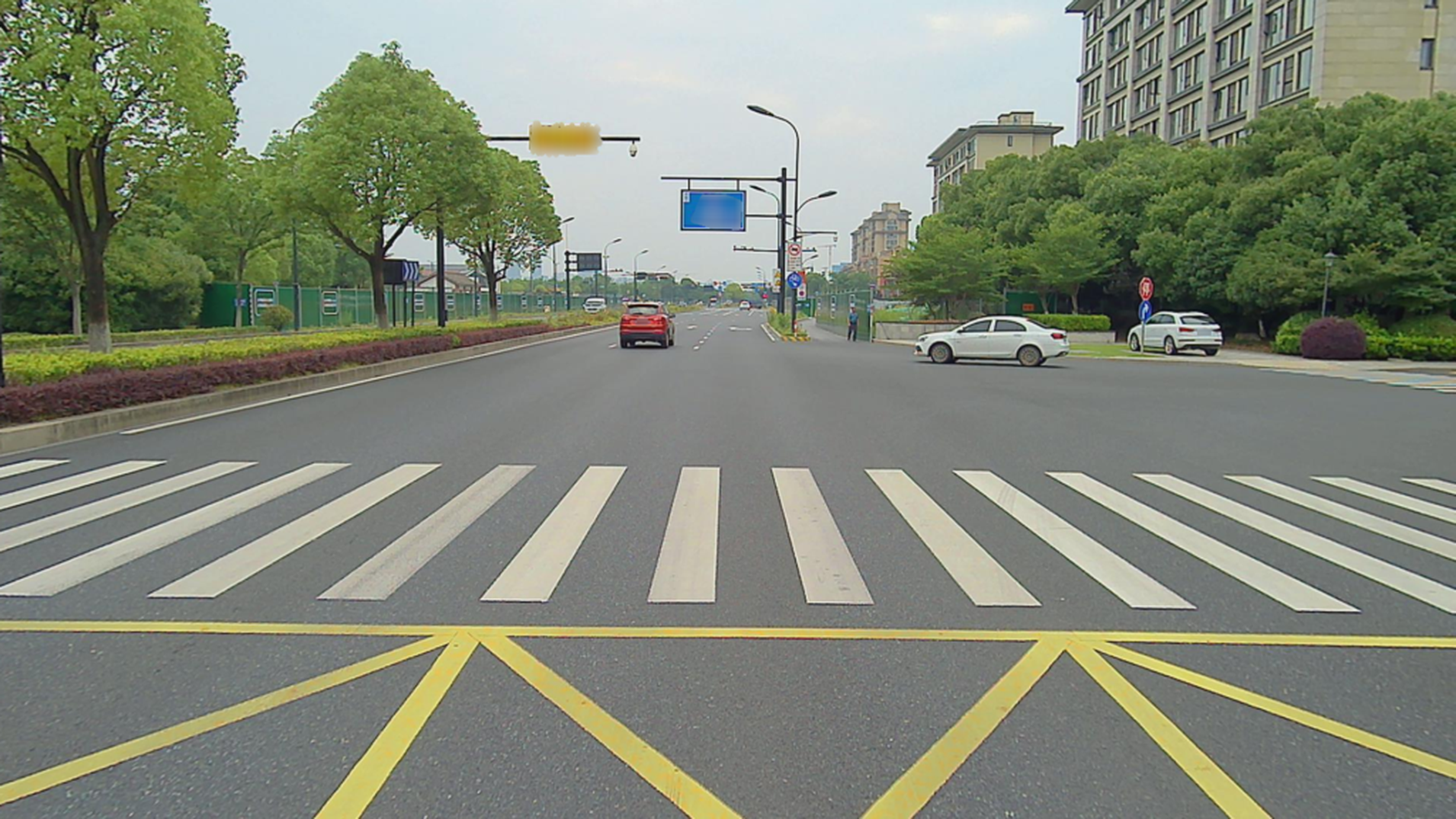}&\hspace{-6pt}
\includegraphics[height=0.092\textwidth]{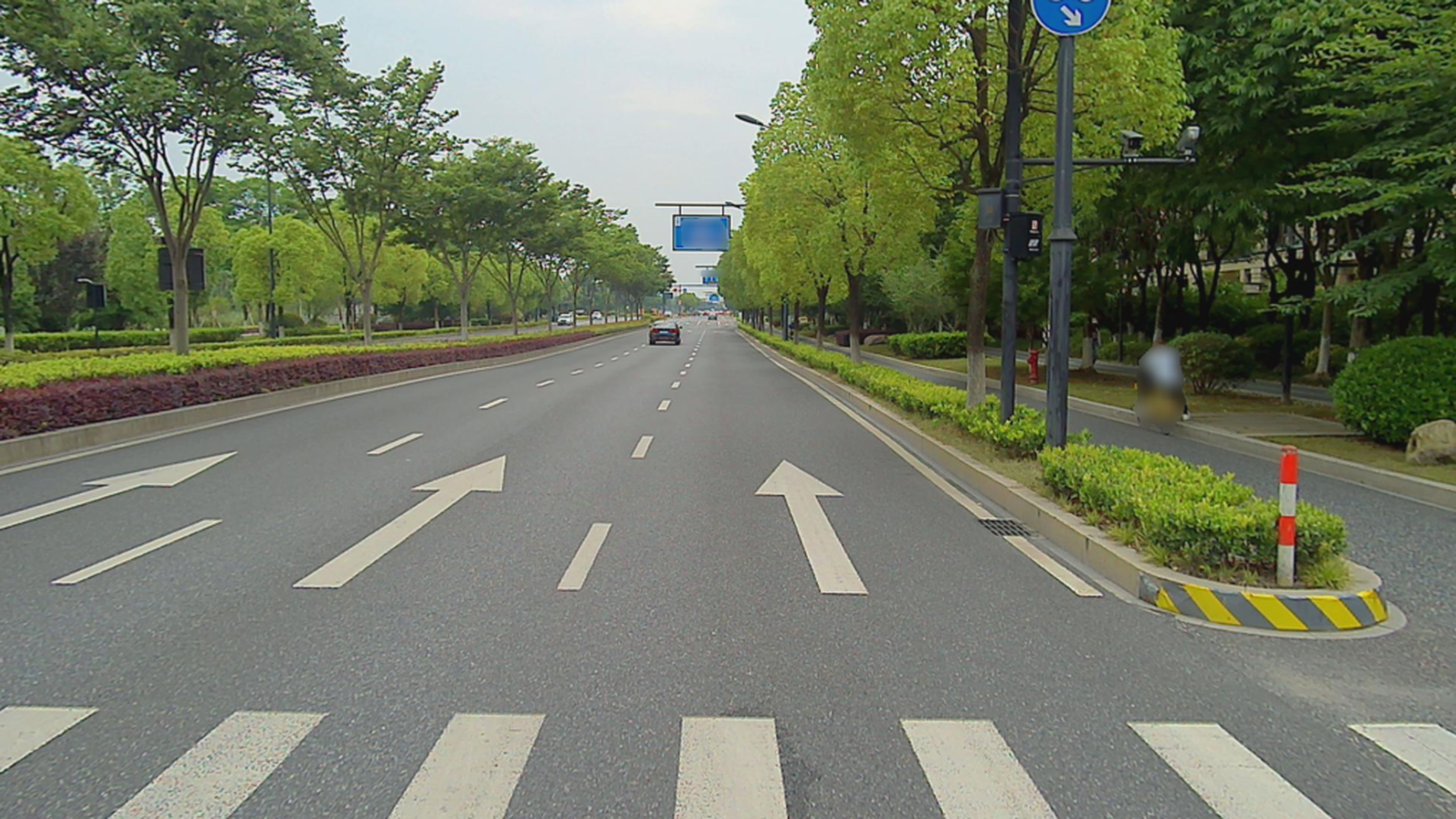} \\
\end{tabular} 
 \\
  \end{tabular}
}
\vspace{-4pt}
\caption{\textbf{Novel-view images of EVUS (col. 1-4) and Para-Lane (col. 5-6).} The qualitative comparison of 3DGS and enhanced versions by DiFix3D and \name. The \name results show a higher fidelity to the scene visual contents.}
\label{fig:euvs_paralene_examples}
\end{figure*}
\subsection{Main Results}
We compare \name against state-of-the-art Gaussian Splatting and diffusion enhancers for novel view synthesis on four datasets as follows.

\noindent\textbf{EUVS results}
\Cref{tab:euvs_table1} reports the performance of 3D reconstruction baselines and diffusion-based enhancement models evaluated on 3DGS-rendered images. Across all three EUVS novel-view configurations, \name consistently improves image fidelity and perceptual realism, outperforming state-of-the-art. The gains are particularly notable in Settings~2 and~3, which involve more challenging extrapolation regimes. Visual comparisons in \Cref{fig:euvs_paralene_examples} illustrate how our method restores missing structure and reduces rendering artifacts in extrapolated views, producing higher-quality results compared to competing techniques.

\noindent\textbf{Para-Lane results}
\Cref{tab:paralane_table1} presents results on the Para-Lane dataset. Our method achieves superior performance across both single- and double-lane shifts, reducing LPIPS by approximately 50\% relative to raw 3DGS renderings. Compared to other diffusion-based enhancers, \name delivers roughly a 1~dB increase in PSNR while achieving lower LPIPS scores, indicating improved visual realism alongside stronger photometric consistency with the underlying scene. Qualitative examples in \Cref{fig:euvs_paralene_examples} demonstrate sharper textures, more coherent geometric details, and robust restoration under cross-lane viewpoint shifts.
\begin{table}[t!]
\centering
\setlength{\tabcolsep}{16pt} % default is 6pt
\renewcommand{\arraystretch}{1.1} % row spacing (default = 1.0)
\resizebox{\textwidth}{!}{
\begin{tabular}{lccc|ccc}
\toprule
\multirow{2}{*}{\textbf{Method}} & \multicolumn{3}{c}{\textbf{Single-lane transition}} & \multicolumn{3}{c}{\textbf{Double-lane transition}} \\
\cmidrule(lr){2-4} \cmidrule(lr){5-7} 
& \textbf{PSNR} & \textbf{SSIM} & \textbf{LPIPS} & \textbf{PSNR} & \textbf{SSIM} & \textbf{LPIPS} \\
\midrule
3DGS~\cite{kerbl2023gaussian} & 17.05 &	0.524 &	0.446 & 16.26 &	0.505 &	0.472  \\
2DGS~\cite{2dgs2023} & 16.79 &	0.523 &	0.469 & 16.01 &	0.510 &	0.494  \\
GSPro~\cite{cheng2024gaussianpro} & 17.01 &	0.521 &	0.446 & 16.29 &	0.505 &	0.472 \\
Scaffold-GS~\cite{lu2024scaffold} & 17.59 &	0.538 &	0.450 & 17.09 & 0.525 & 0.470  \\
StreetGS~\cite{yan2024street} & 17.50 &	0.510 &	0.456 & 16.16 &	0.496 &	0.480 \\
\midrule
3DGS~\cite{kerbl2023gaussian} & 17.05 &	0.524 &	0.446 & 16.26 &	0.505 &	0.472  \\
\quad + DIFIX3D~\cite{wu2025difix3d} & 17.62 & 0.470 & 0.234 & 16.87 & 0.447 & 0.271 \\
\quad + DIFIX3D++~\cite{omran2025hybrid} & 19.25 & 0.559 & 0.275 & 18.55 & 0.538 & 0.311  \\
\rowcolor{lightblue} \quad + \name & \textbf{20.20} & \textbf{0.586} & \textbf{0.208} & \textbf{19.47} & \textbf{0.564} & \textbf{0.233} \\
\bottomrule
\end{tabular}
}
\caption{\small \textbf{SOTA comparisons on  Para-Lane} across two extrapolation settings. Neural reconstruction (top) and diffusion enhancers (bottom) comparisons are reported separately. All diffusion enhancers are applied on the same 3DGS backbone.}
\label{tab:paralane_table1}
\vspace{-5pt}
\end{table}
\begin{table}[t]
\centering
\caption{\textbf{SOTA comparisons on nuScenes} reported on extrapolated trajectories.}
\setlength{\tabcolsep}{24pt} % default is 6pt
\renewcommand{\arraystretch}{1.1} % row spacing (default = 1.0)

\resizebox{.7\textwidth}{!}{
\begin{tabular}{lccc}
    \toprule
    \textbf{Method} & \textbf{PSNR} & \textbf{SSIM} & \textbf{LPIPS} \\
    \midrule
    OmniRe~\cite{chen2025omnire} & 17.15 & 0.645 & 0.448 \\
    \quad + DIFIX3D~\cite{wu2025difix3d} & 17.12 & 0.612 & 0.371 \\
    \quad + DIFIX3D++~\cite{omran2025hybrid} & 18.74 & 0.696 & 0.375 \\
    %\midrule
    \rowcolor{lightblue} \quad + \name & \textbf{21.58} & \textbf{0.716} & \textbf{0.303}  \\
    %ours + dyn. masks & 21.37 & 0.661 & 0.292  \\
    \bottomrule
    \end{tabular}
}
\label{tab:nuscenes_results_tab}
\vspace{-4pt}
\end{table}

\begin{figure*}[!b]
\centering
 \resizebox{.95\linewidth}{!}{
  \includegraphics[width=\textwidth, clip]{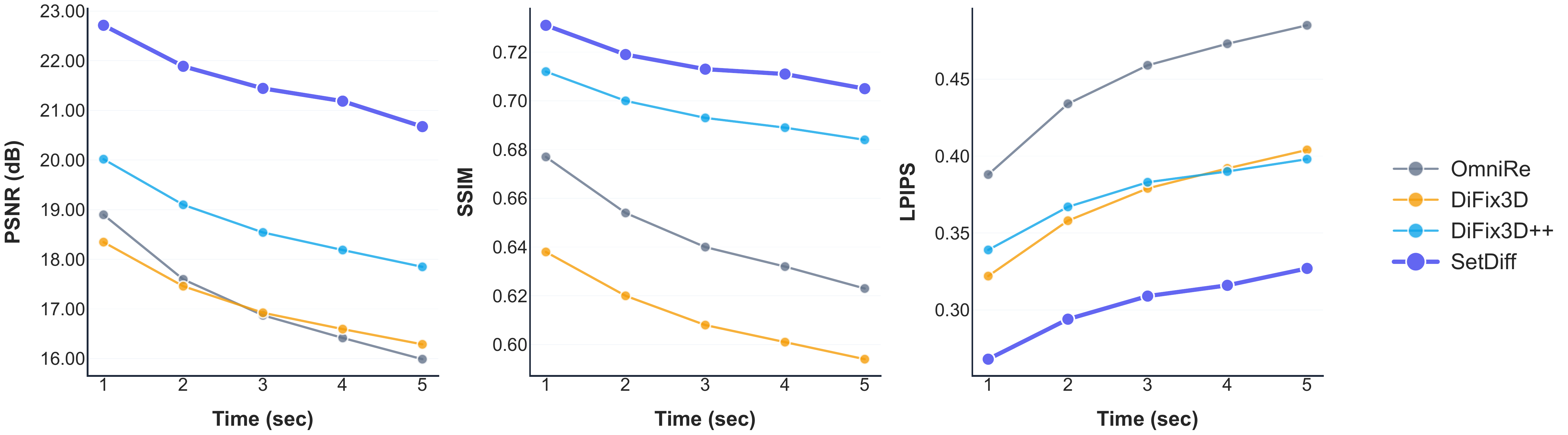}
  }
  
  \caption{
    \textbf{SOTA comparisons on nuScenes} reported under various trajectory lengths.
  }
  \vspace{-5pt}
   \label{fig:nuscenes_plots}
\end{figure*}

\begin{figure*}[!ht]
\centering
 \resizebox{.99\textwidth}{!}{
%\begin{tabular}{c}
%\input{floats/figures/nuscenes/single_cam_ex_1}
%\\
%\input{floats/figures/nuscenes/single_cam_ex_2}
%\end{tabular}
\includegraphics[width=\textwidth, trim={0cm 3.0cm 0cm 0.5cm}, clip]{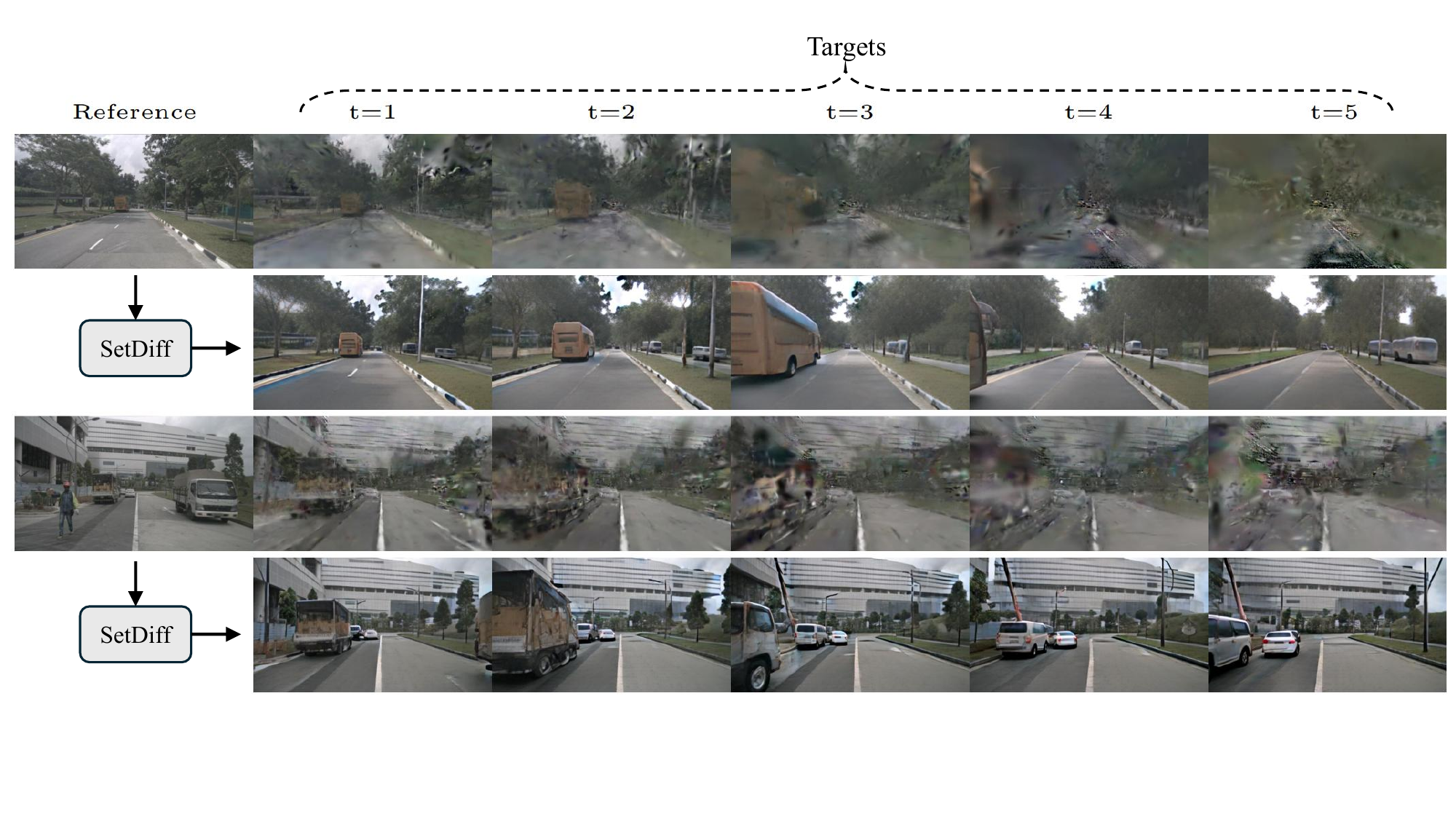}
  }
  \caption{\textbf{Novel-view images of nuScenes}; (odd rows) a single reference image and temporally-extrapolated rendered views from OmniRe (dynamic 3DGS model), up to 5 seconds; (even rows) enhanced images, conditioned on reference view, using \name.}
   \label{fig:nuscenes_single_cam}
   \vspace{-3pt}
\end{figure*}

\noindent\textbf{nuScenes results}
\Cref{tab:nuscenes_results_tab} summarizes novel-view synthesis performance using all six nuScenes cameras over 5-second temporally extrapolated sequences generated by OmniRe~\cite{chen2025omnire}. \name substantially improves image quality by correcting severe distortions and reconstructing missing details. As shown in \Cref{fig:nuscenes_plots}, metrics grouped by temporal offset reveal a progressive drop in Signal-to-noise ratio (SNR) with longer extrapolation, which is also visually evident in \Cref{fig:nuscenes_single_cam}. Across all temporal buckets, \name delivers consistently higher fidelity and more realistic reconstructions, effectively restoring information even under long-horizon predictions. Notably, in low-SNR regimes the C-maps become highly noisy and unreliable; however, training \name across varying SNR levels enables the model to exploit C-map conditioning whenever it is reliable.
\begin{figure*}[ht!]
\centering
\resizebox{1.0\textwidth}{!}{
\begin{tabular}{c}
 \begin{tabular}{@{}c@{\hspace{2pt}}c@{}c@{}c@{}c@{}c@{}c@{}}
 %% Example 1
% \includegraphics[width=.16\textwidth]{images/nuscenes/render_omnire_on_trainval_partial_scene=200/0/source/n008-2018-08-21-11-53-44-0400__CAM_FRONT_LEFT__1534867409504799.jpg} &
%  \includegraphics[width=.16\textwidth]{images/nuscenes/render_omnire_on_trainval_partial_scene=200/0/source/n008-2018-08-21-11-53-44-0400__CAM_FRONT__1534867409512404.jpg} &
%\includegraphics[width=.16\textwidth]{images/nuscenes/render_omnire_on_trainval_partial_scene=200/0/source/n008-2018-08-21-11-53-44-0400__CAM_FRONT_RIGHT__1534867409420482.jpg} &
%  \includegraphics[width=.16\textwidth]{images/nuscenes/render_omnire_on_trainval_partial_scene=200/0/source/n008-2018-08-21-11-53-44-0400__CAM_BACK_RIGHT__1534867409428113.jpg} &
%  \includegraphics[width=.16\textwidth]{images/nuscenes/render_omnire_on_trainval_partial_scene=200/0/source/n008-2018-08-21-11-53-44-0400__CAM_BACK__1534867409437558.jpg} &
%  \includegraphics[width=.16\textwidth]{images/nuscenes/render_omnire_on_trainval_partial_scene=200/0/source/n008-2018-08-21-11-53-44-0400__CAM_BACK_LEFT__1534867409447405.jpg}
%  \\
  \rotatebox{90}{\makebox[1.2cm][c]{\tiny{OmniRe}}} &
  \includegraphics[width=.16\textwidth]{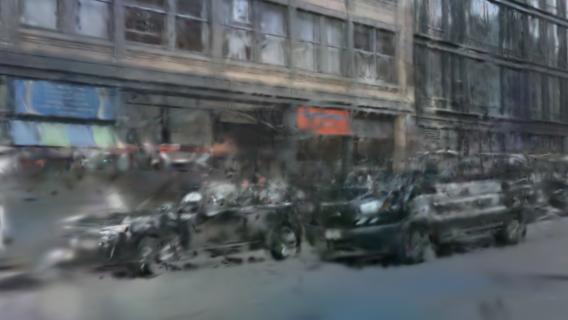} &
  \includegraphics[width=.16\textwidth]{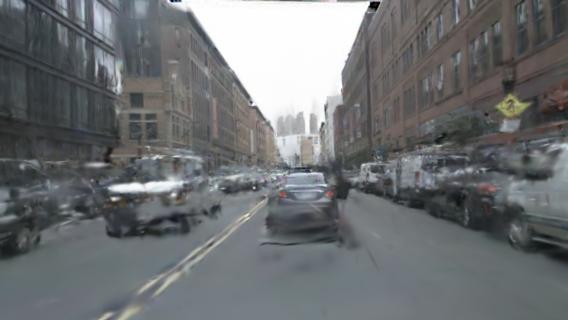} &
  \includegraphics[width=.16\textwidth]{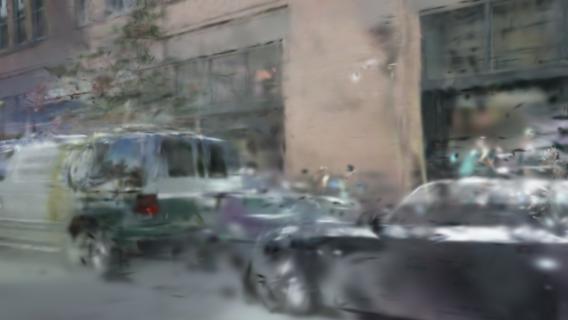} &
  \includegraphics[width=.16\textwidth]{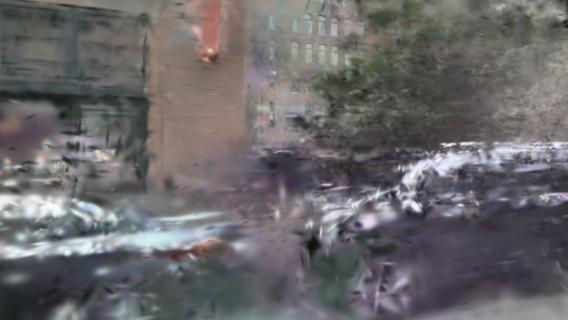} &
 \includegraphics[width=.16\textwidth]{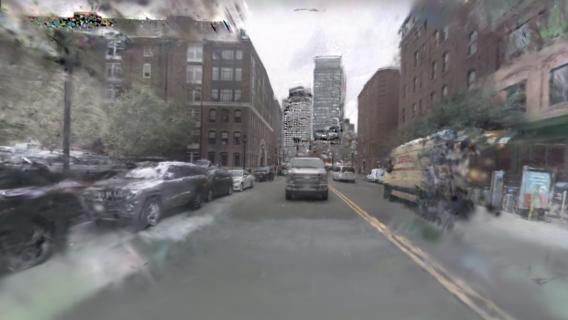} &
 \includegraphics[width=.16\textwidth]{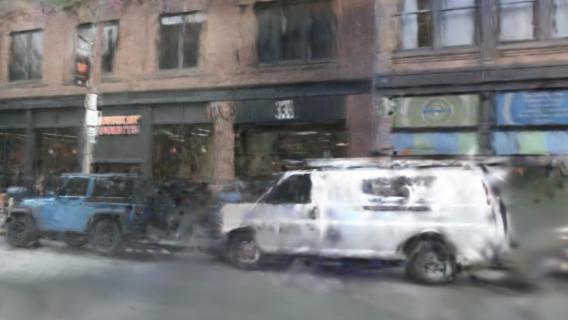}
  \vspace{-5pt}\\
  \rotatebox{90}{\makebox[1.2cm][c]{\tiny{DiFix3D}}} &
  \includegraphics[width=.16\textwidth]{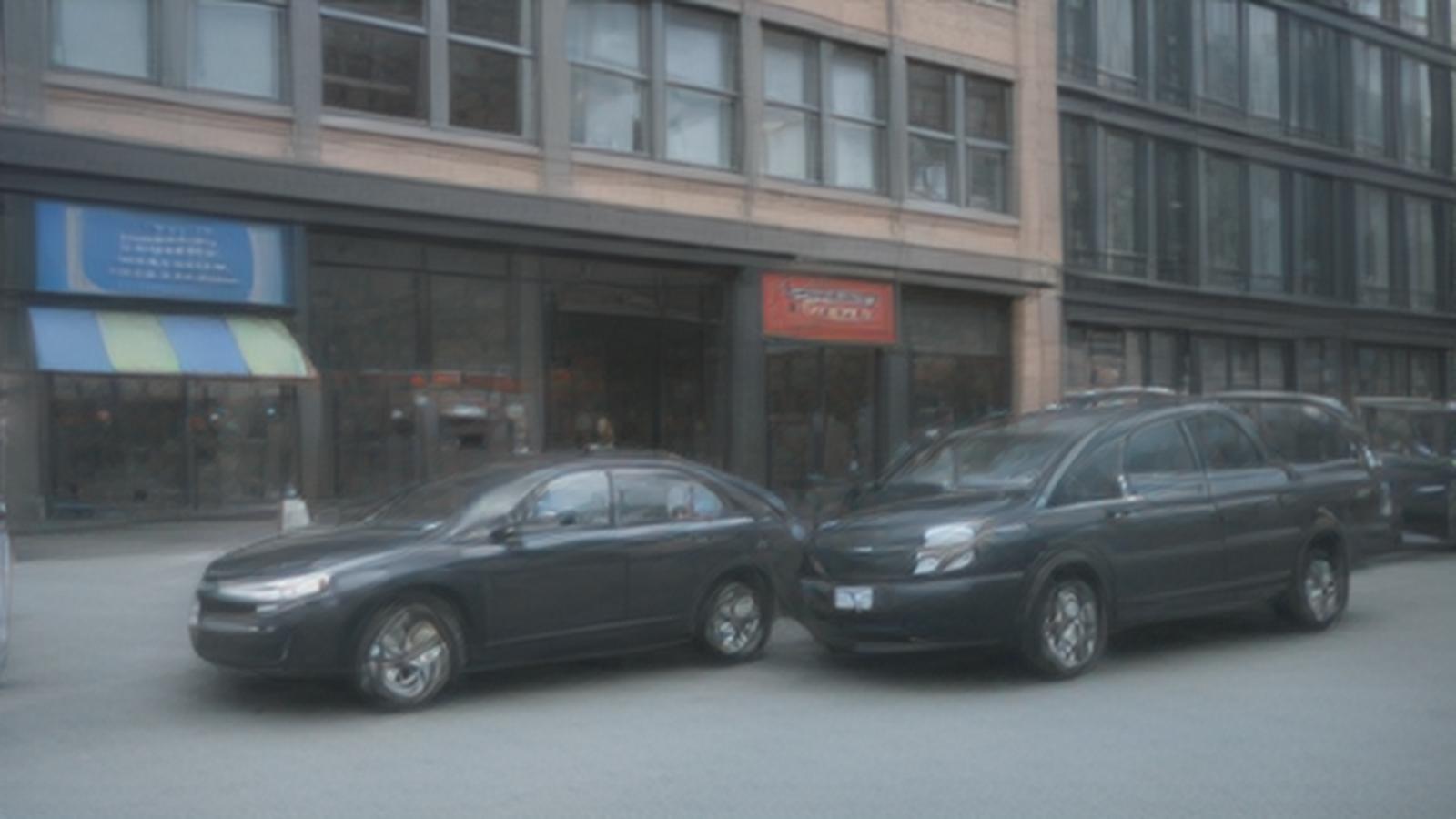} &
  \includegraphics[width=.16\textwidth]{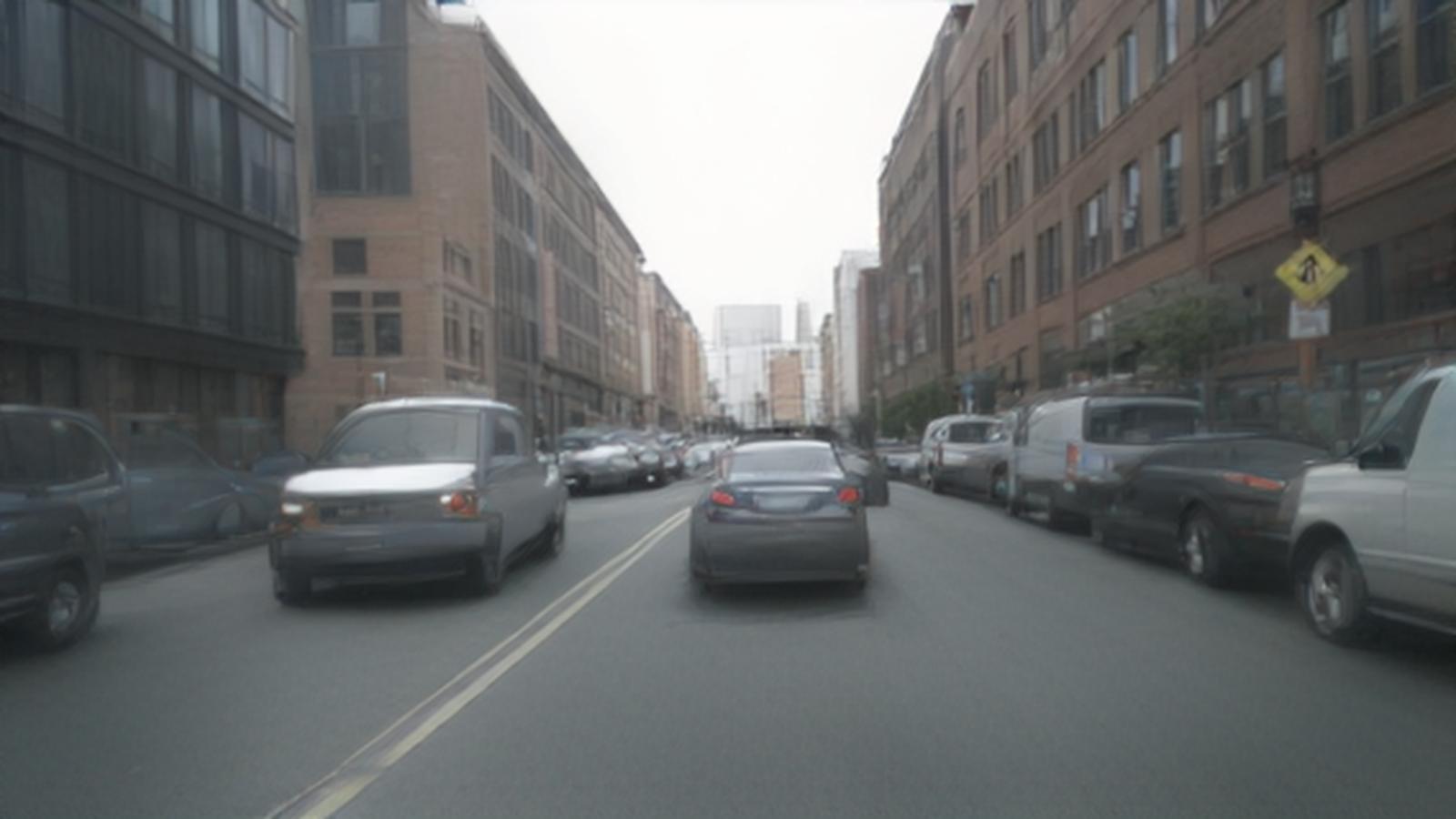} &
  \includegraphics[width=.16\textwidth]{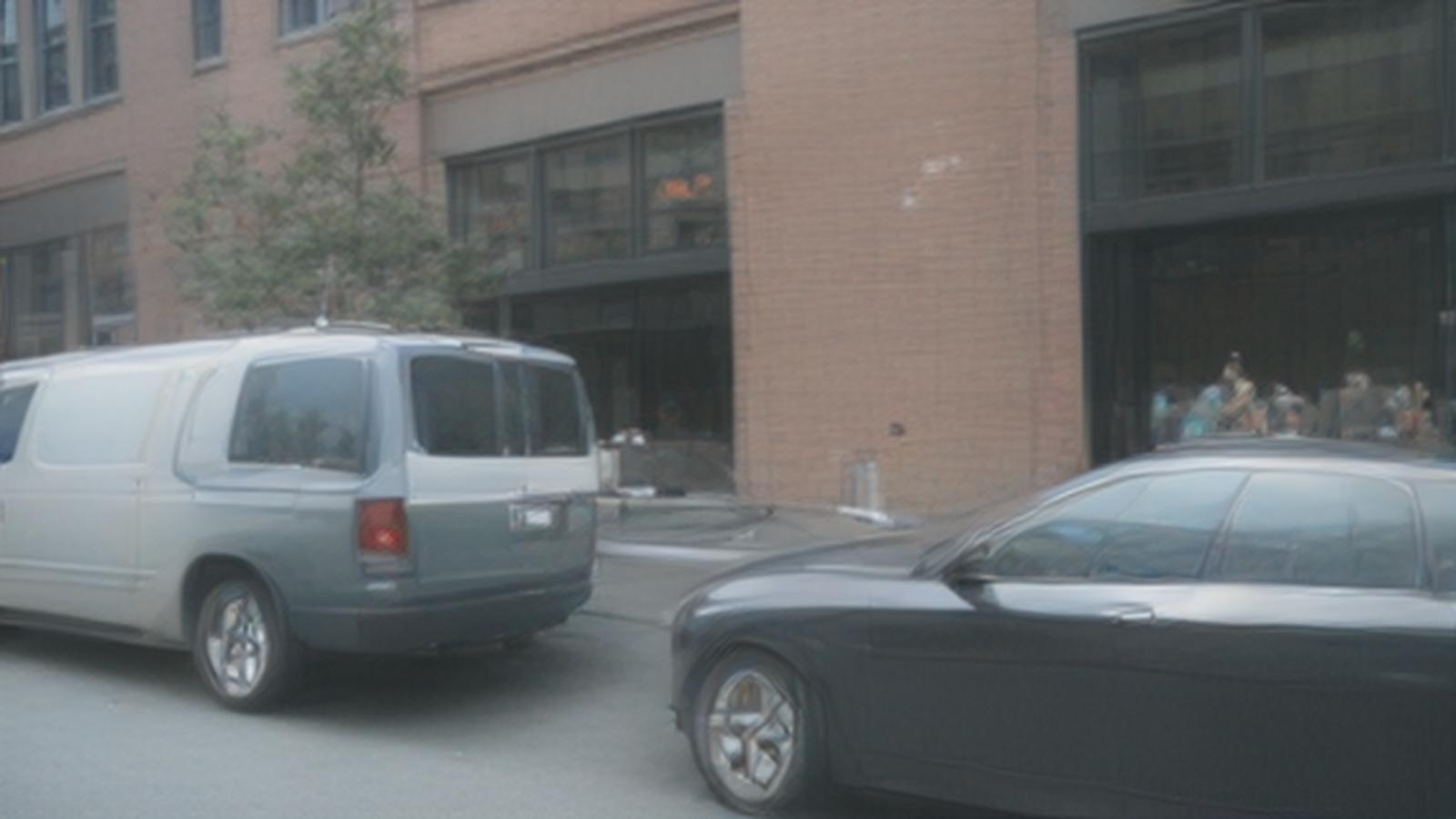} &
  \includegraphics[width=.16\textwidth]{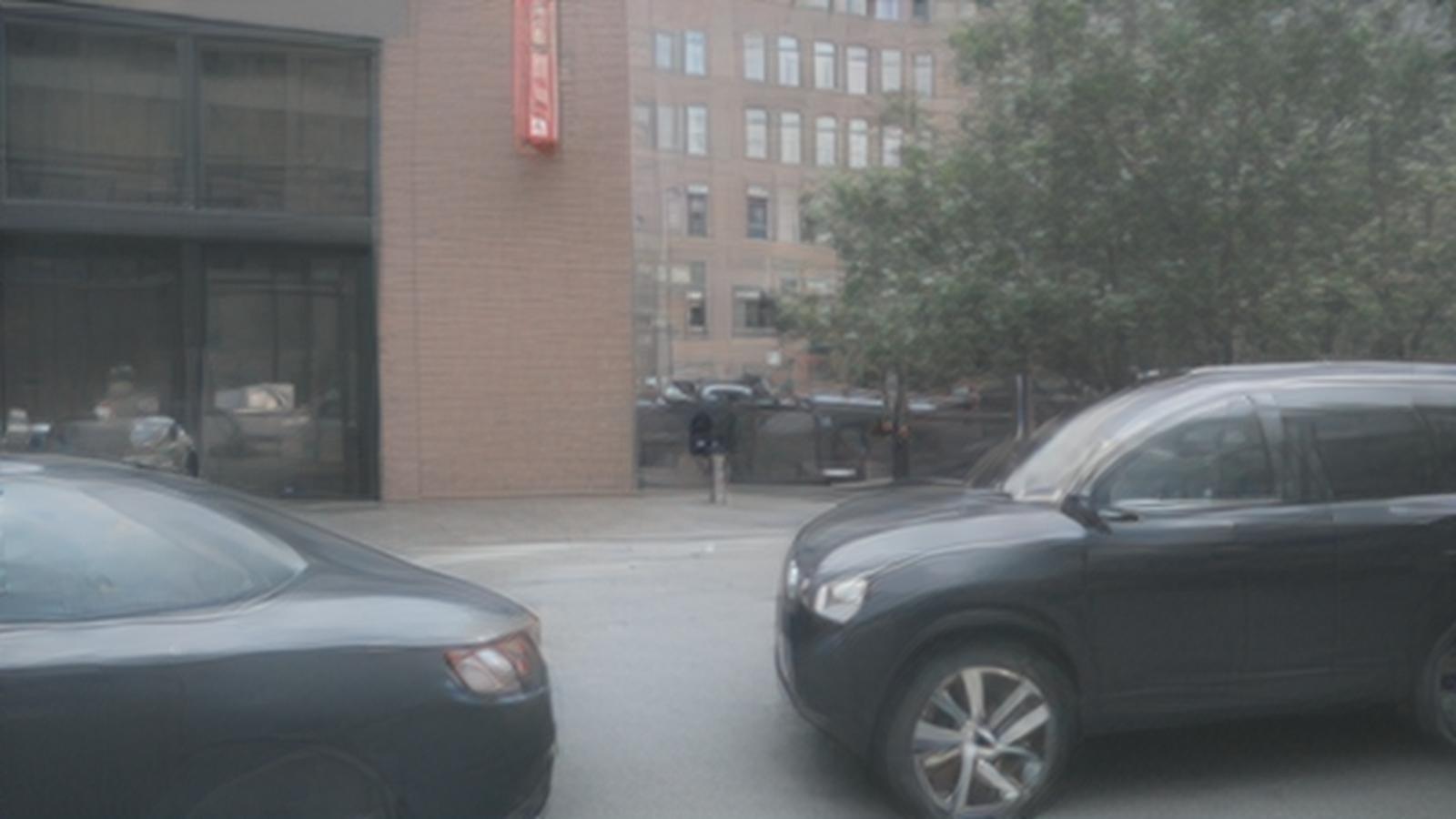} &
 \includegraphics[width=.16\textwidth]{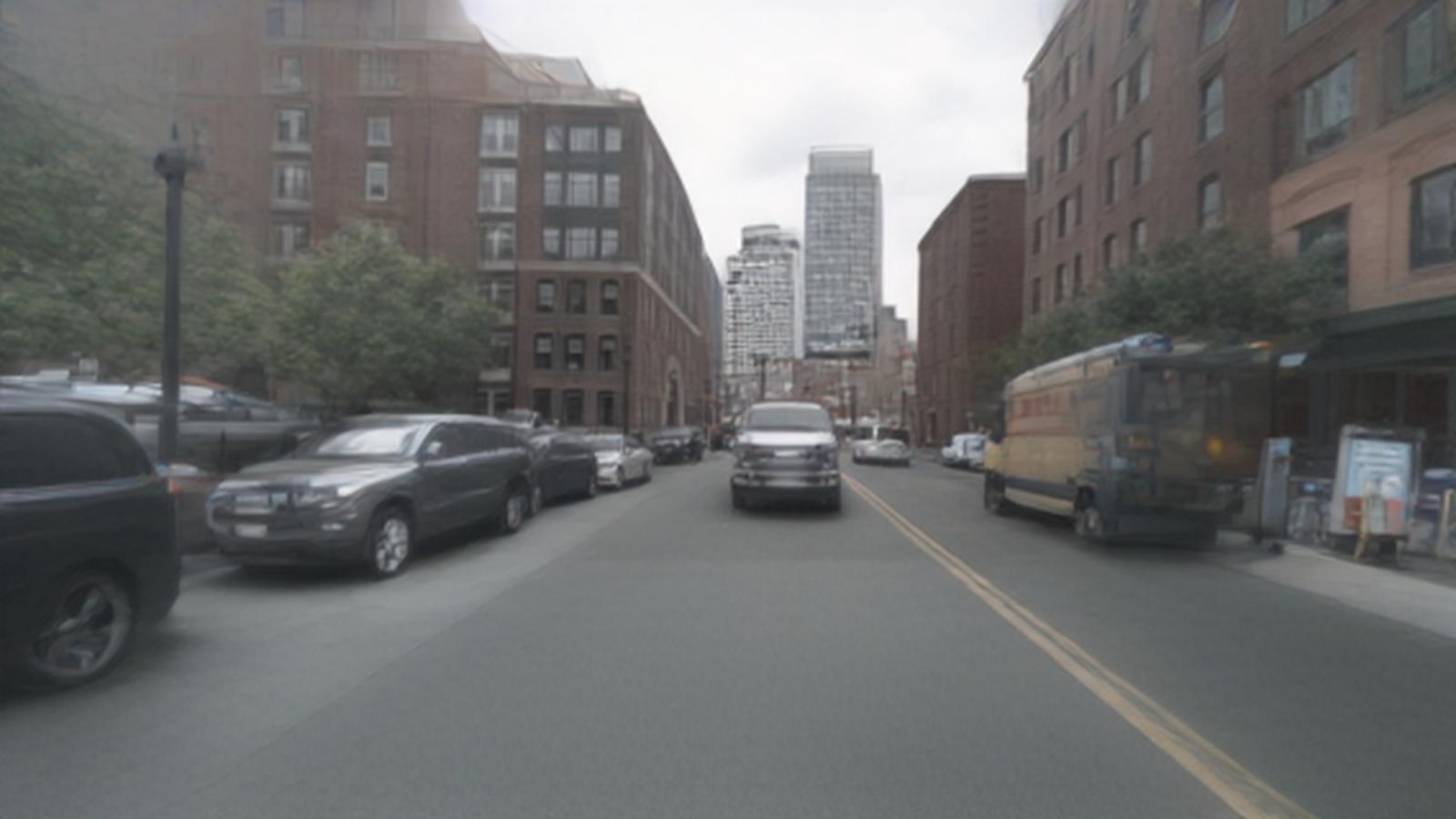} &
 \includegraphics[width=.16\textwidth]{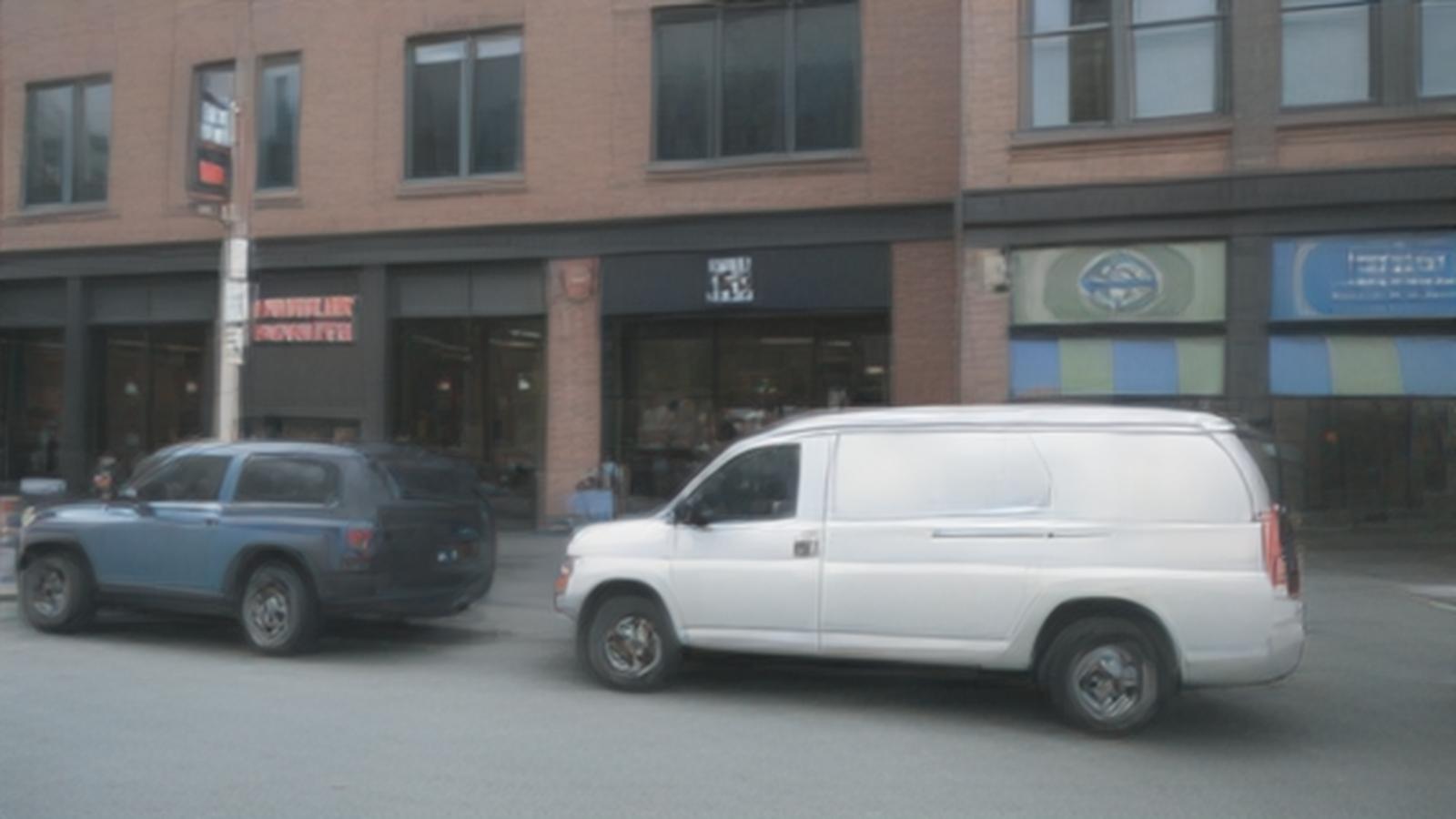}
  \vspace{-5pt}\\
  \rotatebox{90}{\makebox[1.2cm][c]{\tiny{Ours}}} &
  \includegraphics[width=.16\textwidth]{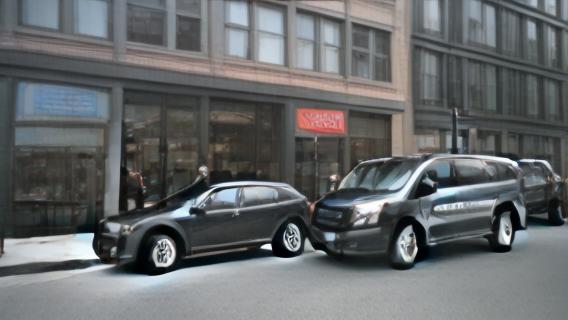} &
  \includegraphics[width=.16\textwidth]{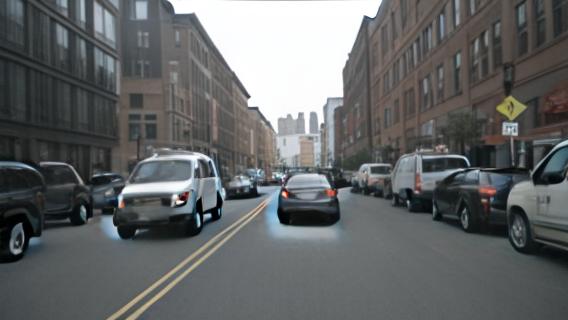} &
  \includegraphics[width=.16\textwidth]{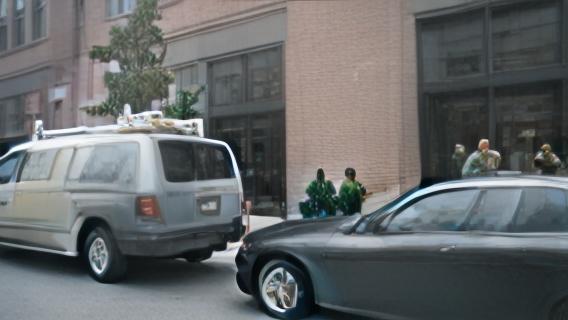} &
  \includegraphics[width=.16\textwidth]{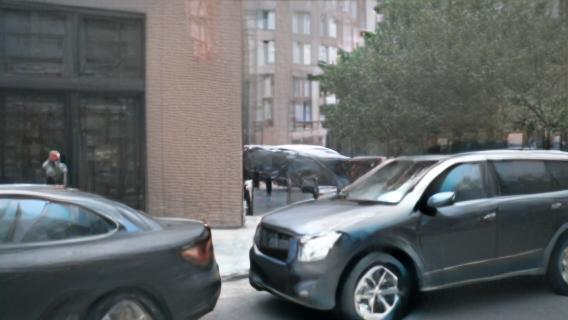} &
 \includegraphics[width=.16\textwidth]{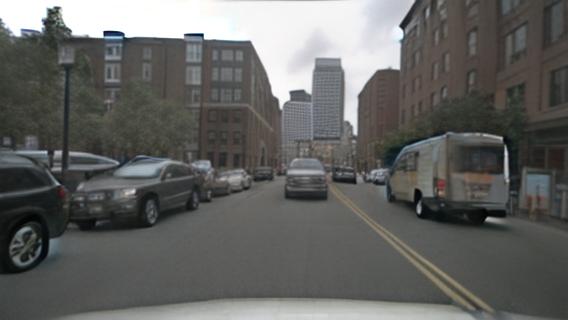} &
 \includegraphics[width=.16\textwidth]{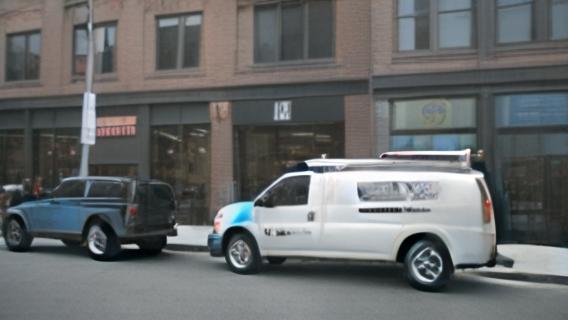}
 \vspace{-5pt}\\
\end{tabular}
\\
 \begin{tabular}{@{}c@{\hspace{2pt}}c@{}c@{}c@{}c@{}c@{}c@{}}
 %% Example 1
% \includegraphics[width=.16\textwidth]{images/nuscenes/render_omnire_on_trainval_partial_scene=200/0/source/n008-2018-08-21-11-53-44-0400__CAM_FRONT_LEFT__1534867409504799.jpg} &
%  \includegraphics[width=.16\textwidth]{images/nuscenes/render_omnire_on_trainval_partial_scene=200/0/source/n008-2018-08-21-11-53-44-0400__CAM_FRONT__1534867409512404.jpg} &
%\includegraphics[width=.16\textwidth]{images/nuscenes/render_omnire_on_trainval_partial_scene=200/0/source/n008-2018-08-21-11-53-44-0400__CAM_FRONT_RIGHT__1534867409420482.jpg} &
%  \includegraphics[width=.16\textwidth]{images/nuscenes/render_omnire_on_trainval_partial_scene=200/0/source/n008-2018-08-21-11-53-44-0400__CAM_BACK_RIGHT__1534867409428113.jpg} &
%  \includegraphics[width=.16\textwidth]{images/nuscenes/render_omnire_on_trainval_partial_scene=200/0/source/n008-2018-08-21-11-53-44-0400__CAM_BACK__1534867409437558.jpg} &
%  \includegraphics[width=.16\textwidth]{images/nuscenes/render_omnire_on_trainval_partial_scene=200/0/source/n008-2018-08-21-11-53-44-0400__CAM_BACK_LEFT__1534867409447405.jpg}
%  \\
 \rotatebox{90}{\makebox[1.0cm][r]{\tiny{OmniRe}}} &
 \includegraphics[width=.16\textwidth]{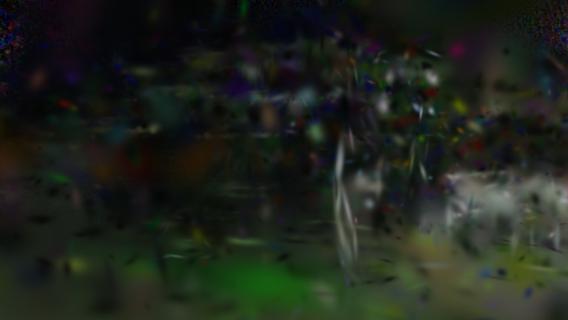} &
  \includegraphics[width=.16\textwidth]{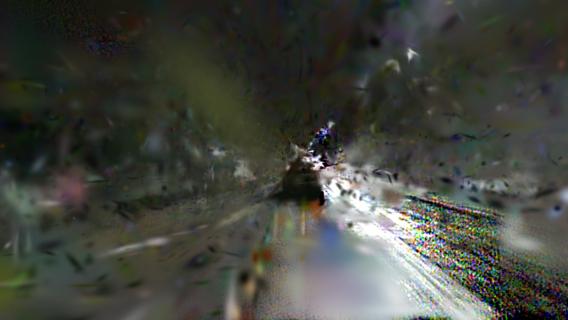} &
  \includegraphics[width=.16\textwidth]{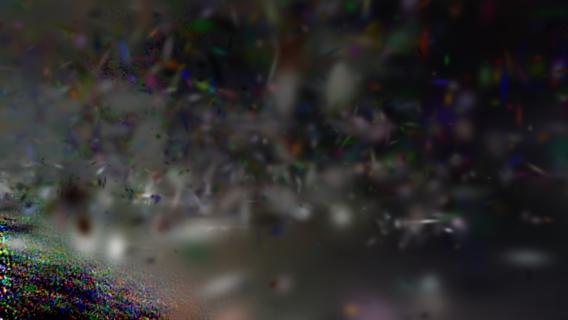}&
 \includegraphics[width=.16\textwidth]{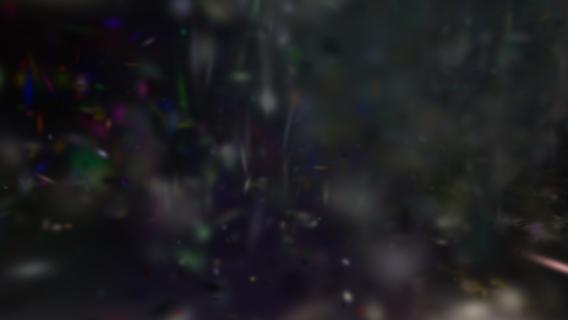} &
  \includegraphics[width=.16\textwidth]{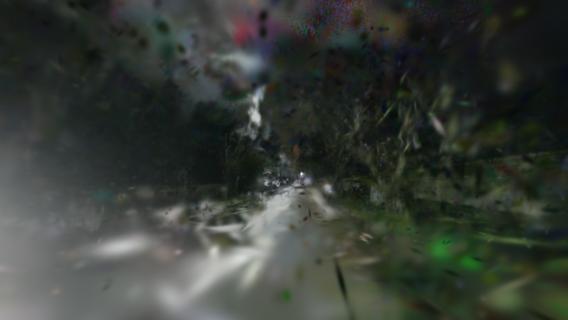} &
  \includegraphics[width=.16\textwidth]{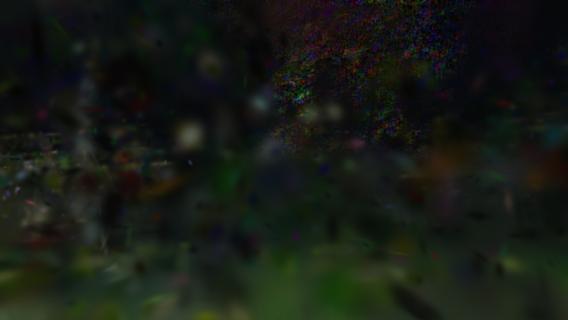} 
  \vspace{-5pt}\\
  \rotatebox{90}{\makebox[1.0cm][r]{\tiny{DiFix3D}}} &
 \includegraphics[width=.16\textwidth]{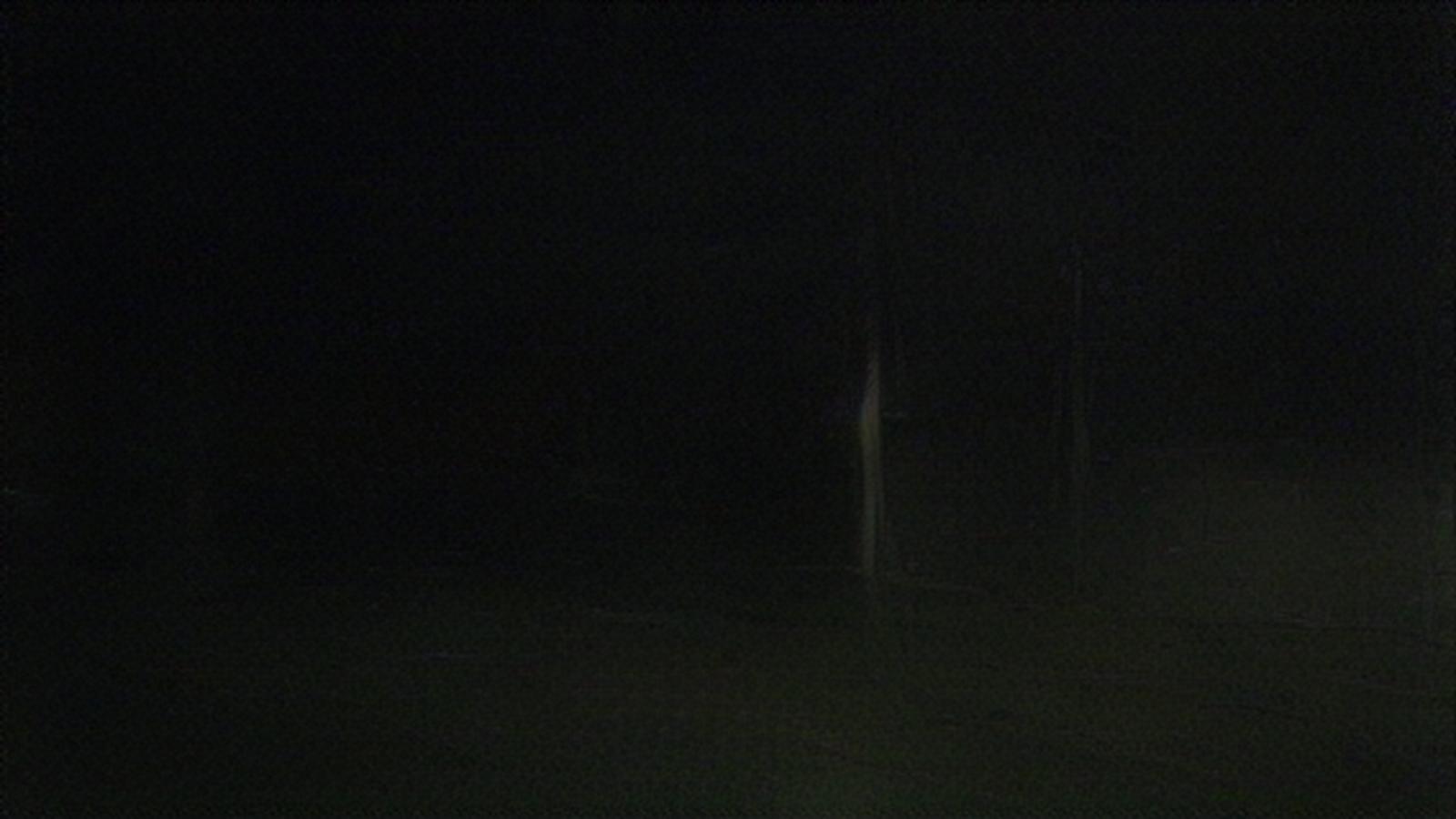} &
  \includegraphics[width=.16\textwidth]{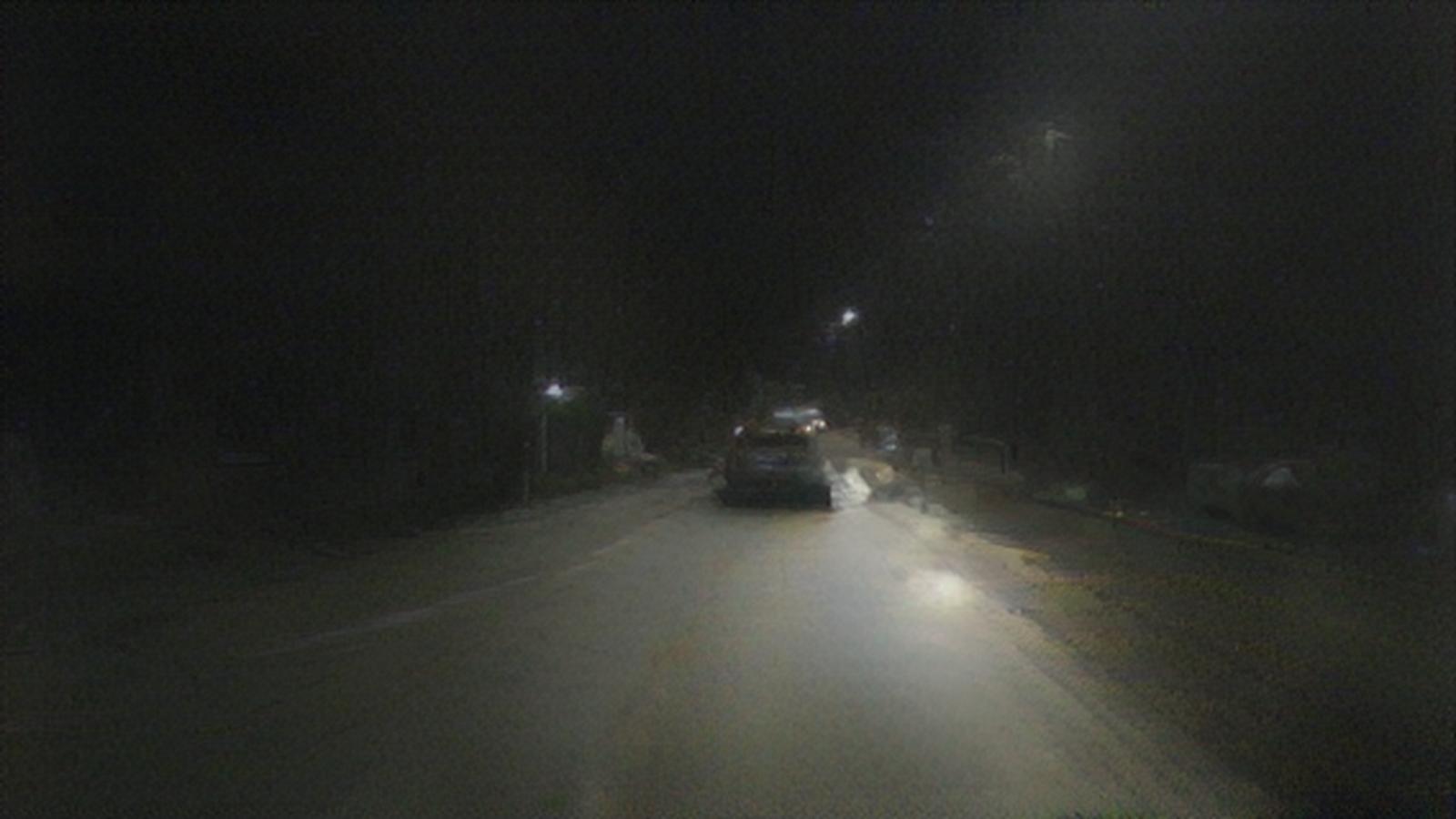} &
  \includegraphics[width=.16\textwidth]{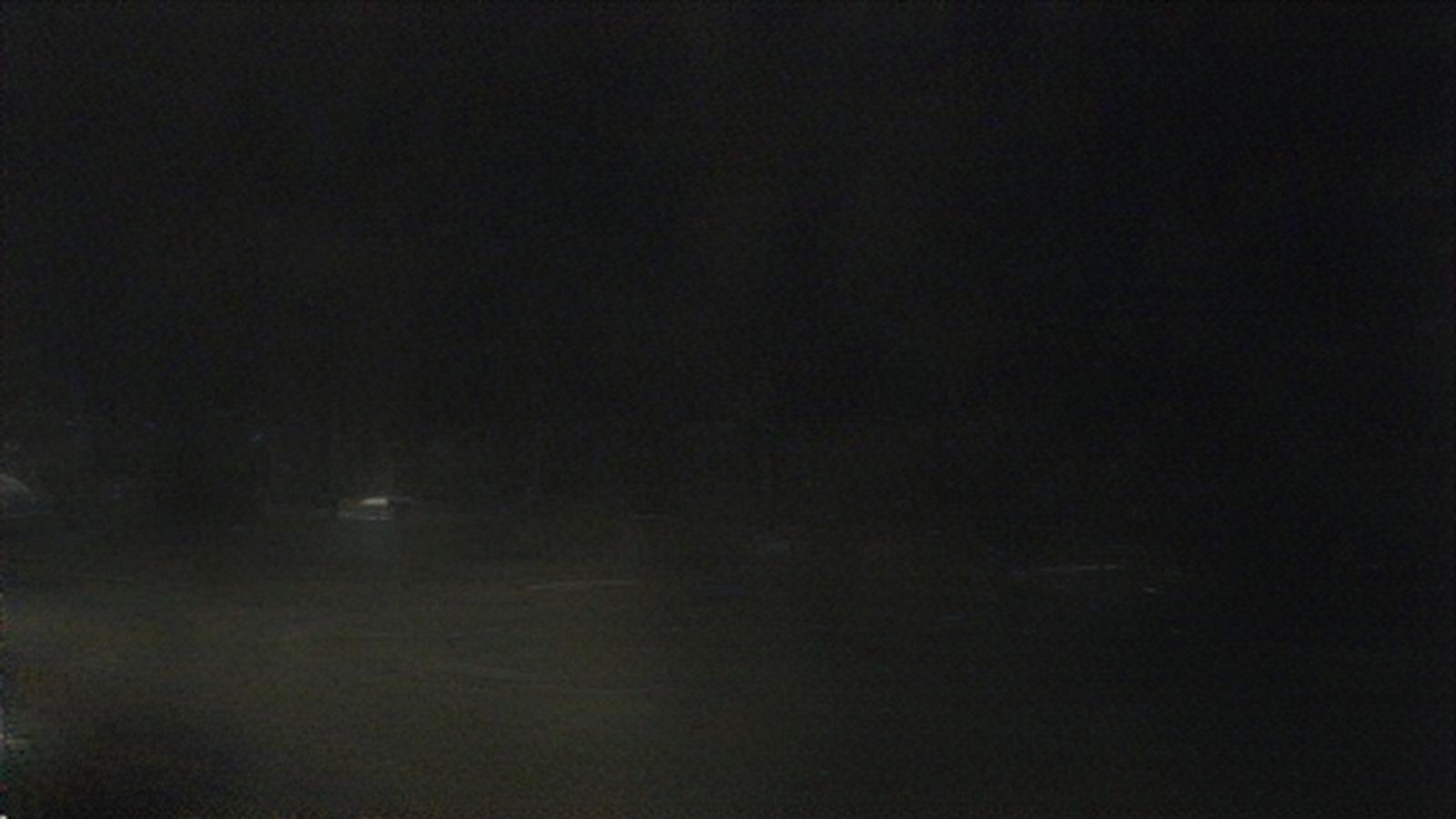} &
  \includegraphics[width=.16\textwidth]{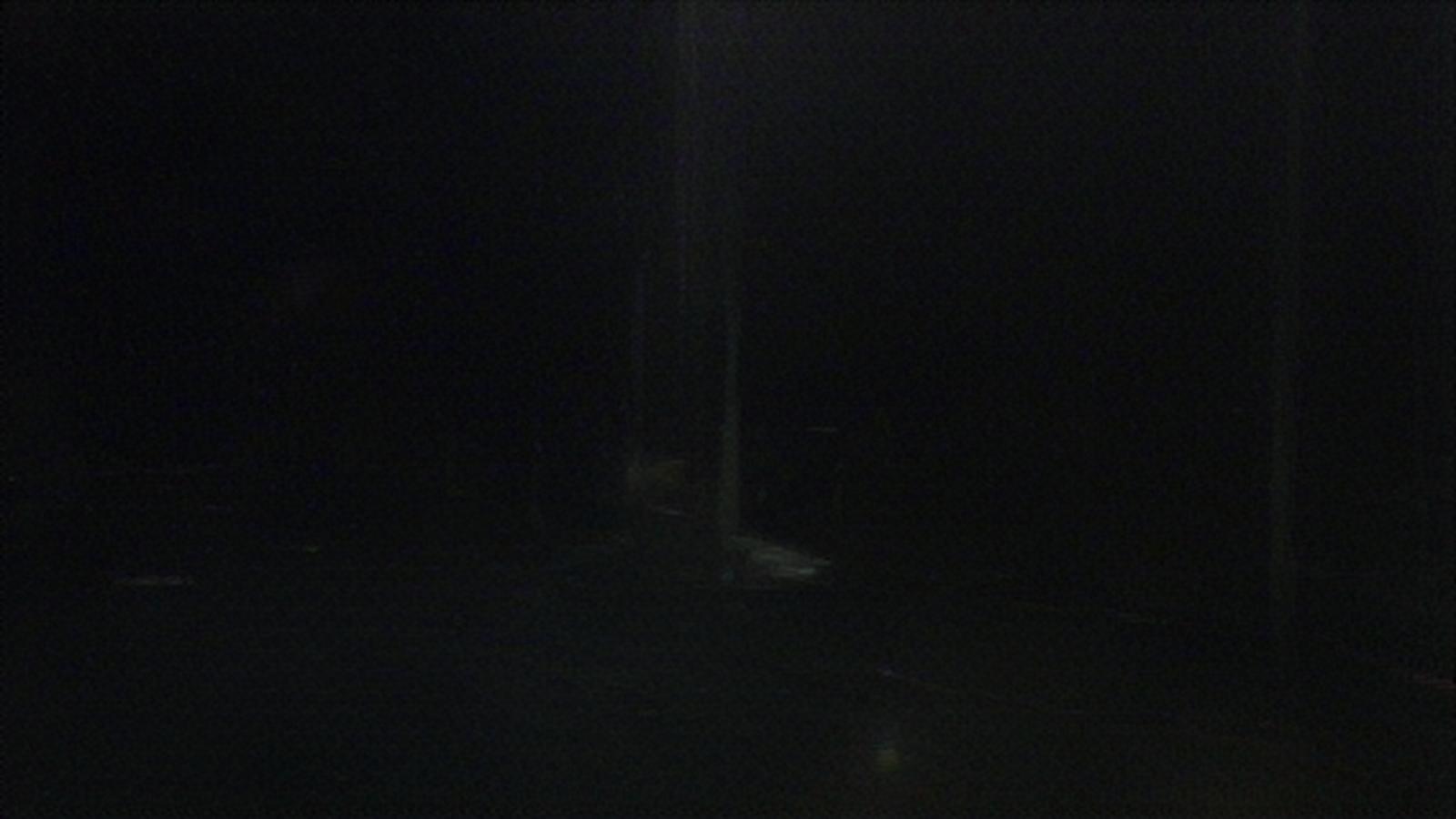} &
   \includegraphics[width=.16\textwidth]{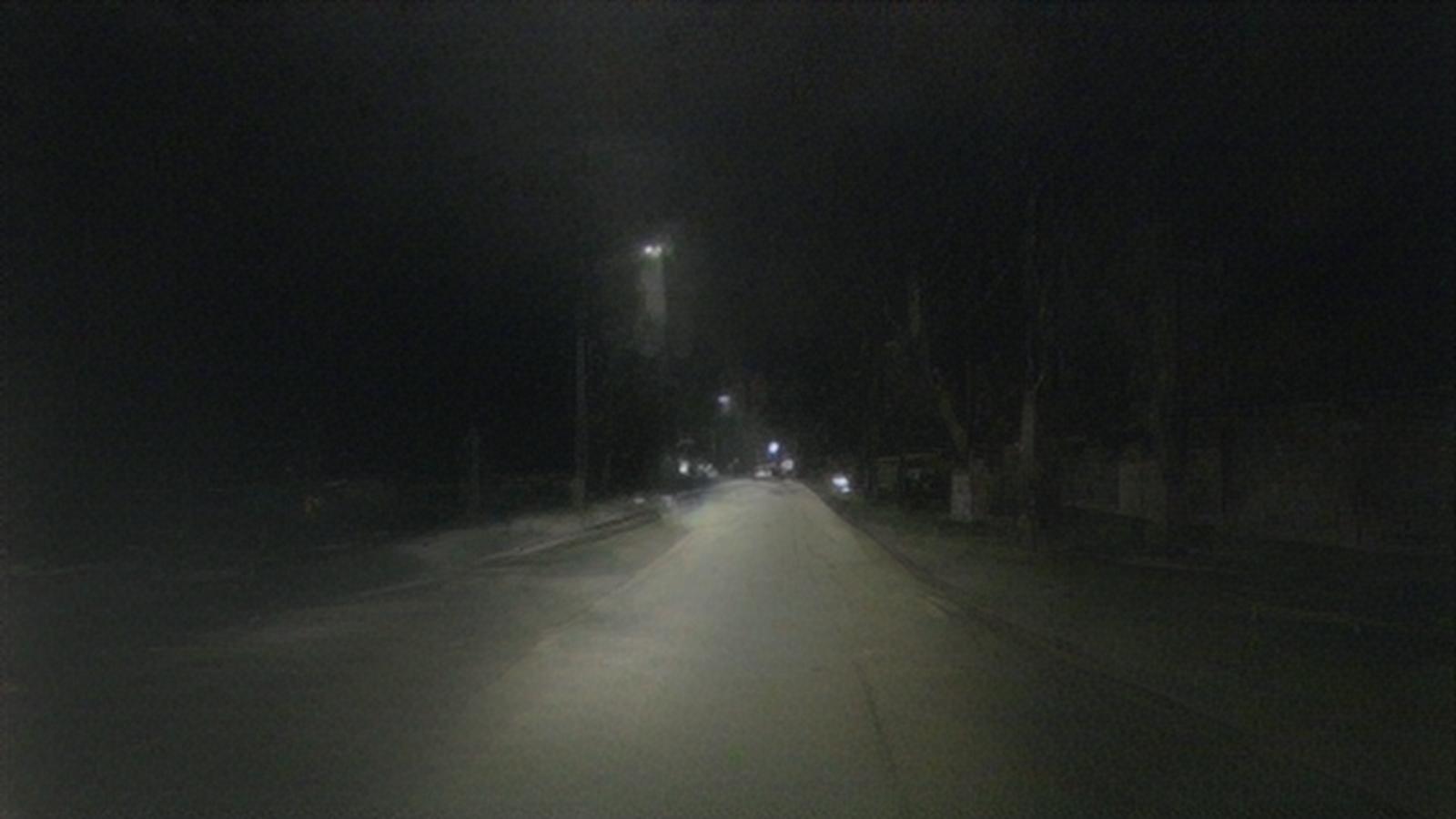} &
  \includegraphics[width=.16\textwidth]{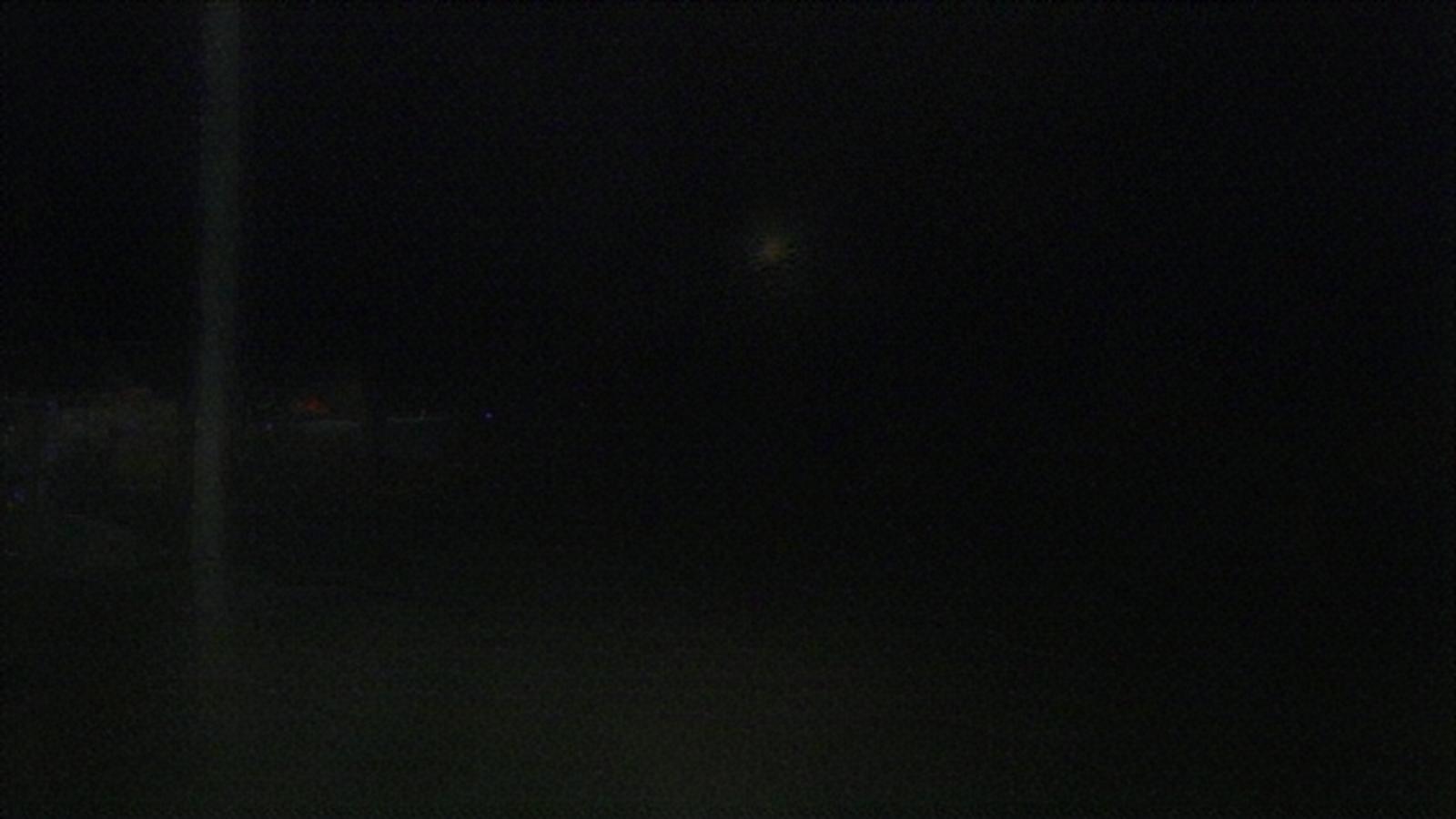}
  \vspace{-5pt}\\
  \rotatebox{90}{\makebox[0.8cm][r]{\tiny{Ours}}} &
\includegraphics[width=.16\textwidth]{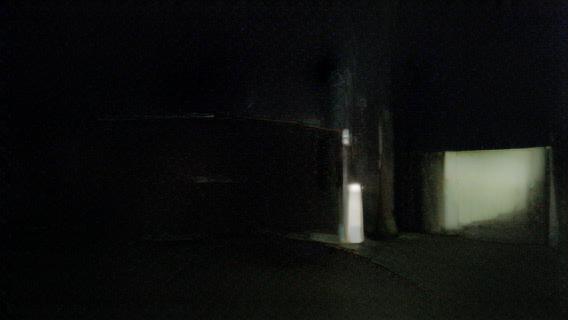} &
  \includegraphics[width=.16\textwidth]{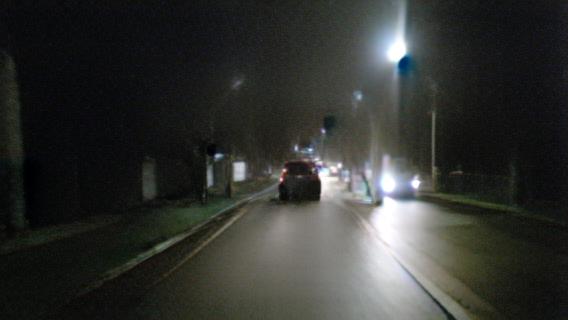} &
  \includegraphics[width=.16\textwidth]{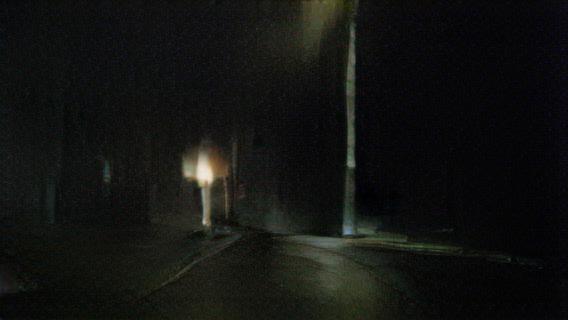}&
  \includegraphics[width=.16\textwidth]{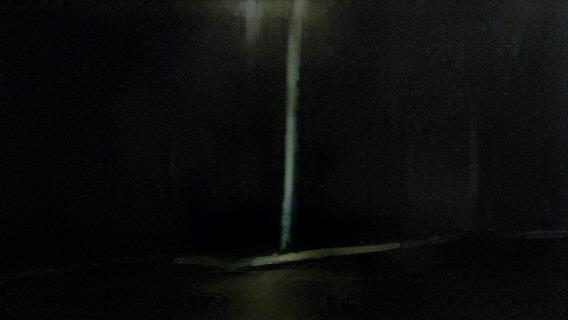} &
   \includegraphics[width=.16\textwidth]{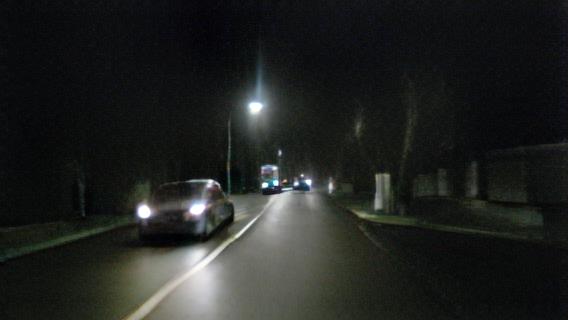} &
  \includegraphics[width=.16\textwidth]{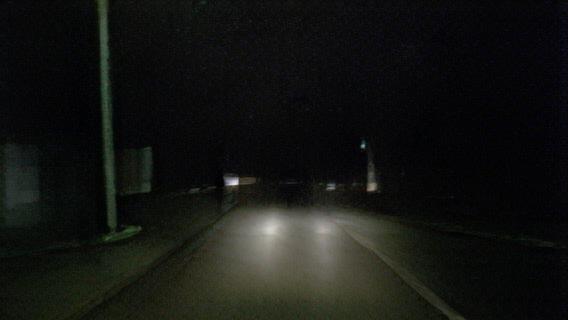}
 \vspace{-5pt} \\
\end{tabular}
\\
 \begin{tabular}{@{}c@{\hspace{2pt}}c@{}c@{}c@{}c@{}c@{}c@{}}
 %% Example 3
 \rotatebox{90}{\makebox[1.0cm][r]{\tiny{OmniRe}}} &
 \includegraphics[width=.16\textwidth]{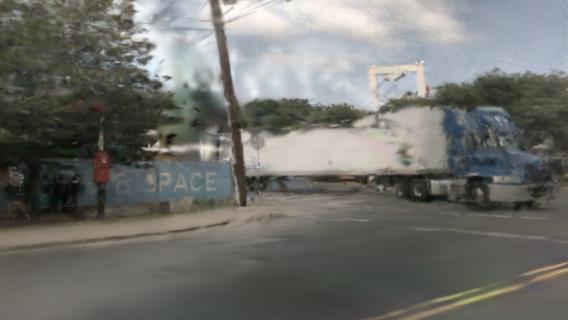} &
  \includegraphics[width=.16\textwidth]{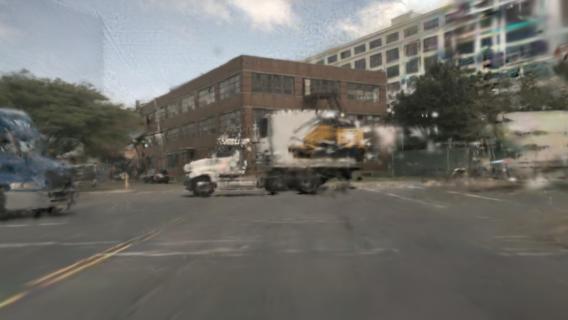} &
 \includegraphics[width=.16\textwidth]{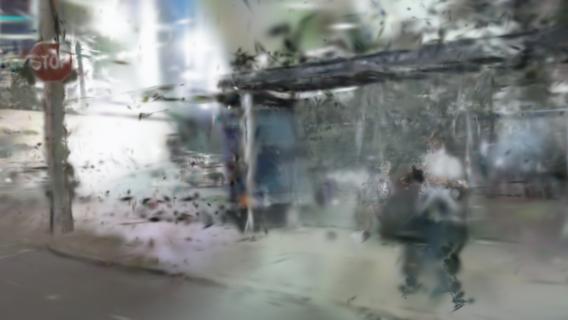} &
  \includegraphics[width=.16\textwidth]{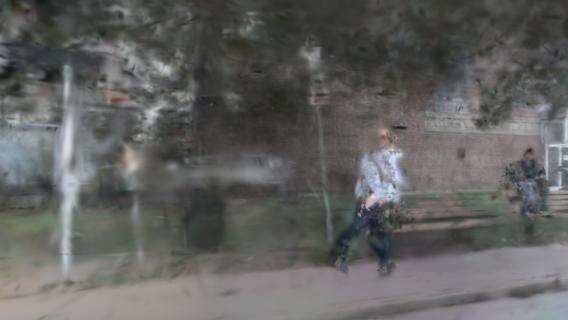}&
  \includegraphics[width=.16\textwidth]{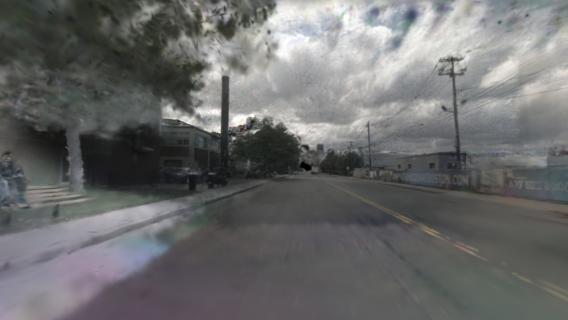} &
  \includegraphics[width=.16\textwidth]{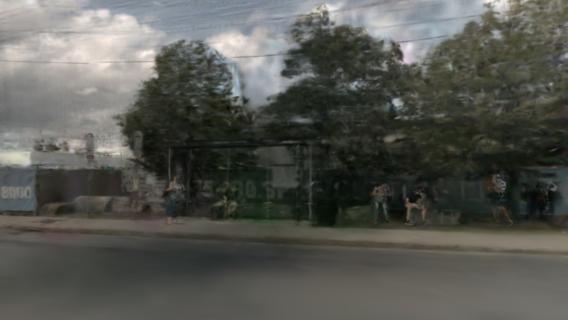}
  \vspace{-5pt}\\
  \rotatebox{90}{\makebox[1.0cm][r]{\tiny{DiFix3D}}} &
 \includegraphics[width=.16\textwidth]{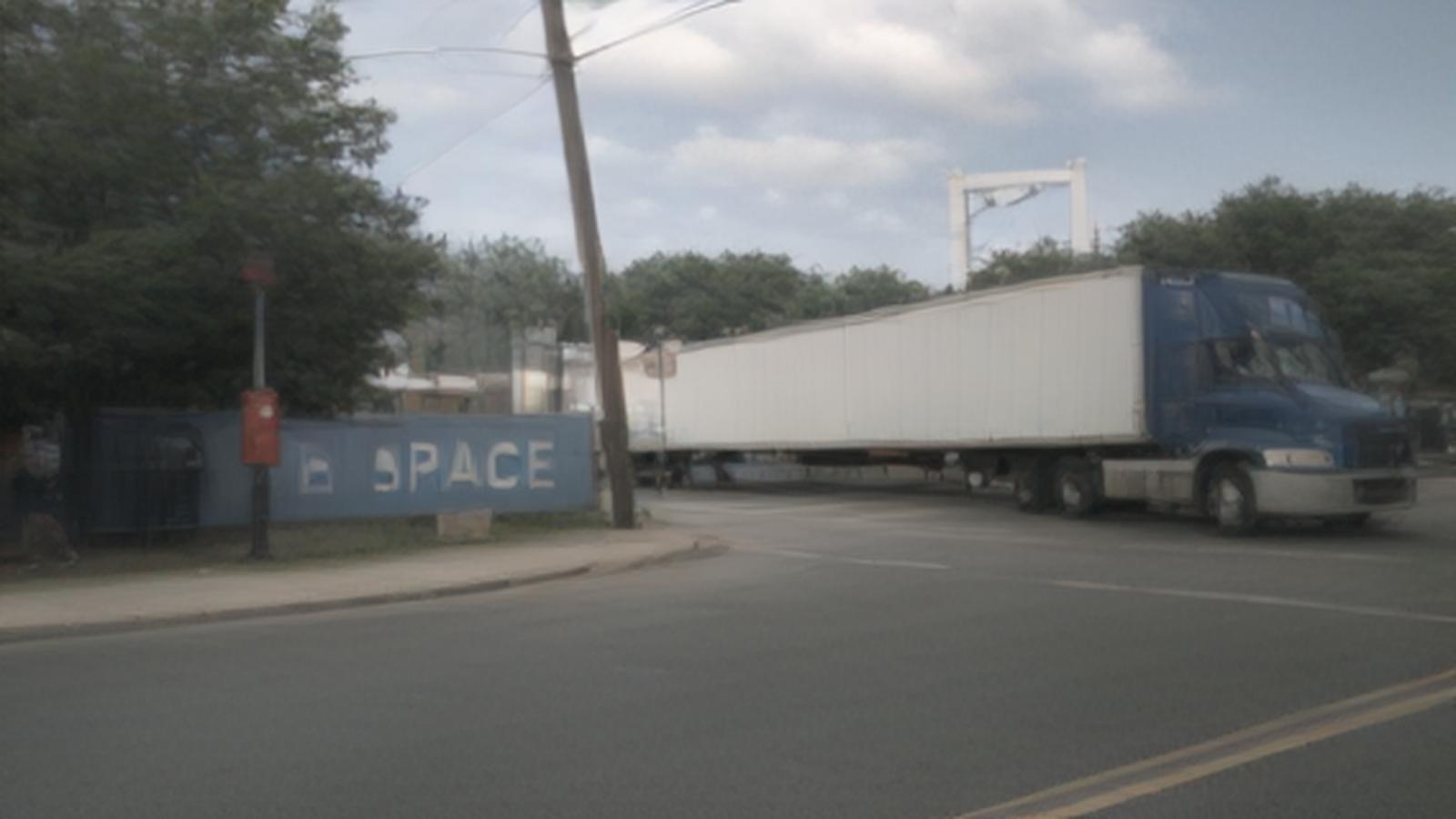} &
  \includegraphics[width=.16\textwidth]{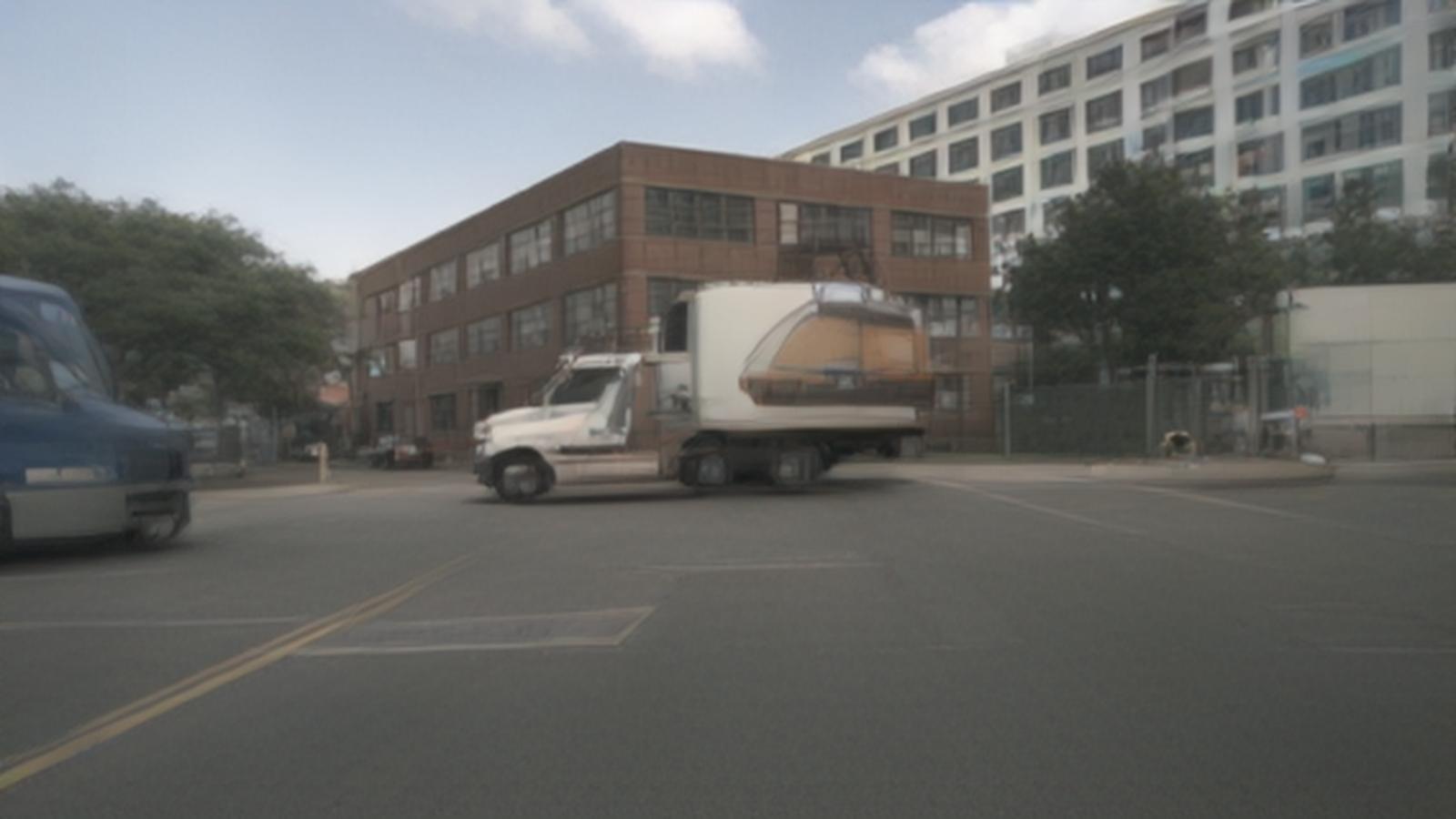} &
 \includegraphics[width=.16\textwidth]{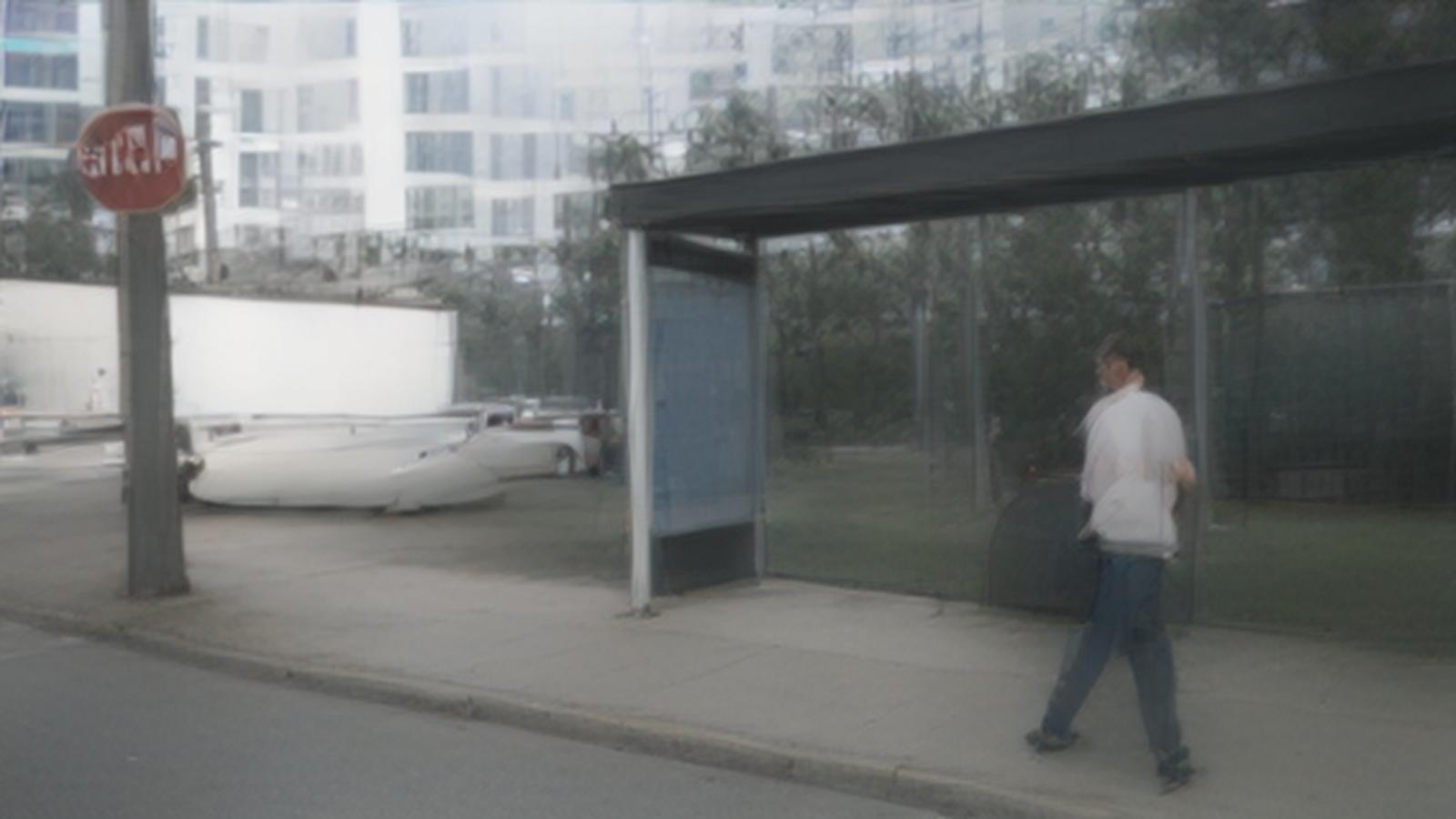} &
  \includegraphics[width=.16\textwidth]{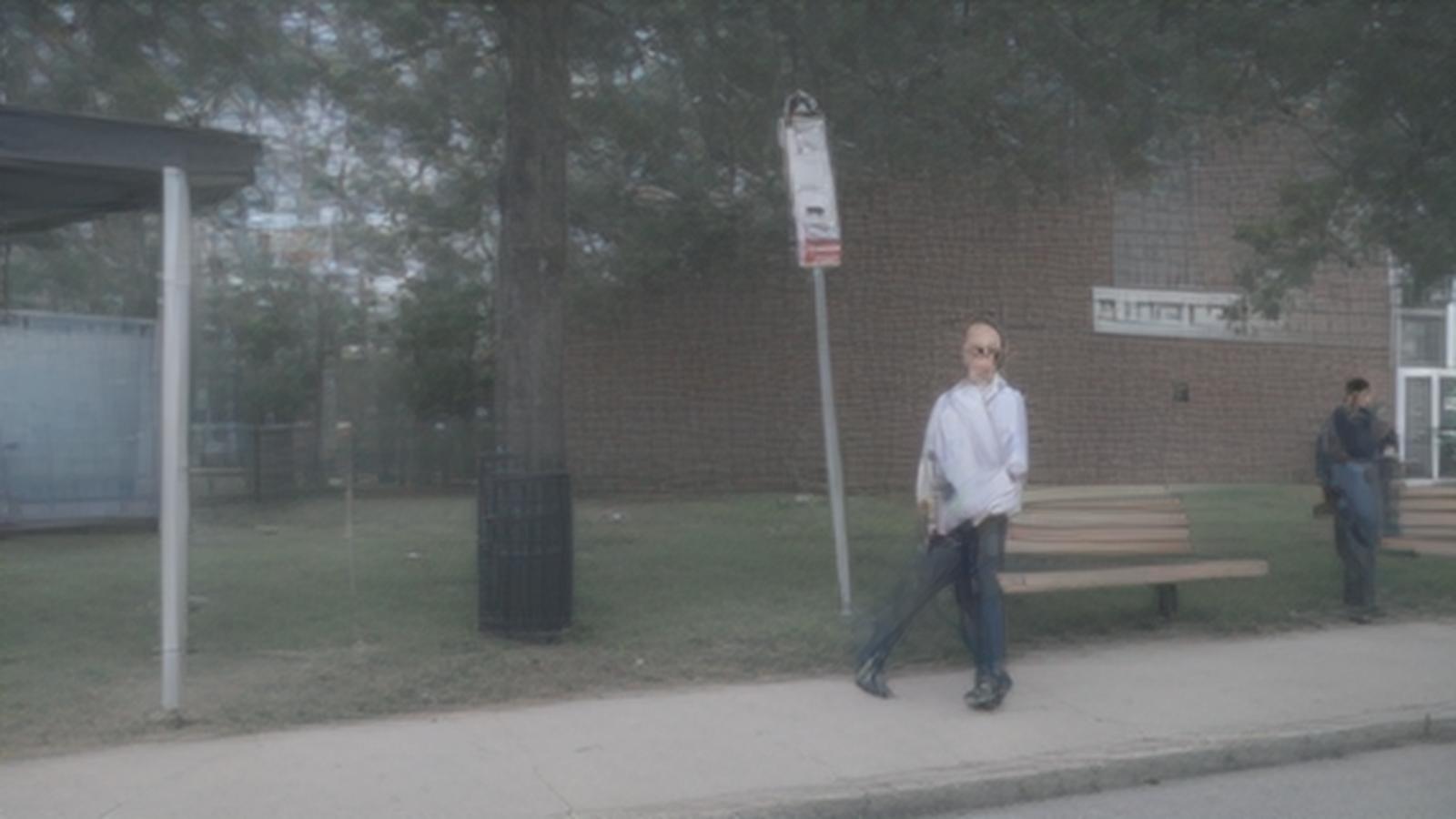}&
  \includegraphics[width=.16\textwidth]{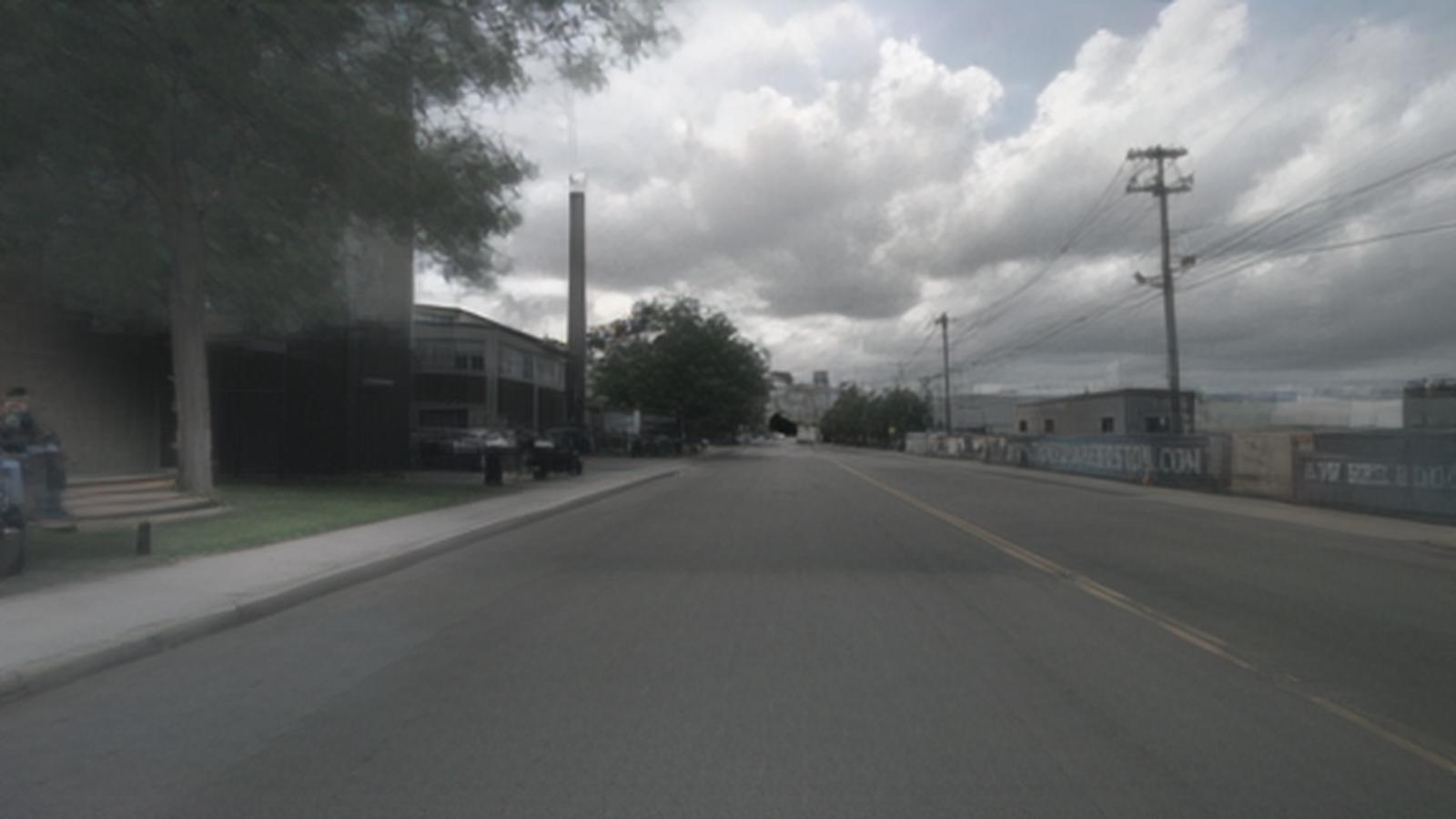} &
  \includegraphics[width=.16\textwidth]{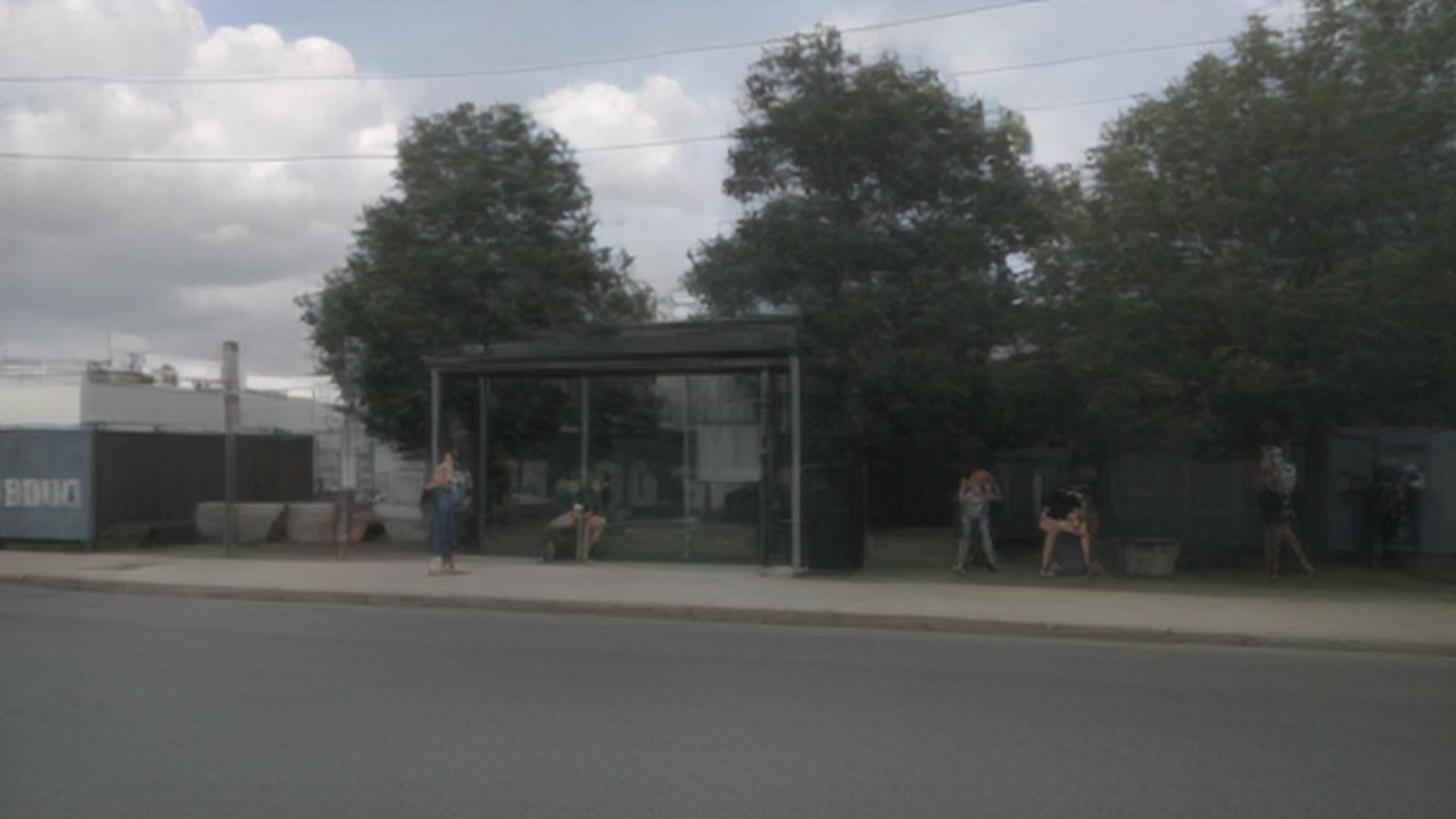}
  \vspace{-5pt}\\
  \rotatebox{90}{\makebox[0.7cm][r]{\tiny{Ours}}} &
 \includegraphics[width=.16\textwidth]{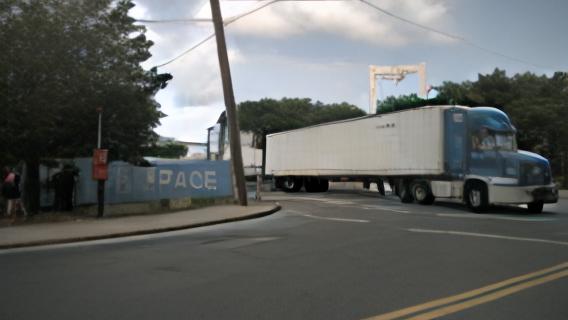} &
  \includegraphics[width=.16\textwidth]{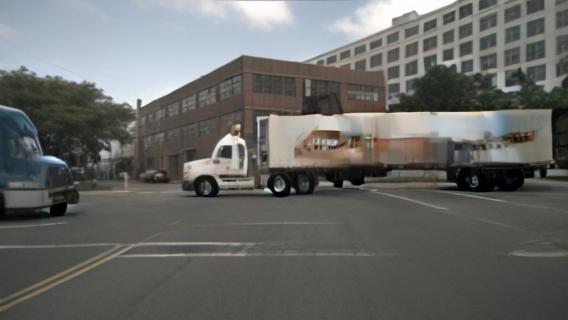} &
 \includegraphics[width=.16\textwidth]{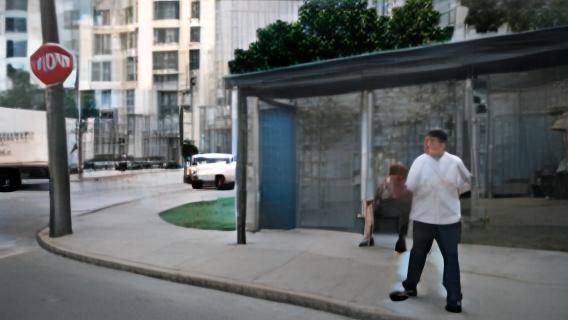} &
  \includegraphics[width=.16\textwidth]{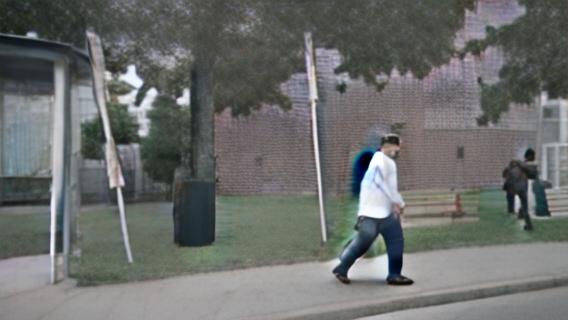}&
  \includegraphics[width=.16\textwidth]{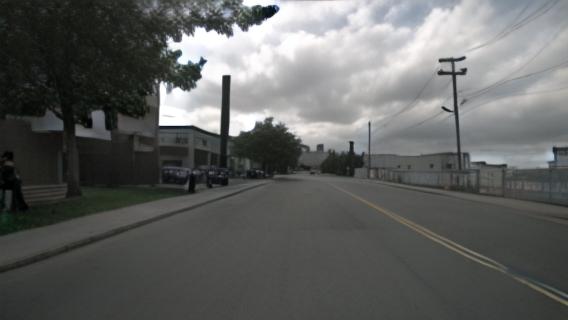} &
  \includegraphics[width=.16\textwidth]{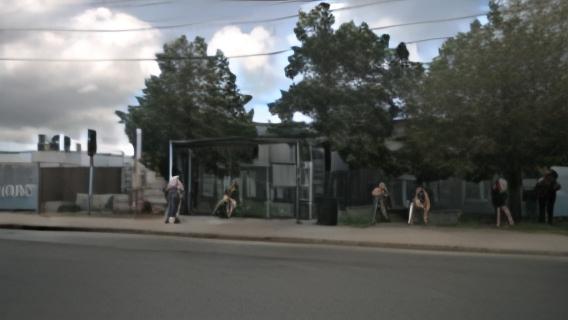}
 \vspace{-5pt}\\
\end{tabular}
\\
\end{tabular}
}
\caption{
\textbf{Multi-camera enhancement} of extrapolated views from nuScenes.
}
\label{fig:nuscenes_multi_cam}
\vspace{-5pt}
\end{figure*}

\noindent\textbf{DL3DV results}
\Cref{tab:dl3dv_table} reports performance on DL3DV under three sparse-view configurations. For diffusion-based enhancement models—including DiFix, 3DGS-Enhancer, DiFix++, and the proposed \name—the 3DGS-rendered novel views are used as input for enhancement. All diffusion-based methods outperform the raw 3DGS outputs, with \name achieving the highest scores across all settings, surpassing DiFix3D and 3DGS-Enhancer. Notably, 3DGS-Enhancer incorporates iterative feedback by reinserting enhanced views into the 3DGS optimization loop, whereas the other methods—including ours—operate purely as post-processors without modifying the underlying geometry. This highlights the effectiveness of \name in improving perceptual quality without requiring retraining, iterative refinement, or geometry updates.
\begin{table*}[!h]
\centering
\setlength{\tabcolsep}{8pt} % default is 6pt
\renewcommand{\arraystretch}{1.1} % row spacing (default = 1.0)

\resizebox{\textwidth}{!}{
\begin{tabular}{lccc|ccc|ccc}
\toprule
\multirow{2}{*}{\textbf{Method}} & \multicolumn{3}{c|}{\textbf{3 views}} & \multicolumn{3}{c|}{\textbf{6 views}} & \multicolumn{3}{c}{\textbf{9 views}} \\
\cmidrule(lr){2-4} \cmidrule(lr){5-7} \cmidrule(lr){8-10}
& \textbf{PSNR} & \textbf{SSIM} & \textbf{LPIPS} & \textbf{PSNR} & \textbf{SSIM} & \textbf{LPIPS} & \textbf{PSNR} & \textbf{SSIM} & \textbf{LPIPS} \\
\midrule
3DGS~\cite{kerbl2023gaussian} & 10.97 & 0.248 & 0.567 & 13.34 & 0.332 & 0.498 & 14.99 & 0.403 & 0.446 \\
Mip-NeRF~\cite{barron2021mipnerf} & 10.92 & 0.191 & 0.618 & 11.56 & 0.199 & 0.608 & 12.42 & 0.218 & 0.600 \\
RegNeRF~\cite{niemeyer2022regnerf} & 11.46 & 0.214 & 0.600 & 12.69 & 0.236 & 0.579 & 12.33 & 0.219 & 0.598 \\
FreeNeRF~\cite{yu2021freenerf} & 10.91 & 0.211 & 0.595 & 12.13 & 0.230 & 0.576 & 12.85 & 0.241 & 0.573 \\
DNGaussian~\cite{wu2023dngaussian} & 11.10 & 0.273 & 0.579 & 12.67 & 0.329 & 0.547 & 13.44 & 0.365 & 0.539 \\
\midrule
3DGS~\cite{kerbl2023gaussian} & 10.97 & 0.248 & 0.567 & 13.34 & 0.332 & 0.498 & 14.99 & 0.403 & 0.446 \\
\quad + DIFIX3D~\cite{wu2025difix3d} & 14.90 & 0.460 & 0.338 & 16.90 & 0.527 & 0.266 & 17.86 & 0.560 & 0.235 \\
\quad + 3DGS-Enhancer~\cite{liu20243dgs} & 14.33 & 0.424 & 0.464 & 16.94 & 0.565 & 0.356 & 18.50 & \textbf{0.630} & 0.305 \\
\quad + DIFIX3D++~\cite{omran2025hybrid} & 15.93 & 0.528 & 0.361 & 17.95 & \textbf{0.589} & 0.291 & 18.93 & 0.619 & 0.262 \\
\rowcolor{lightblue} \quad + \name & \textbf{16.87} & \textbf{0.533} & \textbf{0.335} & \textbf{18.74} & 0.587 & \textbf{0.264} & \textbf{19.60} & 0.614 & \textbf{0.232} \\
\bottomrule
\end{tabular}
}
%\vspace{-8pt}
\caption{\textbf{SOTA comparisons on DL3DV} on few-shot 3D reconstruction following~\cite{liu20243dgs}. Neural reconstruction (top) and diffusion enhancers (bottom) comparisons are reported separately. All diffusion enhancers are applied on the same 3DGS backbone.}.
\label{tab:dl3dv_table}
\end{table*}
\vspace{-9pt}
\begin{figure*}[ht!]
\centering
\resizebox{\textwidth}{!}{
\begin{tabular}{@{}c@{\hspace{-6pt}}@{}c@{\hspace{-6pt}}@{}c@{\hspace{-6pt}}@{}c@{\hspace{-6pt}}@{}c@{\hspace{-6pt}}@{}}
\begin{tabular}{@{}c@{\hspace{3pt}}@{}c@{}}
%% Example 1
\rotatebox{90}{\makebox[1.0cm][r]{\tiny{3DGS}}} &
\includegraphics[width=.2\textwidth]{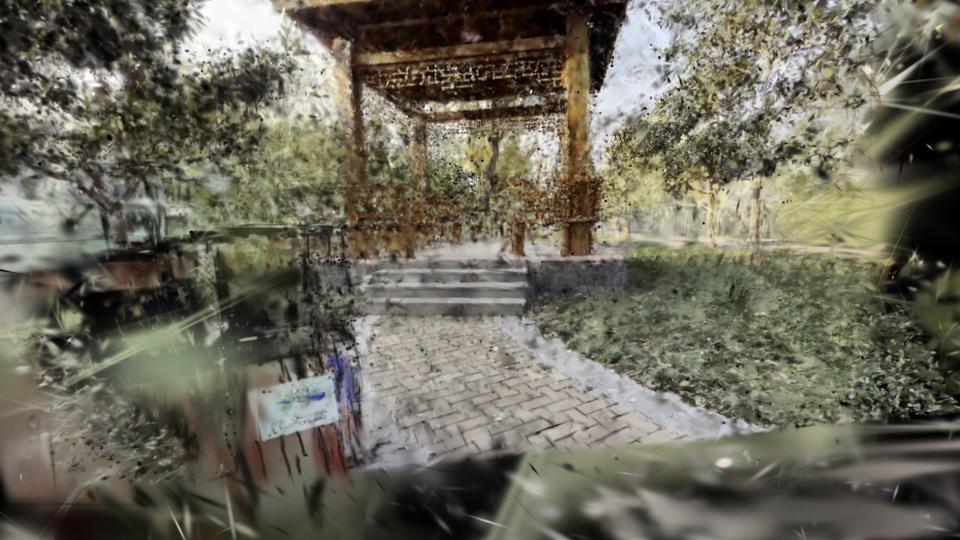}
\vspace{-3pt}\\
\rotatebox{90}{\makebox[1.2cm][r]{\tiny{DiFix3D}}} &
\includegraphics[width=.2\textwidth]{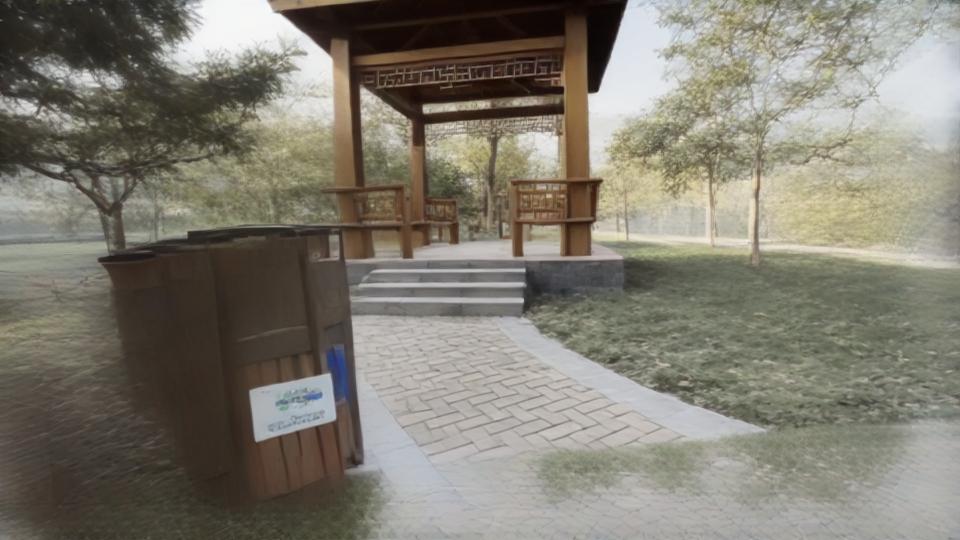}
\vspace{-3pt}\\
\rotatebox{90}{\makebox[1.0cm][r]{\tiny{Ours}}} &
\includegraphics[width=.2\textwidth]{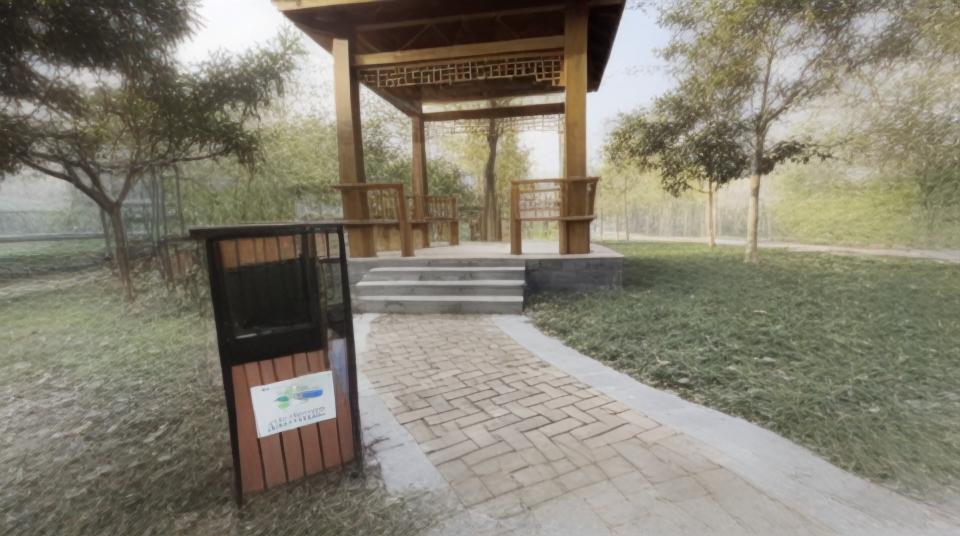}
\vspace{-3pt}\\
\rotatebox{90}{\makebox[1.0cm][r]{\tiny{GT}}} &\hspace{-6pt}
\includegraphics[width=.2\textwidth]{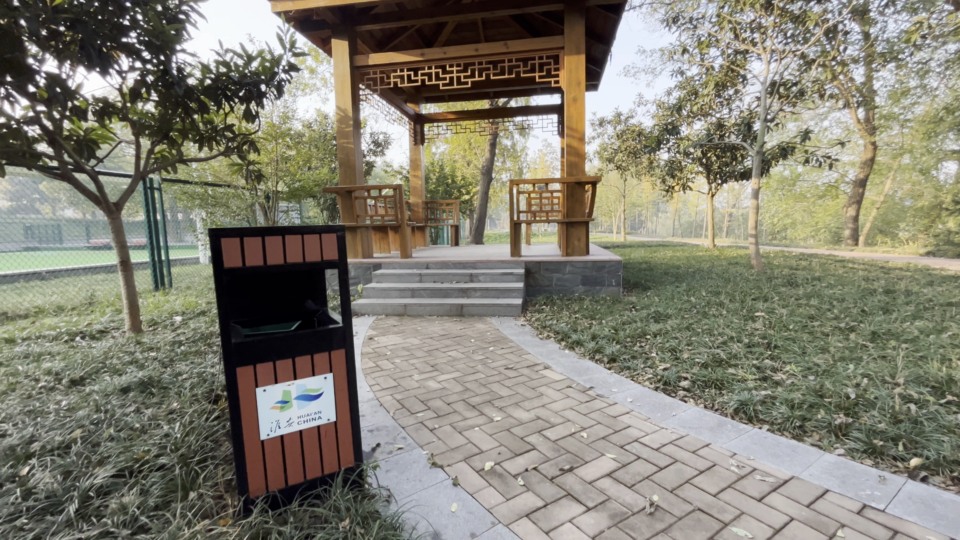} \\
\end{tabular}
 &
\begin{tabular}{@{}c@{}}
%% Example 2
\includegraphics[width=.2\textwidth]{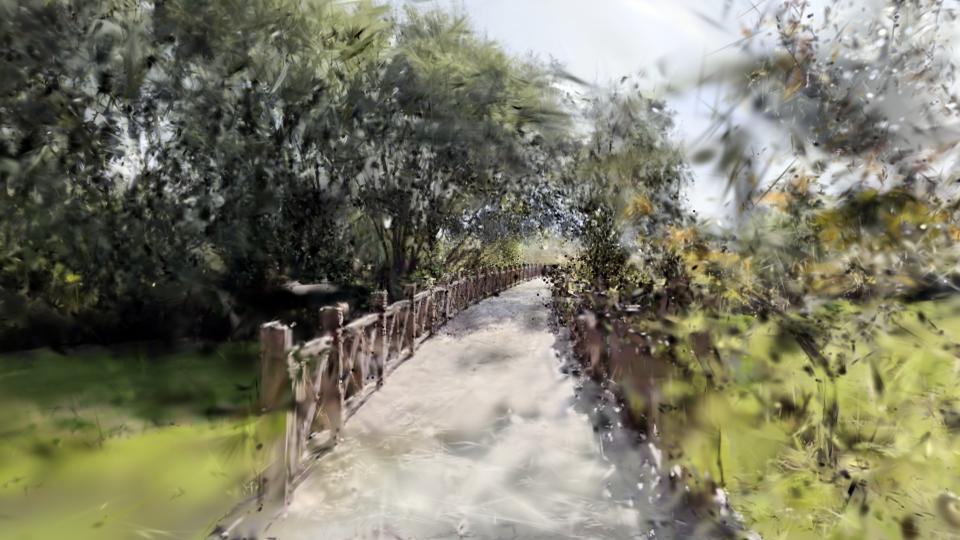}
\vspace{-3pt}\\
\includegraphics[width=.2\textwidth]{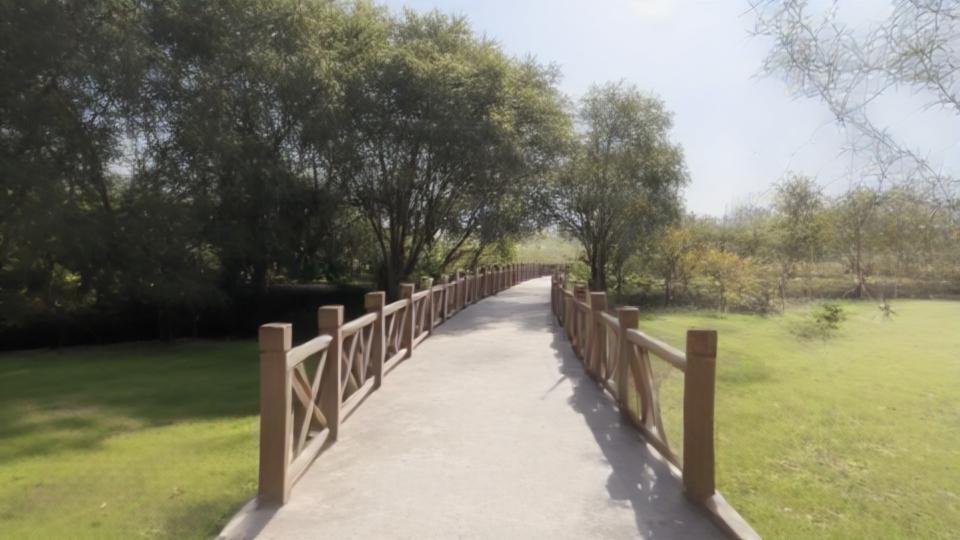}
\vspace{-3pt}\\
\includegraphics[width=.2\textwidth]{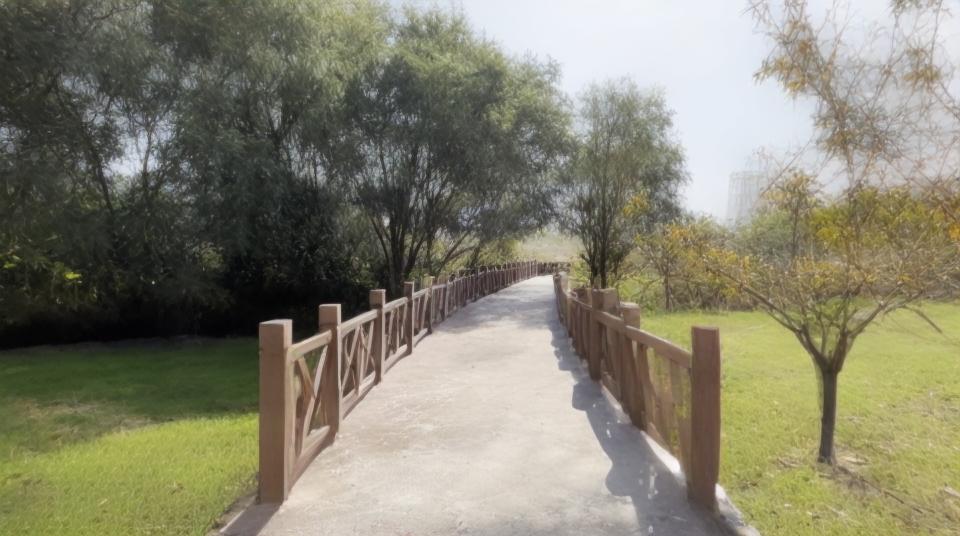}
\vspace{-3pt}\\
\hspace{-6pt}
\includegraphics[width=.2\textwidth]{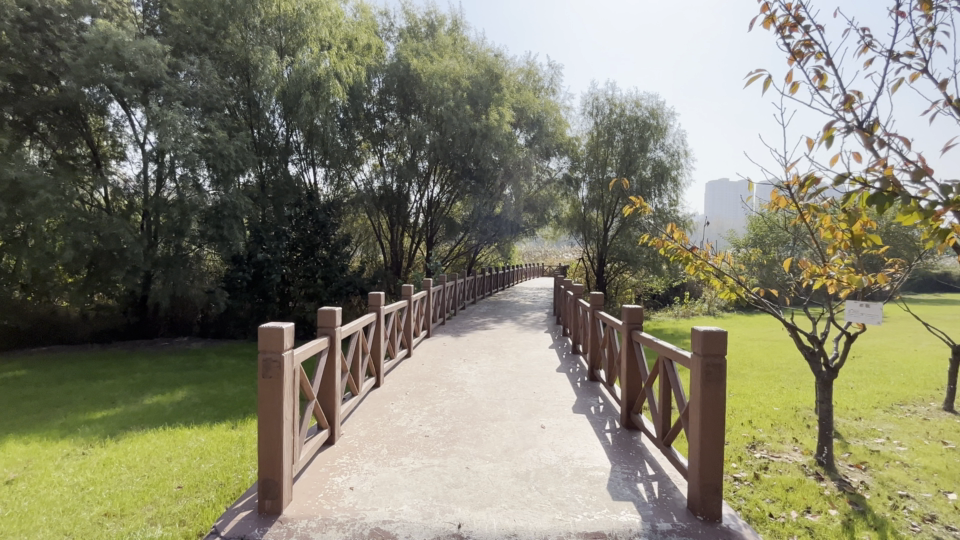} \\
\end{tabular} &
\begin{tabular}{@{}c@{}}
%% Example 2
\includegraphics[width=.2\textwidth]{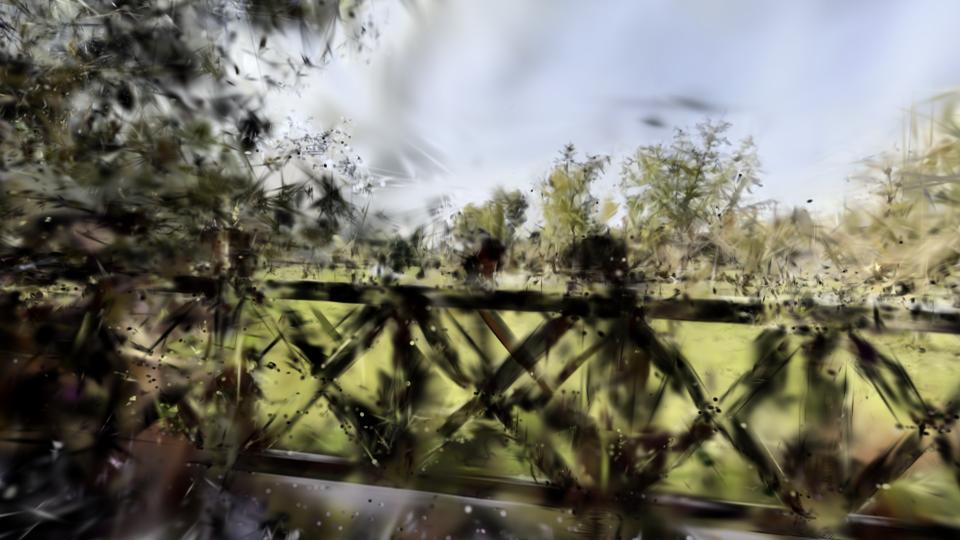}
\vspace{-3pt}\\
\includegraphics[width=.2\textwidth]{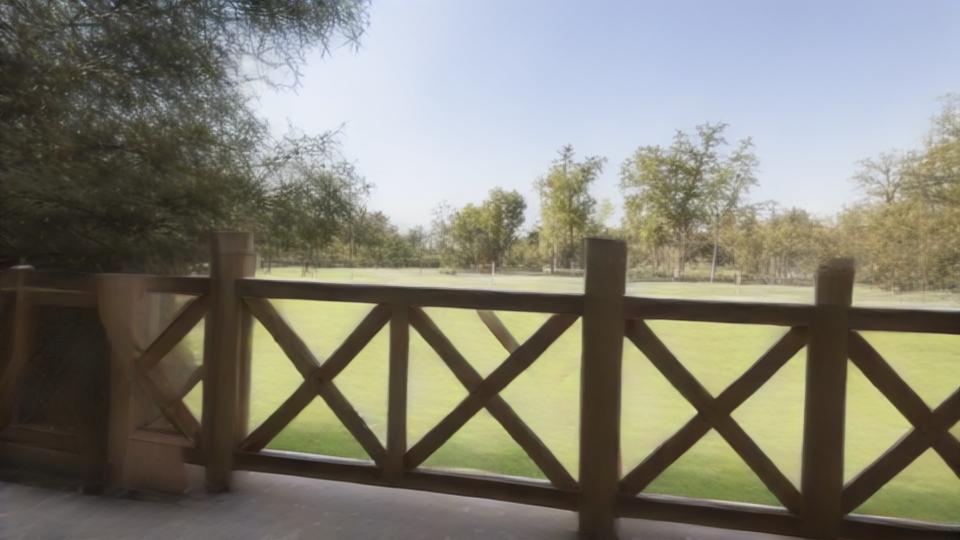}
\vspace{-3pt}\\
\includegraphics[width=.2\textwidth]{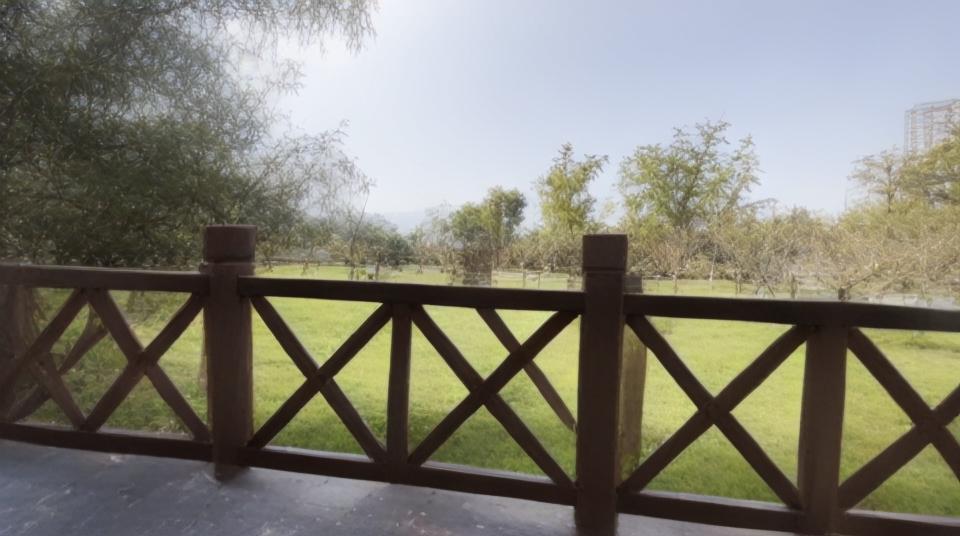}
\vspace{-3pt}\\
\hspace{-6pt}
\includegraphics[width=.2\textwidth]{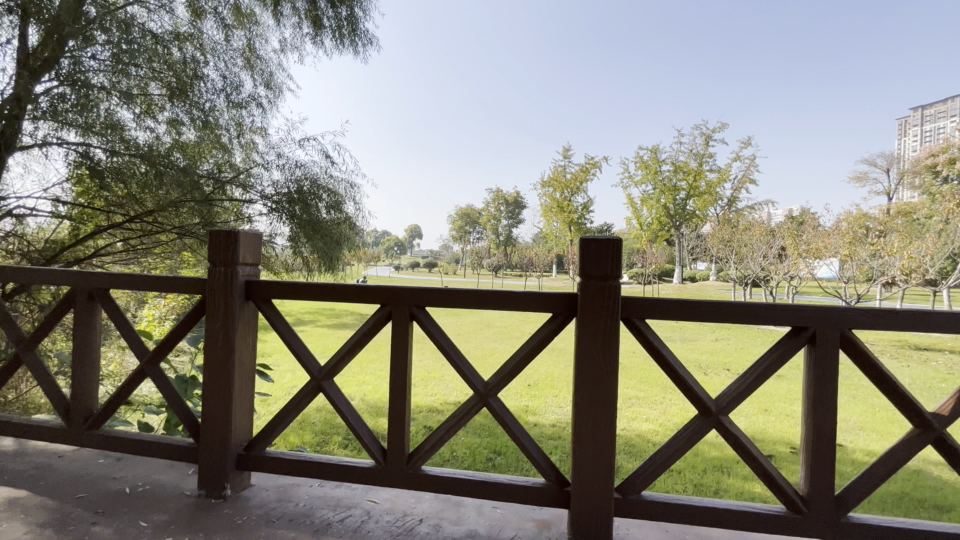} \\
\end{tabular} &
\begin{tabular}{@{}c@{}}
%% Example 2
\includegraphics[width=.2\textwidth]{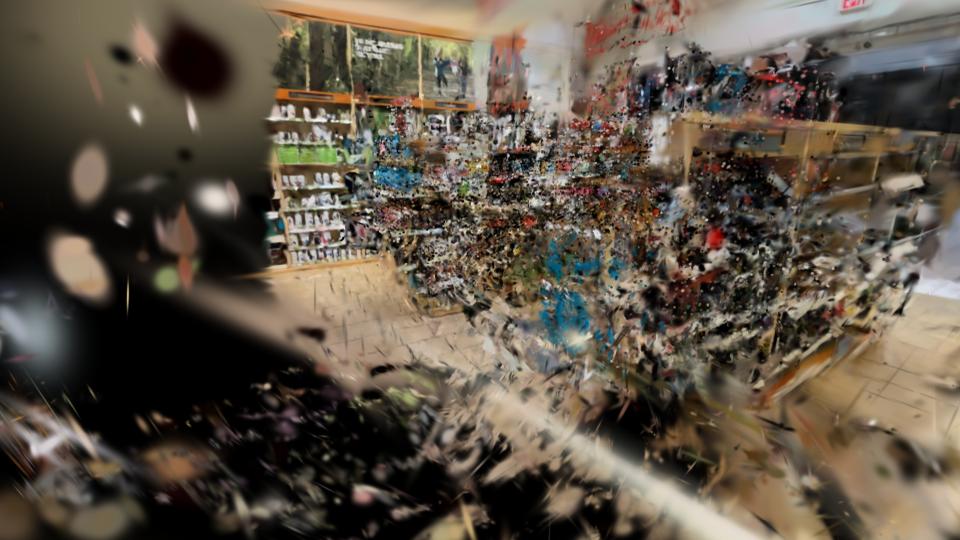}
\vspace{-3pt}\\
\includegraphics[width=.2\textwidth]{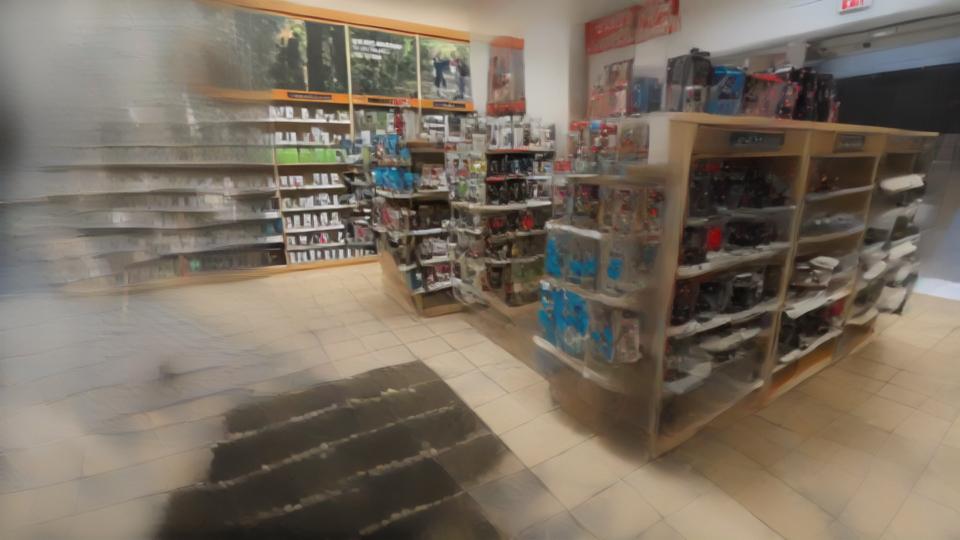}
\vspace{-3pt}\\
\includegraphics[width=.2\textwidth]{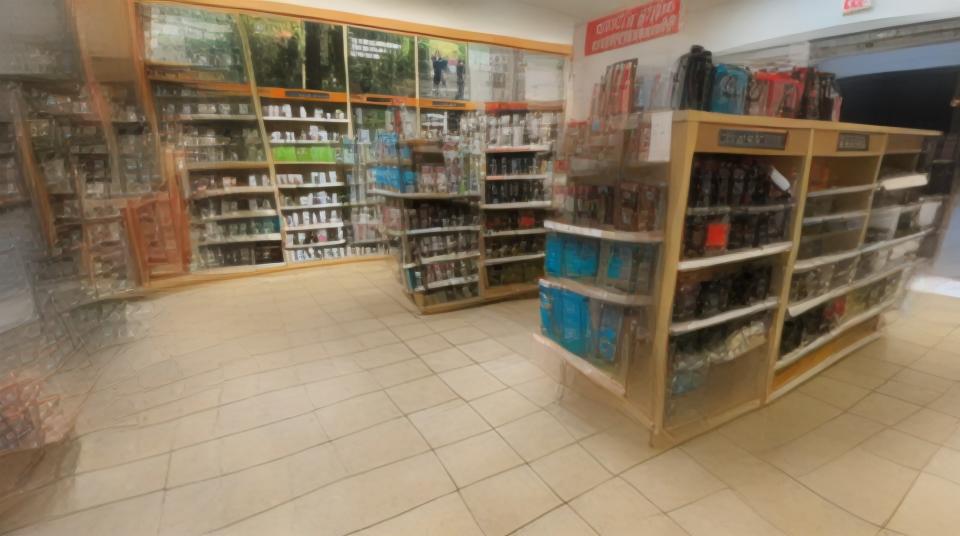}
\vspace{-3pt}\\
\hspace{-6pt}
\includegraphics[width=.2\textwidth]{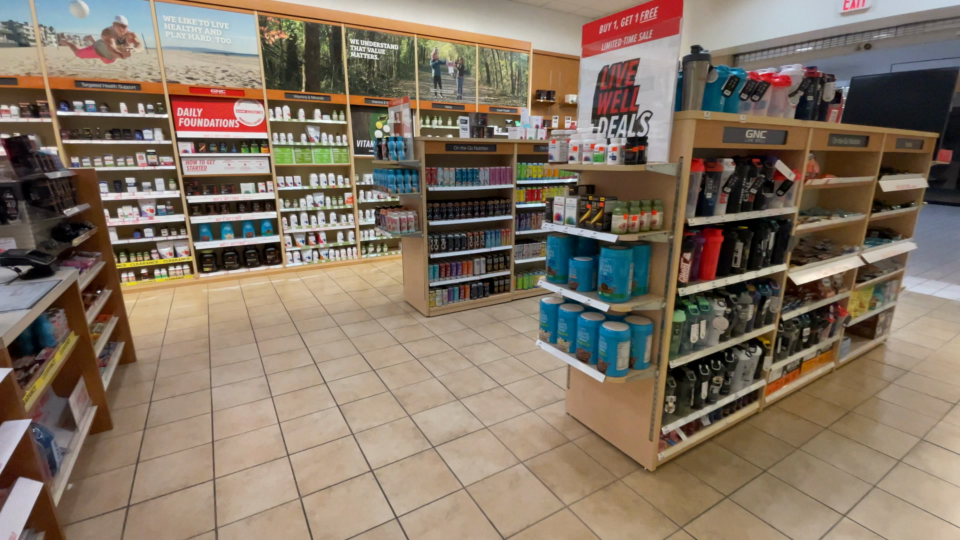} \\
\end{tabular} &
\begin{tabular}{@{}c@{}}
%% Example 5
\includegraphics[width=.2\textwidth]{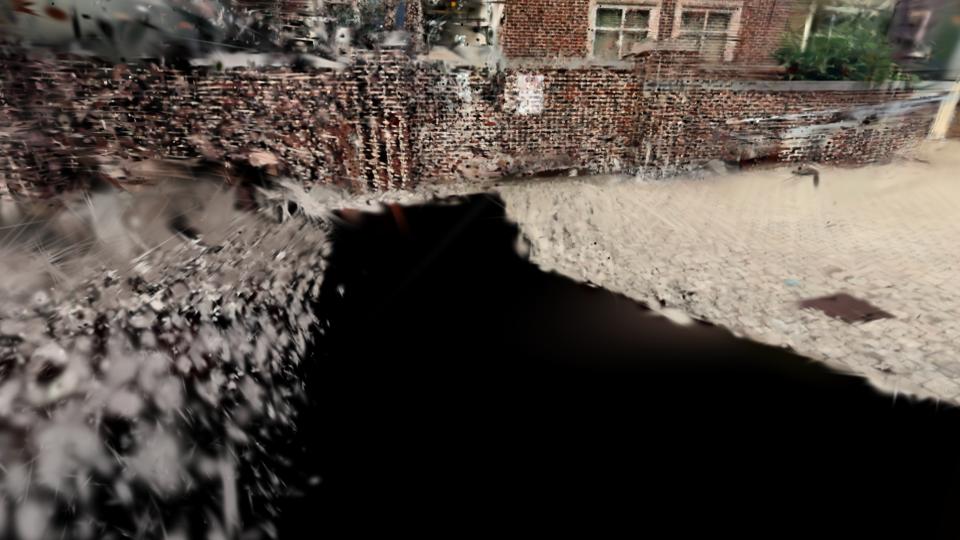}
\vspace{-3pt}\\
\includegraphics[width=.2\textwidth]{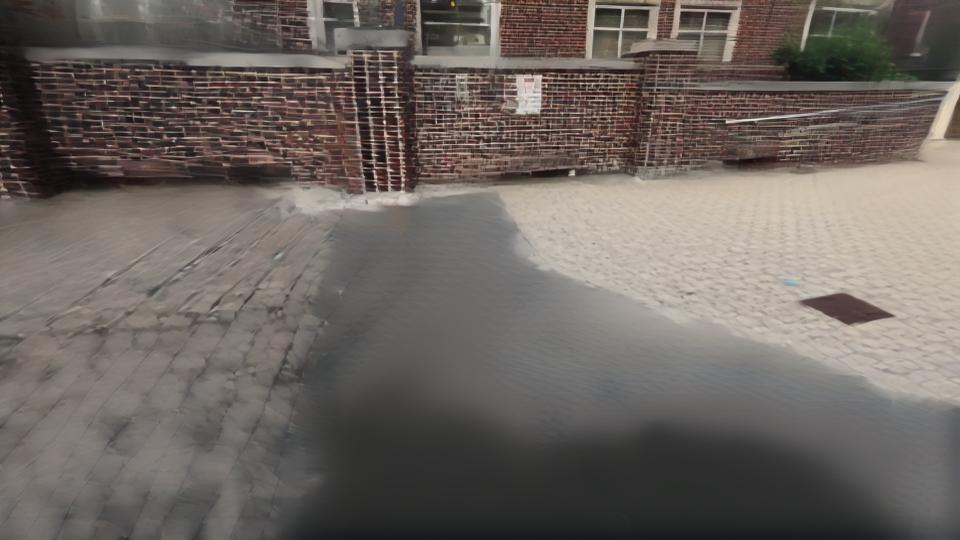}
\vspace{-3pt}\\
\includegraphics[width=.2\textwidth]{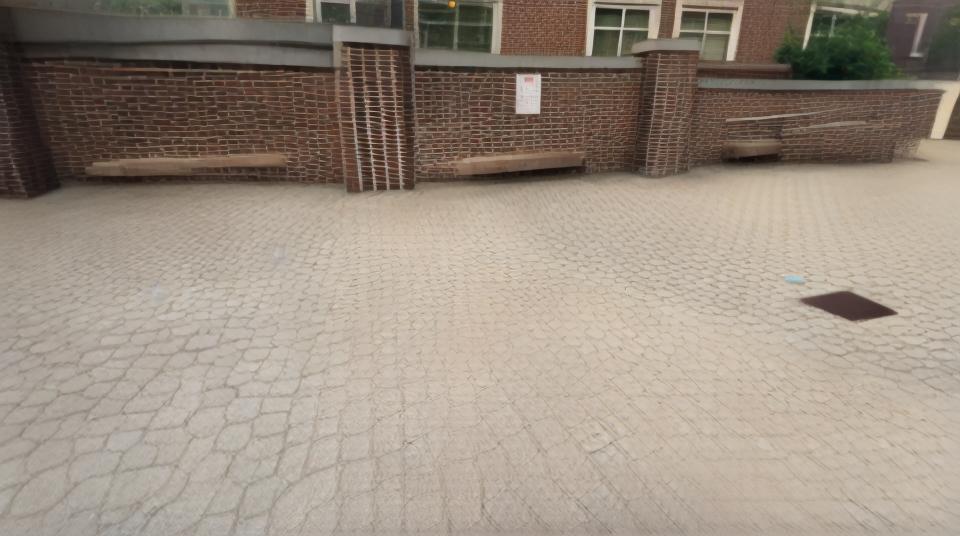}
\vspace{-3pt}\\
\hspace{-6pt}
\includegraphics[width=.2\textwidth]{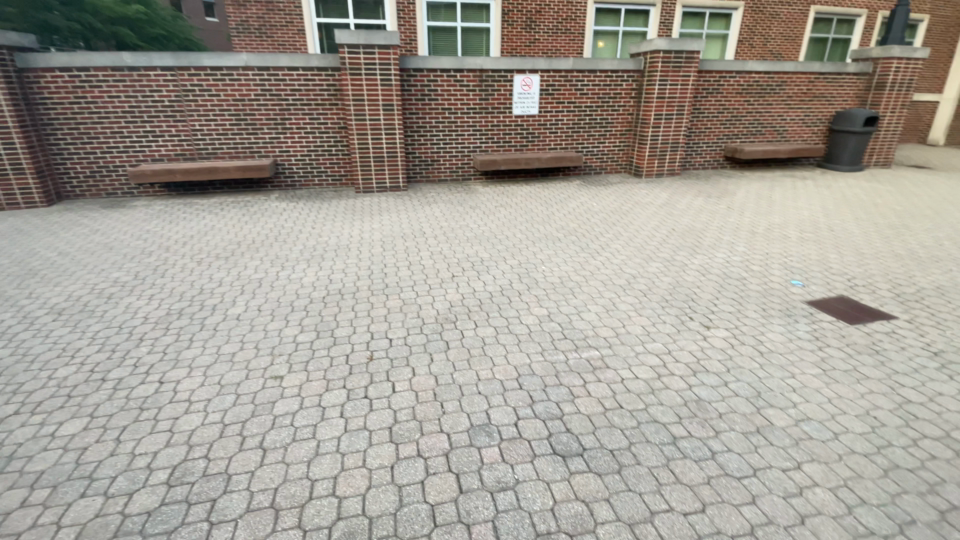} \\
\end{tabular} 
\\
\end{tabular}
  }
  \vspace{-5pt}
  \caption{
     \textbf{Novel-view images on DL3DV} rendered from sparse-view 3DGS.
  }
   \label{fig:dl3dv_examples}   
\end{figure*}
%\vspace{-5pt}

\subsection{Ablation Study}
\begin{table}[!t]
\vspace{-8pt}
\centering
\caption{Ablating the impact of geometric grounding on enhancement qualities.
}
\setlength{\tabcolsep}{24pt} % default is 6pt
\renewcommand{\arraystretch}{1.1} % row spacing (default = 1.0)

\resizebox{.7\textwidth}{!}{
\begin{tabular}[t]{lccc}
\toprule
\textbf{Geometric Grounding} & \textbf{PSNR} & \textbf{SSIM} & \textbf{LPIPS} \\
\midrule

None & 17.59 & 0.164 & 0.278 \\
Coordinate maps & 18.39 & 0.557 & 0.282 \\
Camera poses & 18.43 & 0.560 & 0.288 \\
\cellcolor{lightblue}Both & \cellcolor{lightblue}\textbf{18.74} & \cellcolor{lightblue}\textbf{0.587} & \cellcolor{lightblue}\textbf{0.265} \\
% \multirow{2}{*}{Architecture} & w/o LoRA & 18.604 & 0.584 & 0.287\\
% & cross-attn & 18.673 & 0.587 & 0.268 \\
\bottomrule
\end{tabular}
}
\label{tab:ablation_geo}
\vspace{-4pt}
\end{table}

To evaluate the contributions of the proposed geometric grounding and set-based diffusion enhancement, we conduct a series of ablation experiments on DL3DV dataset. Additional ablations are provided in~\Cref{sec:sup_ablation}.

\noindent\textbf{Geometric grounding}
We assess the influence of the proposed geometric conditionings to ground the diffusion enhancer by selectively including camera C-maps, camera poses, or both. As shown in \Cref{tab:ablation_geo}, using no grounding results in the lowest performance, confirming the importance of explicit geometric priors. Conditioning on either camera poses or C-maps individually yields comparable improvements, indicating that each provides useful geometric grounding. Their combination (“both”) achieves the best overall performance, demonstrating they be both required and complementary.

\noindent\textbf{Set diffusion}
\begin{table}[b!]
\centering
\caption{Ablating the impact of set diffusion on enhancement qualities.}

\setlength{\tabcolsep}{24pt}
\renewcommand{\arraystretch}{1.15}

\resizebox{0.9\textwidth}{!}{
\begin{tabular}[t]{lcccccc}
\toprule
$\boldsymbol{\mathcal{I}}^r + \boldsymbol{\mathcal{I}}^t$
& \textbf{1 + 1} & \textbf{2 + 1} & \textbf{3 + 1} & \textbf{3 + 2} & \textbf{3 + 4} & \textbf{3 + 6} \\
\midrule
\textbf{PSNR}  & 17.80 & 18.57 & 18.67 & 18.69 & 18.72 &\textbf{18.79} \\
\textbf{SSIM}  & 0.552 & 0.578 & 0.581 & 0.582 & 0.583 &\textbf{0.590} \\
\textbf{LPIPS} & 0.280 & 0.258 & \textbf{0.255} & 0.255 & 0.255 &0.256 \\
\bottomrule
\end{tabular}
}
\label{tab:ablation_set}
\vspace{-8pt}
\end{table}
We evaluate the impact of using set-based reference and target inputs, denoted by $\mathcal{I}^r$ and $\mathcal{I}^t$, on enhancement quality. Starting from single-frame enhancement  ($|\mathcal{I}^r|=|\mathcal{I}^t|=1$), we progressively increase the sizes of the reference and target sets and analyze their effect on diffusion-based enhancement performance, as reported in \Cref{tab:ablation_set}. The results show that enlarging the reference set leads to a substantial performance gain, as it enables the model to aggregate missing information across multiple views -an essential factor for accurate reconstruction. Performance gains begin to saturate beyond three reference views, which appears sufficient to cover the DL3DV dataset. Increasing the size of the target set also improves enhancement quality, albeit to a lesser extent, by exploiting cross-frame correlations.

\subsection{Runtime Analysis}
We measure average enhancement time per $578 \times 1024$ frame on an A100–80GB GPU as $66.6$,ms for OmniRe, $800.0$,ms for DiFix, and $555.5$,ms for \name. Using the single-step SD‑Turbo denoiser, DiFix and \name reach real-time throughput of $1.2$ and $1.8$ FPS, enabling integration into simulation frameworks such as AlpaSim~\cite{alpasim_2025}.
Despite jointly processing multiple views, \name is more efficient than single-view DiFix due to parallelizable multi-view batching that amortizes computation with minimal overhead. The added attention from the extended multi-view token sequence contributes only a small portion of the total cost: the \name UNet backbone runs at 9 FPS, while the VAE decoder remains the bottleneck at 3.25 FPS. With single-step denoising, the denoiser contributes only a minor share of runtime, and the modest increase in attention tokens has negligible impact on total compute.

\subsection{Failure cases}
\name leverages visual information from clean reference views to guide enhancement when sufficient geometric overlap exists between reference and target views. When this overlap becomes very limited—due to extreme viewpoint shifts or heavy occlusions—the geometric conditioning becomes less reliable. In such cases, the model relies more heavily on its diffusion prior, yielding visually plausible results but potentially introducing misalignment or localized hallucinations. These situations are uncommon in autonomous driving scenarios, where novel viewpoints typically retain substantial visibility with recorded frames. Nonetheless, improving robustness under minimal geometric overlap remains an important direction for future work, for example through uncertainty-aware conditioning or adaptive weighting of geometric cues.

%\textbf{Inference-time Reference Views}: We test the effect of varying the number of reference views at inference time. Increasing the number of reference views from 1 to 6 does not significantly improve PSNR or SSIM, and only improves LPIPS. We hypothesize that in sparse-view setups of DL3DV, the selected single reference view closest to the target view dominates the contribution, while additional views offer limited overlap and thus minimal benefit to pixel-level reconstruction but provide additional context that improves the perceptual similarity.

\section{Conclusion}

We introduced \name, a geometry-grounded set-based diffusion framework to enhance novel-view renderings in 3D Gaussian Splatting. By conditioning the denoiser on pixel-aligned coordinate maps and pose-aware Plücker ray embeddings, \name improves photometric fidelity and reduces artifacts under sparse observations and wide-baseline extrapolation. Its permutation-equivariant set formulation enables joint processing of multiple reference and target views, yielding better image enhancement with minimal computational overhead. Across four challenging datasets, \name consistently outperforms existing diffusion-based enhancers and geometry-regularized baselines, establishing a robust and scalable solution for high-fidelity multi-camera view synthesis in autonomous driving scenarios. 
\clearpage
{
    \small
    \bibliographystyle{splncs04}
    \bibliography{main}

@String(PAMI = {IEEE Trans. Pattern Anal. Mach. Intell.})

@String(CVPR= {IEEE Conf. Comput. Vis. Pattern Recog.})

@String(ICCV= {Int. Conf. Comput. Vis.})

@String(ECCV= {Eur. Conf. Comput. Vis.})

@String(TOG= {ACM Trans. Graph.})

@String(PAMI  = {IEEE TPAMI})

@String(CVPR  = {CVPR})

@String(ICCV  = {ICCV})

@String(ECCV  = {ECCV})

@String(TOG   = {ACM TOG})

@inproceedings{mildenhall2020nerf,
  title={NeRF: Representing Scenes as Neural Radiance Fields for View Synthesis},
  author={Mildenhall, Ben and Tancik, Matthew and Srinivasan, Pratul P and Barron, Jonathan T and Ramamoorthi, Ravi and Ng, Ren},
  booktitle={ECCV},
  year={2020}
}

@article{kerbl2023gaussian,
  title={3D Gaussian Splatting for Real-Time Radiance Field Rendering},
  author={Kerbl, Bernhard and Kopanas, Georgios and Leimk{\"u}hler, Thomas and Drettakis, George},
  journal={ACM Transactions on Graphics (TOG)},
  year={2023}
}

@inproceedings{reconfusion2023,
  title={Reconfusion: 3d reconstruction with diffusion priors},
  author={Wu, Rundi and Mildenhall, Ben and Henzler, Philipp and Park, Keunhong and Gao, Ruiqi and Watson, Daniel and Srinivasan, Pratul P and Verbin, Dor and Barron, Jonathan T and Poole, Ben and others},
  booktitle={Proceedings of the IEEE/CVF conference on computer vision and pattern recognition},
  pages={21551--21561},
  year={2024}
}

@article{wu2025difix3d,
  title={Difix3D+: Improving 3D Reconstructions with Single-Step Diffusion Models},
  author={Wu, Jay Zhangjie and Zhang, Yuxuan and Turki, Haithem and Ren, Xuanchi and Gao, Jun and Shou, Mike Zheng and Fidler, Sanja and Gojcic, Zan and Ling, Huan},
  journal={CVPR},
  year={2025},
  archivePrefix={arXiv},
  eprint={2503.01774}
}

@article{omran2025hybrid,
  title={Hybrid Gaussian Splatting for Novel Urban View Synthesis},
  author={Omran, Mohamed and Zanjani, Farhad and Abati, Davide and Petersen, Jens and Habibian, Amirhossein},
  journal={arXiv preprint arXiv:2510.12308},
  year={2025}
}

@inproceedings{veg2023,
  title={Vegs: View extrapolation of urban scenes in 3d gaussian splatting using learned priors},
  author={Hwang, Sungwon and Kim, Min-Jung and Kang, Taewoong and Kang, Jayeon and Choo, Jaegul},
  booktitle={European Conference on Computer Vision},
  pages={1--18},
  year={2024},
  organization={Springer}
}

@inproceedings{2dgs2023,
  title={2D gaussian splatting for geometrically accurate radiance fields},
  author={Huang, Binbin and Yu, Zehao and Chen, Anpei and Geiger, Andreas and Gao, Shenghua},
  booktitle={ACM SIGGRAPH 2024 conference papers},
  pages={1--11},
  year={2024}
}

@inproceedings{zipnerf2023,
  title={Zip-nerf: Anti-aliased grid-based neural radiance fields},
  author={Barron, Jonathan T and Mildenhall, Ben and Verbin, Dor and Srinivasan, Pratul P and Hedman, Peter},
  booktitle={Proceedings of the IEEE/CVF International Conference on Computer Vision},
  pages={19697--19705},
  year={2023}
}

@inproceedings{instantngp2022,
  title={Instant Neural Graphics Primitives with a Multiresolution Hash Encoding},
  author={Müller, Thomas and Evans, Alex and Schied, Christoph and Keller, Alexander},
  booktitle={SIGGRAPH},
  year={2022}
}

@article{3DGM2024,
  title={Memorize what matters: Emergent scene decomposition from multitraverse},
  author={Li, Yiming and Wang, Zehong and Wang, Yue and Yu, Zhiding and Gojcic, Zan and Pavone, Marco and Feng, Chen and Alvarez, Jose M},
  journal={Advances in Neural Information Processing Systems},
  volume={37},
  pages={108389--108438},
  year={2024}
}

@inproceedings{lu2024scaffold,
  title={Scaffold-gs: Structured 3d gaussians for view-adaptive rendering},
  author={Lu, Tao and Yu, Mulin and Xu, Linning and Xiangli, Yuanbo and Wang, Limin and Lin, Dahua and Dai, Bo},
  booktitle={Proceedings of the IEEE/CVF Conference on Computer Vision and Pattern Recognition},
  pages={20654--20664},
  year={2024}
}

@inproceedings{barron2021mipnerf,
  title={Mip-NeRF: A Multiscale Representation for Anti-Aliasing Neural Radiance Fields},
  author={Barron, Jonathan T and Mildenhall, Ben and Verbin, Daniel and Srinivasan, Pratul P and Hedman, Peter},
  booktitle={ICCV},
  year={2021}
}

@inproceedings{niemeyer2022regnerf,
  title={RegNeRF: Regularizing Neural Radiance Fields for Few-Shot Novel View Synthesis},
  author={Niemeyer, Michael and Barron, Jonathan T and Mildenhall, Ben and others},
  booktitle={CVPR},
  year={2022}
}

@inproceedings{yu2021freenerf,
  title={FreeNeRF: Improving Few-Shot Neural Rendering with Free Viewpoint Priors},
  author={Yu, Alex and Ye, Zexiang and Tancik, Matthew and Kanazawa, Angjoo},
  booktitle={CVPR},
  year={2021}
}

@inproceedings{wu2023dngaussian,
  title={DNGaussian: Depth-Normal Regularized Gaussian Splatting for Improved 3D Reconstruction},
  author={Wu, Jay Zhangjie and others},
  booktitle={CVPR},
  year={2023}
}

@article{xu2023dmv3d,
  title={Dmv3d: Denoising multi-view diffusion using 3d large reconstruction model},
  author={Xu, Yinghao and Tan, Hao and Luan, Fujun and Bi, Sai and Wang, Peng and Li, Jiahao and Shi, Zifan and Sunkavalli, Kalyan and Wetzstein, Gordon and Xu, Zexiang and others},
  journal={arXiv preprint arXiv:2311.09217},
  year={2023}
}

@article{liu20243dgs,
  title={3dgs-enhancer: Enhancing unbounded 3d gaussian splatting with view-consistent 2d diffusion priors},
  author={Liu, Xi and Zhou, Chaoyi and Huang, Siyu},
  journal={Advances in Neural Information Processing Systems},
  volume={37},
  pages={133305--133327},
  year={2024}
}

@inproceedings{ling2024dl3dv,
  title={Dl3dv-10k: A large-scale scene dataset for deep learning-based 3d vision},
  author={Ling, Lu and Sheng, Yichen and Tu, Zhi and Zhao, Wentian and Xin, Cheng and Wan, Kun and Yu, Lantao and Guo, Qianyu and Yu, Zixun and Lu, Yawen and others},
  booktitle={Proceedings of the IEEE/CVF Conference on Computer Vision and Pattern Recognition},
  pages={22160--22169},
  year={2024}
}

@article{salimans2022progressive,
  title={Progressive Distillation for Fast Sampling of Diffusion Models},
  author={Tim Salimans and Jonathan Ho},
  journal={arXiv preprint arXiv:2202.00512},
  year={2022},
  url={https://arxiv.org/abs/2202.00512}
}

@inproceedings{yan2024street,
  title={Street gaussians: Modeling dynamic urban scenes with gaussian splatting},
  author={Yan, Yunzhi and Lin, Haotong and Zhou, Chenxu and Wang, Weijie and Sun, Haiyang and Zhan, Kun and Lang, Xianpeng and Zhou, Xiaowei and Peng, Sida},
  booktitle={European Conference on Computer Vision},
  pages={156--173},
  year={2024},
  organization={Springer}
}

@inproceedings{sauer2024adversarial,
  title={Adversarial diffusion distillation},
  author={Sauer, Axel and Lorenz, Dominik and Blattmann, Andreas and Rombach, Robin},
  booktitle={European Conference on Computer Vision},
  pages={87--103},
  year={2024},
  organization={Springer}
}

@inproceedings{chen2025omnire,
  title={OmniRe: Omni Urban Scene Reconstruction},
  author={Ziyu Chen and Jiawei Yang and Jiahui Huang and Riccardo de Lutio and Janick Martinez Esturo and Boris Ivanovic and Or Litany and Zan Gojcic and Sanja Fidler and Marco Pavone and Li Song and Yue Wang},
  booktitle={The Thirteenth International Conference on Learning Representations},
  year={2025}
}

@inproceedings{cheng2024gaussianpro,
  title={Gaussianpro: 3d gaussian splatting with progressive propagation},
  author={Cheng, Kai and Long, Xiaoxiao and Yang, Kaizhi and Yao, Yao and Yin, Wei and Ma, Yuexin and Wang, Wenping and Chen, Xuejin},
  booktitle={Forty-first International Conference on Machine Learning},
  year={2024}
}

@inproceedings{zhou2024feature,
  title={Feature 3dgs: Supercharging 3d gaussian splatting to enable distilled feature fields},
  author={Zhou, Shijie and Chang, Haoran and Jiang, Sicheng and Fan, Zhiwen and Zhu, Zehao and Xu, Dejia and Chari, Pradyumna and You, Suya and Wang, Zhangyang and Kadambi, Achuta},
  booktitle={Proceedings of the IEEE/CVF Conference on Computer Vision and Pattern Recognition},
  pages={21676--21685},
  year={2024}
}

@inproceedings{jain2021putting,
  title={Putting nerf on a diet: Semantically consistent few-shot view synthesis},
  author={Jain, Ajay and Tancik, Matthew and Abbeel, Pieter},
  booktitle={Proceedings of the IEEE/CVF international conference on computer vision},
  pages={5885--5894},
  year={2021}
}

@inproceedings{zhou2024drivinggaussian,
  title={Drivinggaussian: Composite gaussian splatting for surrounding dynamic autonomous driving scenes},
  author={Zhou, Xiaoyu and Lin, Zhiwei and Shan, Xiaojun and Wang, Yongtao and Sun, Deqing and Yang, Ming-Hsuan},
  booktitle={Proceedings of the IEEE/CVF conference on computer vision and pattern recognition},
  pages={21634--21643},
  year={2024}
}

@inproceedings{wu2023mars,
  title={Mars: An instance-aware, modular and realistic simulator for autonomous driving},
  author={Wu, Zirui and Liu, Tianyu and Luo, Liyi and Zhong, Zhide and Chen, Jianteng and Xiao, Hongmin and Hou, Chao and Lou, Haozhe and Chen, Yuantao and Yang, Runyi and others},
  booktitle={CAAI International Conference on Artificial Intelligence},
  pages={3--15},
  year={2023},
  organization={Springer}
}

@article{zhou2024hugsim,
  title={Hugsim: A real-time, photo-realistic and closed-loop simulator for autonomous driving},
  author={Zhou, Hongyu and Lin, Longzhong and Wang, Jiabao and Lu, Yichong and Bai, Dongfeng and Liu, Bingbing and Wang, Yue and Geiger, Andreas and Liao, Yiyi},
  journal={arXiv preprint arXiv:2412.01718},
  year={2024}
}

@inproceedings{lai2018learning,
  title={Learning blind video temporal consistency},
  author={Lai, Wei-Sheng and Huang, Jia-Bin and Wang, Oliver and Shechtman, Eli and Yumer, Ersin and Yang, Ming-Hsuan},
  booktitle={Proceedings of the European conference on computer vision (ECCV)},
  pages={170--185},
  year={2018}
}

@article{parmar2024one,
  title={One-step image translation with text-to-image models},
  author={Parmar, Gaurav and Park, Taesung and Narasimhan, Srinivasa and Zhu, Jun-Yan},
  journal={arXiv preprint arXiv:2403.12036},
  year={2024}
}

@inproceedings{han2025extrapolated,
  title={Extrapolated urban view synthesis benchmark},
  author={Han, Xiangyu and Jia, Zhen and Li, Boyi and Wang, Yan and Ivanovic, Boris and You, Yurong and Liu, Lingjie and Wang, Yue and Pavone, Marco and Feng, Chen and others},
  booktitle={Proceedings of the IEEE/CVF International Conference on Computer Vision},
  pages={28718--28728},
  year={2025}
}

@article{ni2025lane,
  title={Para-Lane: Multi-Lane Dataset Registering Parallel Scans for Benchmarking Novel View Synthesis},
  author={Ni, Ziqian and Du, Sicong and Hou, Zhenghua and Wu, Chenming and Yang, Sheng},
  journal={arXiv preprint arXiv:2502.15635},
  year={2025}
}

@inproceedings{caesar2020nuscenes,
  title={nuscenes: A multimodal dataset for autonomous driving},
  author={Caesar, Holger and Bankiti, Varun and Lang, Alex H and Vora, Sourabh and Liong, Venice Erin and Xu, Qiang and Krishnan, Anush and Pan, Yu and Baldan, Giancarlo and Beijbom, Oscar},
  booktitle={Proceedings of the IEEE/CVF conference on computer vision and pattern recognition},
  pages={11621--11631},
  year={2020}
}

@article{chen2024pgsr,
  title={Pgsr: Planar-based gaussian splatting for efficient and high-fidelity surface reconstruction},
  author={Chen, Danpeng and Li, Hai and Ye, Weicai and Wang, Yifan and Xie, Weijian and Zhai, Shangjin and Wang, Nan and Liu, Haomin and Bao, Hujun and Zhang, Guofeng},
  journal={IEEE Transactions on Visualization and Computer Graphics},
  year={2024},
  publisher={IEEE}
}

@software{alpasim_2025,
  author       = {
    NVIDIA and
    Yulong Cao and
    Riccardo de Lutio and
    Sanja Fidler and
    Guillermo Garcia Cobo and
    Zan Gojcic and
    Maximilian Igl and
    Boris Ivanovic and
    Peter Karkus and
    Janick Martinez Esturo and
    Marco Pavone and
    Aaron Smith and
    Ellie Tanimura and
    Michal Tyszkiewicz and
    Michael Watson and
    Qi Wu and
    Le Zhang
  },
  title        = {AlpaSim: A Modular, Lightweight, and Data-Driven Research Simulator for Autonomous Driving},
  year         = {2025},
  month        = {October},
  url          = {https://github.com/NVlabs/alpasim},
}

@article{ali2025world,
  title={World simulation with video foundation models for physical ai},
  author={Ali, Arslan and Bai, Junjie and Bala, Maciej and Balaji, Yogesh and Blakeman, Aaron and Cai, Tiffany and Cao, Jiaxin and Cao, Tianshi and Cha, Elizabeth and Chao, Yu-Wei and others},
  journal={arXiv preprint arXiv:2511.00062},
  year={2025}
}

@article{russell2025gaia,
  title={Gaia-2: A controllable multi-view generative world model for autonomous driving},
  author={Russell, Lloyd and Hu, Anthony and Bertoni, Lorenzo and Fedoseev, George and Shotton, Jamie and Arani, Elahe and Corrado, Gianluca},
  journal={arXiv preprint arXiv:2503.20523},
  year={2025}
}

@article{zhou2025hugsim,
  title={Hugsim: A real-time, photo-realistic and closed-loop simulator for autonomous driving},
  author={Zhou, Hongyu and Lin, Longzhong and Wang, Jiabao and Lu, Yichong and Bai, Dongfeng and Liu, Bingbing and Wang, Yue and Geiger, Andreas and Liao, Yiyi},
  journal=PAMI,
  year={2025},  
}

@inproceedings{gao2025magicdrive,
  title={MagicDrive-V2: High-resolution long video generation for autonomous driving with adaptive control},
  author={Gao, Ruiyuan and Chen, Kai and Xiao, Bo and Hong, Lanqing and Li, Zhenguo and Xu, Qiang},
  booktitle=ICCV,
  year={2025}
}

@inproceedings{ljungbergh2024neuroncap,
  title={Neuroncap: Photorealistic closed-loop safety testing for autonomous driving},
  author={Ljungbergh, William and Tonderski, Adam and Johnander, Joakim and Caesar, Holger and {\AA}str{\"o}m, Kalle and Felsberg, Michael and Petersson, Christoffer},
  booktitle=eccv,
  year={2024},
}

@inproceedings{yang2023unisim,
  title={Unisim: A neural closed-loop sensor simulator},
  author={Yang, Ze and Chen, Yun and Wang, Jingkang and Manivasagam, Sivabalan and Ma, Wei-Chiu and Yang, Anqi Joyce and Urtasun, Raquel},
  booktitle=cvpr,
  year={2023}
}

@article{li2025recogdrive,
  title={Recogdrive: A reinforced cognitive framework for end-to-end autonomous driving},
  author={Li, Yongkang and Xiong, Kaixin and Guo, Xiangyu and Li, Fang and Yan, Sixu and Xu, Gangwei and Zhou, Lijun and Chen, Long and Sun, Haiyang and Wang, Bing and others},
  journal={arXiv preprint arXiv:2506.08052},
  year={2025}
}

@inproceedings{renz2025simlingo,
  title={Simlingo: Vision-only closed-loop autonomous driving with language-action alignment},
  author={Renz, Katrin and Chen, Long and Arani, Elahe and Sinavski, Oleg},
  booktitle=cvpr,  
  year={2025}
}

@article{kim2024openvla,
  title={Openvla: An open-source vision-language-action model},
  author={Kim, Moo Jin and Pertsch, Karl and Karamcheti, Siddharth and Xiao, Ted and Balakrishna, Ashwin and Nair, Suraj and Rafailov, Rafael and Foster, Ethan and Lam, Grace and Sanketi, Pannag and others},
  journal={arXiv preprint arXiv:2406.09246},
  year={2024}
}

@article{zanjani2025gaussian,
  title={Gaussian Splatting is an Effective Data Generator for 3D Object Detection},
  author={Zanjani, Farhad G and Abati, Davide and Wiggers, Auke and Kalatzis, Dimitris and Petersen, Jens and Cai, Hong and Habibian, Amirhossein},
  journal={arXiv preprint arXiv:2504.16740},
  year={2025}
}
}

\clearpage
\appendix
\onecolumn

\begin{center}
\textbf{\Large Enhancing Novel View Synthesis via \\ Geometry Grounded Set Diffusion \\[0.5cm] --- Supplementary Material ---}
\end{center}

\vspace{0.5cm}

\setcounter{section}{1}
\subsection{Datasets and Experimental Setup}
\label{sec:sup_dataset}
To comprehensively assess the extrapolation capability of novel view synthesis (NVS) methods under challenging conditions—including lateral camera translation, camera rotation, combined translation and rotation, temporal extrapolation, camera‑view sparsity, dynamic scene variation, and both daytime and nighttime environments—we conduct experiments across four diverse datasets. Below, we detail the experimental setup for each dataset.

\subsubsection{EUVS Dataset}
The Extrapolated Urban View Synthesis (EUVS) dataset~\cite{han2025extrapolated} is a large-scale autonomous driving benchmark designed to evaluate NVS performance under significant viewpoint shifts, representative of vision-centric autonomous driving systems. It contains 104 real-world urban scenes captured across multiple vehicles, routes, and camera configurations, along with official train/test splits. EUVS specifies three evaluation protocols:
\begin{enumerate}
    \item \emph{Translation-only}: cross-lane lateral displacement;
    \item \emph{Rotation-only}: holding out side-view cameras for testing;
    \item \emph{Translation + rotation}: \eg, driving on different routes in cross sections (considered the most challenging setting).
\end{enumerate}

\paragraph{Training data generation for diffusion enhancers.}
Each EUVS scene consists of multiple traversals. Diffusion-based enhancers require paired distorted/clean images. To construct these pairs, we fit a 3DGS model on one traversal and render images at the camera poses of the remaining traversals. This produces aligned low-quality (rendered/distorted) and high-quality (captured/clean) image pairs. We repeat this process by permuting over all trajectory combinations within each scene.

\paragraph{Test data.}
To ensure fair evaluation, we adopt a strict scene-level 60\%/40\% train/test split, preventing any overlap or near-duplicate content across sets. For the 40\% held-out scenes, we follow the official EUVS protocols: after fitting 3DGS on the training images, we evaluate \name and all baselines on the prescribed rendered test views.

\paragraph{Dynamic region handling.}
Scenes contain multiple asynchronous traversals, causing dynamic objects to differ between training and inference views. Consequently, consistent reconstruction of dynamic regions is not feasible for 3DGS. Following~\cite{han2025extrapolated}, dynamic regions are excluded during both training and evaluation using the dynamic-region masks provided in dataset.

\subsubsection{Para-Lane Dataset}
The Para-Lane dataset~\cite{ni2025lane} provides multi-lane urban driving sequences captured from parallel trajectories, enabling evaluation of cross-lane NVS under realistic lane shifts. It includes three cameras and over 16K front-view and 64K surround-view images recorded from synchronized multi-pass runs. As with EUVS, we adopt a strict scene-level 60\%/40\% split for training diffusion-enhancement models.

For 3DGS reconstruction, we fit one trajectory at a time and treat the remaining parallel trajectories as test sequences—each requiring view-specific enhancement at its corresponding extrapolated viewpoints.

\subsubsection{nuScenes Dataset}
To evaluate performance under strong scene dynamics (moving vehicles, pedestrians, and cyclists), diverse weather conditions, and both day and night settings, we use the nuScenes dataset~\cite{caesar2020nuscenes}, which contains 750 sequences from six cameras in complex urban environments.

Since each scene contains only a single traversal, we construct a synthetic temporal-extrapolation benchmark. Each sequence is split into two disjoint temporal segments: the first 10 seconds are used to fit a dynamic 3DGS model (OmniRe), and the subsequent 5 seconds serve as novel future views. These future frames are grouped into five temporal buckets according to timestamp, allowing controlled evaluation across increasing extrapolation distances. The final bucket (14–15 sec.) is the most challenging due to accumulated 3DGS drift, dynamic object motion, and scene incompleteness. Dynamic objects are retained in both training and evaluation.

For diffusion-enhancer training, we employ a scene-level random split: 650 training scenes and 150 test scenes. Concretely, after sorting scene names, every fifth scene is assigned to the test set.

\subsubsection{DL3DV Dataset}
To assess performance beyond autonomous driving, we additionally evaluate on the DL3DV-10K dataset~\cite{ling2024dl3dv}, which spans a broad spectrum of bounded and unbounded indoor and outdoor scenes. This enables direct comparison with 3DGS‑Enhancer~\cite{liu20243dgs} under varying input sparsity levels (3, 6, and 12 training views). We follow the original splitting strategy from~\cite{liu20243dgs}, comprising 130 training scenes and 20 test scenes.

For each sparsity level, we adopt the exact train/test camera selections defined in~\cite{liu20243dgs}. A 3DGS model is fitted using only the sparse training views, and rendered images at the held-out camera poses are used to evaluate enhancement quality.

\subsection{Implementation Details}
\paragraph{Training Configuration.}
The proposed \name framework is trained on packets consisting of $N + M$ frames, where $N$ denotes the number of reference views and $M$ denotes target views. For the EUVS, Para-Lane, and DL3DV datasets, the packet length is fixed to $8$. For nuScenes, which contains $6$ cameras, we use a packet length of $12$ (i.e., $6$ reference and $6$ target views). During inference, the packet length is increased to $16$ for EUVS, Para-Lane, and DL3DV, and to $24$ for nuScenes.

Training is conducted on four NVIDIA A100-80G GPUs with a total batch size of $4$, using approximately $40$K iterations. For classifier-free guidance (CFG), we apply a dropout rate of $15\%$ to both the C-map and camera pose inputs. At inference time, the CFG scale is set to $2.0$. All experiments are performed at a resolution of $576 \times 1024$ pixels. Following DiFix3D~\cite{wu2025difix3d}, we fine-tune SD-Turbo using a reduced noise level ($\tau = 200$ instead of $\tau = 1000$), rather than introducing random Gaussian noise.

\paragraph{3D Priors.}
For each packet containing reference and target frames, camera poses and their corresponding C-maps are transformed from world coordinates into the coordinate frame of the first target camera. Prior to computing Plücker ray embeddings, we normalize the camera poses. This is essential because the cross-product term $\mathbf{o} \times \mathbf{d}$ is not scale-invariant, as it directly depends on the magnitude of $\mathbf{o}$. Moreover, camera poses estimated by Structure-from-Motion (SfM) often exhibit scale inconsistencies across scenes and datasets.

To address these issues, we normalize all camera positions during both training and inference. Specifically, the positions are scaled such that the maximum pairwise distance among cameras within a packet is equal to unity. This normalization alleviates scale ambiguity, stabilizes training, and ensures consistent behavior across diverse scene configurations.

\subsubsection{Additional Ablations}
\label{sec:sup_ablation}
\begin{table}[htbp]
\centering
\caption{Ablating the architecture choices on enhancement qualities.
}

\setlength{\tabcolsep}{12pt} % default is 6pt
\renewcommand{\arraystretch}{1.1} % row spacing (default = 1.0)

\resizebox{.6\textwidth}{!}{
\begin{tabular}[t]{l|ccc}
\toprule
\textbf{Architecture} & \textbf{PSNR} & \textbf{SSIM} & \textbf{LPIPS} \\
\midrule

w/o LoRA & 18.604 & 0.584 & 0.287 \\
cross-attn & 18.673 & 0.587 & 0.268 \\
\cellcolor{lightblue}Ours & \cellcolor{lightblue}\textbf{18.74} & \cellcolor{lightblue}\textbf{0.587} & \cellcolor{lightblue}\textbf{0.265} \\
\bottomrule
\end{tabular}
}
\label{tab:additional_ablation}
\vspace{-4pt}
\end{table}

We evaluate two architectural variants: (i) removing the LoRA-based fine-tuning of the VAE decoder (\emph{``w/o LoRA''}), and (ii) replacing additive conditioning with cross-attention in the 2D UNet (\emph{``cross-attn''}). As shown in \Cref{tab:additional_ablation}, LoRA fine-tuning yields a modest improvement in LPIPS, suggesting enhanced perceptual fidelity and realism. The cross-attention variant performs comparably to the baseline, indicating that it provides a viable alternative mechanism for injecting latent priors into the UNet.

\subsubsection{Consistency Evaluation}
\begin{table}[htbp]
\centering
\caption{Evaluating the enhancement consistency.
}

\setlength{\tabcolsep}{12pt} % default is 6pt
\renewcommand{\arraystretch}{1.1} % row spacing (default = 1.0)

\resizebox{.7\textwidth}{!}
{\begin{tabular}{l|cccc}
    \toprule
    & \textbf{Consistency}$\downarrow$ & \textbf{PSNR} & \textbf{SSIM} & \textbf{LPIPS} \\
    \midrule
    OmniRe & 111.0 & 17.15 & 0.645 & 0.448 \\
    \quad + DiFix++ & 110.2 & 18.74 & 0.696 & 0.375 \\
    \cellcolor{lightblue}\quad + \name & \cellcolor{lightblue}\textbf{108.9} & \cellcolor{lightblue}\textbf{21.58} & \cellcolor{lightblue}\textbf{0.716} & \cellcolor{lightblue}\textbf{0.303}  \\
    \bottomrule
\end{tabular}}
\label{tab:consistency}
\vspace{-0.5em}
\end{table}

We assess consistency on nuScenes using the Warp error metric~\cite{lai2018learning}. As reported in \Cref{tab:consistency}, \name improves consistency over OmniRe and DiFix++. Notably, despite substantial gains in image-quality metrics such as PSNR, SSIM, and LPIPS, diffusion-based enhancers (DiFix++ and \name) provide only limited improvements in neural reconstruction consistency. This is expected, as geometry-based reconstruction models already enforce strong cross-view consistency through explicit geometric constraints, leaving limited headroom for further enhancement.

\subsection{Qualitative Video Results}
For qualitative visualizations of enhanced test videos, please refer to the project webpage provided in HTML format.

\end{document}